%% file: main.tex
\documentclass[10pt,journal,compsoc]{IEEEtran}

\usepackage{url}            
\usepackage{amsmath,amsfonts,amssymb}       
\usepackage{multirow}
\usepackage{algorithm,algorithmic}
\usepackage{xcolor}
\usepackage[normalem]{ulem}
\usepackage{arydshln}
\usepackage{caption}
\usepackage{enumitem}
\usepackage{wrapfig}
\usepackage{microtype}
\usepackage{graphicx}
\usepackage{subfigure}
\usepackage{booktabs} 
\usepackage{cutwin}
\usepackage{bbm}
\usepackage[pangram]{blindtext}
\usepackage[calc]{adjustbox}

\usepackage{tabularx}
\usepackage{color, colortbl}
\usepackage{kotex}
\usepackage{lipsum}

\usepackage{enumitem} 

\makeatletter
\def\hlinewd#1{%
	\noalign{\ifnum0=`}\fi\hrule \@height #1 %
	\futurelet\reserved@a\@xhline}
\makeatother

\renewcommand{\algorithmiccomment}[1]{\bgroup\hfill$\triangleright$~#1\egroup}

\definecolor{crimson}{rgb}{0.86, 0.08, 0.24}
\definecolor{orange-red}{rgb}{1.0, 0.27, 0.0}

\newcommand{\newor}{%
  \mathbin{%
    {\vee}\mspace{-2.9mu}
  }%
}
\usepackage{bm}
\newcommand{\bs}[1] {\bm{#1}}

\usepackage{svg}
\DeclareMathOperator*{\minimize}{minimize}
\DeclareMathOperator*{\logicalor}{\vee}

\usepackage{gensymb}
\usepackage{marvosym}

\usepackage{mathtools,tikz}
\DeclareRobustCommand\sampleline[1]{%
  \tikz\draw[#1] (0,0) (0,\the\dimexpr\fontdimen22\textfont2\relax)
  -- (2em,\the\dimexpr\fontdimen22\textfont2\relax);%
}

\definecolor{Red}{rgb}{1,0,0}
\definecolor{Blue}{rgb}{0,0,0.8}
\definecolor{Green}{rgb}{0,0.7,0.2}
\definecolor{airforceblue}{rgb}{0.36, 0.54, 0.66}
\definecolor{ao(english)}{rgb}{0.0, 0.5, 0.0}
\definecolor{azure(colorwheel)}{rgb}{0.0, 0.5, 1.0}
\definecolor{crimson}{rgb}{0.86, 0.08, 0.24}
\definecolor{darkcerulean}{rgb}{0.03, 0.27, 0.49}
\definecolor{cobalt}{rgb}{0.0, 0.28, 0.67}
\definecolor{rosegold}{rgb}{0.72, 0.43, 0.47}
\definecolor{orange-red}{rgb}{1.0, 0.27, 0.0}
\definecolor{mountainmeadow}{rgb}{0.19, 0.73, 0.56}
\definecolor{malachite}{rgb}{0.04, 0.85, 0.32}
\definecolor{darkblue}{rgb}{0.0, 0.0, 0.55}

\definecolor{customblue}{rgb}{0.2, 0.35, 0.8}
\definecolor{gg}{gray}{0.9}

\usepackage[capitalize]{cleveref}
\crefname{section}{Sec.}{Secs.}
\Crefname{section}{Section}{Sections}
\Crefname{table}{Table}{Tables}
\crefname{table}{Tab.}{Tabs.}

\usepackage{makecell}

\newcommand{\eat}[1]{{}}
\creflabelformat{equation}{#2\textup{#1}#3} 

\usepackage{mathtools,tikz,caption}
\DeclareRobustCommand\sampleline[1]{%
  \tikz\draw[#1] (0,0) (0,\the\dimexpr\fontdimen22\textfont2\relax)
  -- (2em,\the\dimexpr\fontdimen22\textfont2\relax);%
}

\newtheorem{theorem}{Theorem}[section]

\newtheorem{lemma}[theorem]{Lemma}

\ifCLASSOPTIONcompsoc
  \usepackage[nocompress]{cite}
\else
  \usepackage{cite}
\fi

\begin{document}

%
\title{Continual Learning: Forget-free Winning Subnetworks for Video Representations}

\author{Haeyong~Kang,
        Jaehong~Yoon, 
        Sung Ju Hwang~\IEEEmembership{Member,~IEEE},
        and~Chang~D.~Yoo~\IEEEmembership{Senior~Member,~IEEE}
\IEEEcompsocitemizethanks{\IEEEcompsocthanksitem Haeyong Kang and Chang D. Yoo are with the School of Electrical Engineering, KAIST, Republic of Korea, 34141. Email: \{haeyong.kang, cd\_yoo\}@kaist.ac.kr}
\IEEEcompsocitemizethanks{\IEEEcompsocthanksitem Jaehong~Yoon and Sung Ju Hwang are with the School of Computing, KAIST, Republic of Korea, 34141. Email: \{jaehong.yoon, sjhwang82\}@kaist.ac.kr} 
\IEEEcompsocitemizethanks{\IEEEcompsocthanksitem Corresponding author: Haeyong Kang and Chang D. Yoo}

}
\markboth{IEEE TRANSACTIONS ON PATTERN ANALYSIS AND MACHINE INTELLIGENCE (TPAMI, 2024)}%
{Shell \MakeLowercase{\textit{et al.}}: Bare Demo of IEEEtran.cls for Computer Society Journals}

\IEEEtitleabstractindextext{%
\input{1_abstract}

\begin{IEEEkeywords}
Continual Learning (CL), Task Incremental Learning (TIL),Task-agnostic Incremental Learning (TaIL), Video Incremental Learning (VIL), Few-shot Class Incremental Learning (FSCIL), Regularized Lottery Ticket Hypothesis (RLTH), Wining SubNetworks (WSN), Soft-Subnetwork (SoftNet), Fourier Subneural Operator (FSO)
\end{IEEEkeywords}}

\maketitle

\IEEEdisplaynontitleabstractindextext

\IEEEpeerreviewmaketitle

\input{2_intro}

\input{3_related_work}

\input{4_approach}

\input{5_experiment}

\input{6_conclusion}

\noindent 

\textbf{Acknowledgement}. This work was supported by Institute for Information \& communications Technology Promotion (IITP) grant funded by (No. RS-2021-II211381), (RS-2022-II220184, 2022-0-00184), and (NRF- 2022R1A2C2012706).


\bibliographystyle{IEEEtran}
\bibliography{reference}

\vspace{-0.3in}


\begin{IEEEbiography}[{\includegraphics[width=1in,height=1.25in,clip,keepaspectratio]{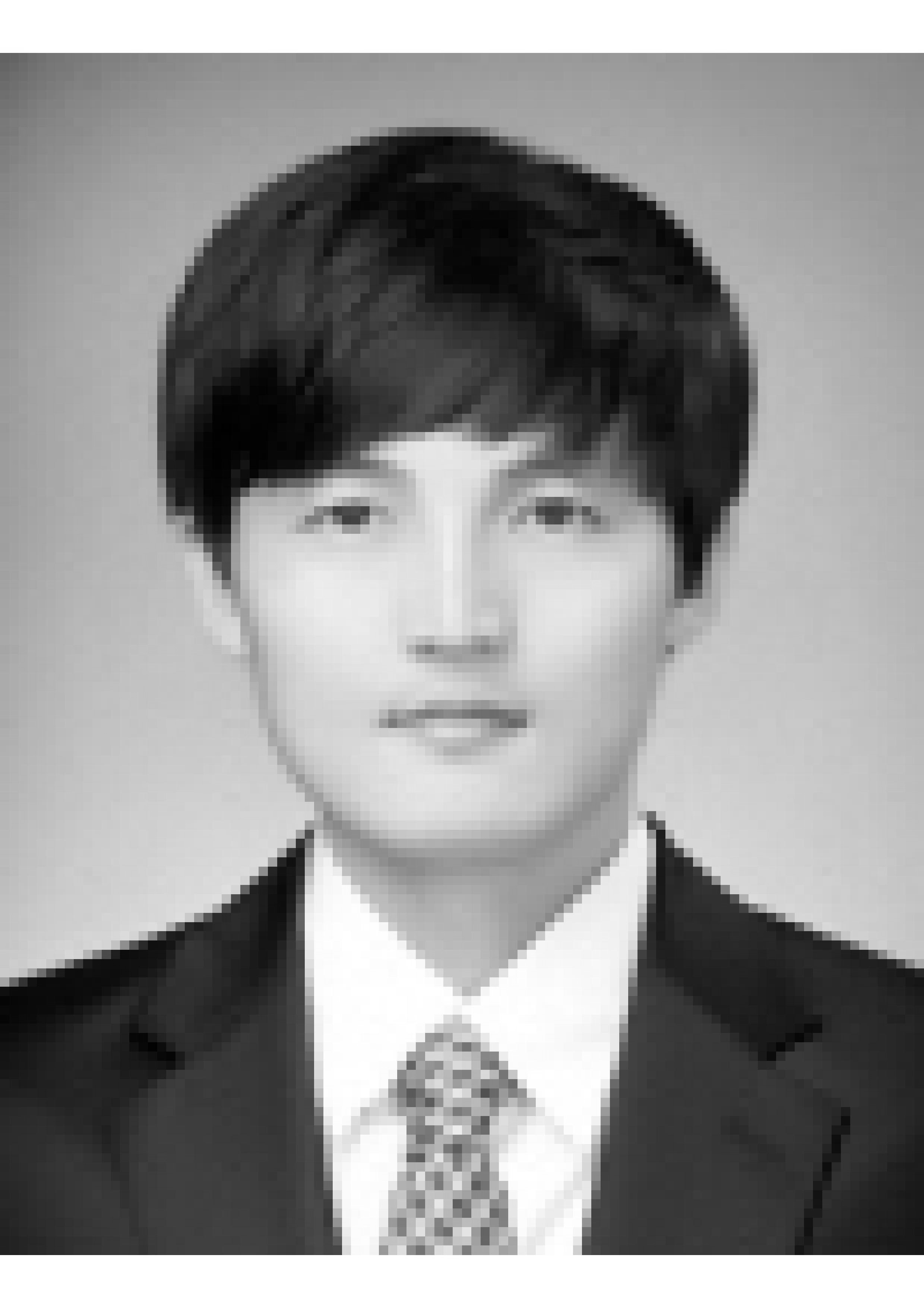}}]{Haeyong Kang}(Member, IEEE), (S'05) received the M.S. degree in Systems and Information Engineering from University of Tsukuba in 2007. From April 2007 to October 2010, he worked as an associate research engineer at LG Electronics. With working experiences at the Korea Institute of Science and Technology (KIST) and the University of Tokyo, He received a Ph.D. at the School of Electrical Engineering, the Korea Advanced Institute of Science and Technology (KAIST) with a dissertation on forget-free continual learning in 2023. He is pursuing research such as unbiased machine learning and continual learning as a postdoctoral researcher at KAIST.
\end{IEEEbiography}

\vspace{-13mm}

\begin{IEEEbiography}[{\includegraphics[width=1in,height=1.25in,clip,keepaspectratio]{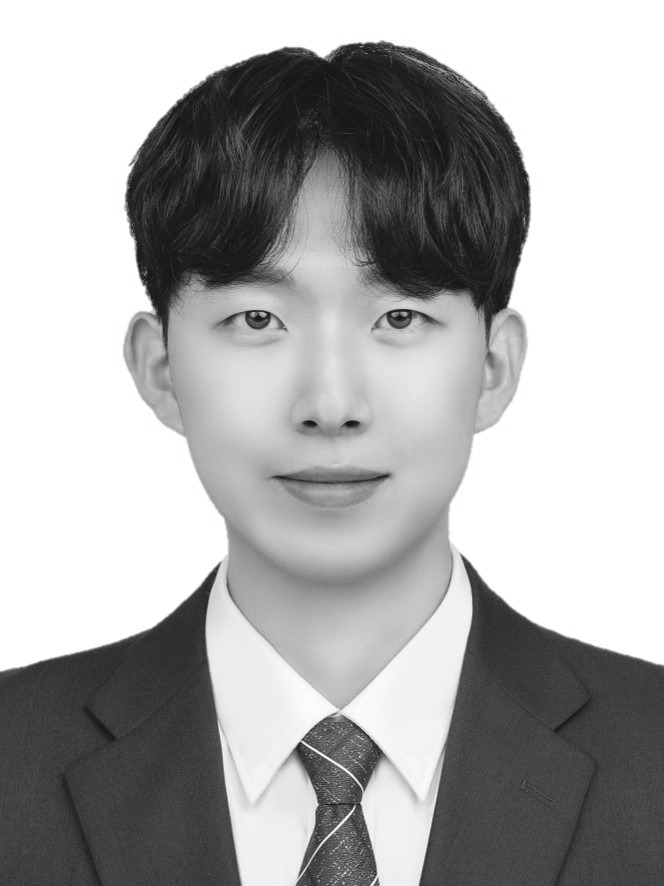}}]{Jaehong Yoon}
(Member, IEEE) He received the B.S. and M.S. degrees in Computer Science from Ulsan National Institute of Science and Technology (UNIST), and received the Ph.D. degree in the School of Computing from Korea Advanced Institute of Science and Technology (KAIST). He is currently working as a postdoctoral research associate at the University of North Carolina at Chapel Hill. His current research interests include multimodal learning, video understanding, efficient deep learning, continual learning, and learning with real-world data.
\end{IEEEbiography}

\vspace{-13mm}

\begin{IEEEbiography}[{\includegraphics[width=1in,height=1.25in,clip,keepaspectratio]{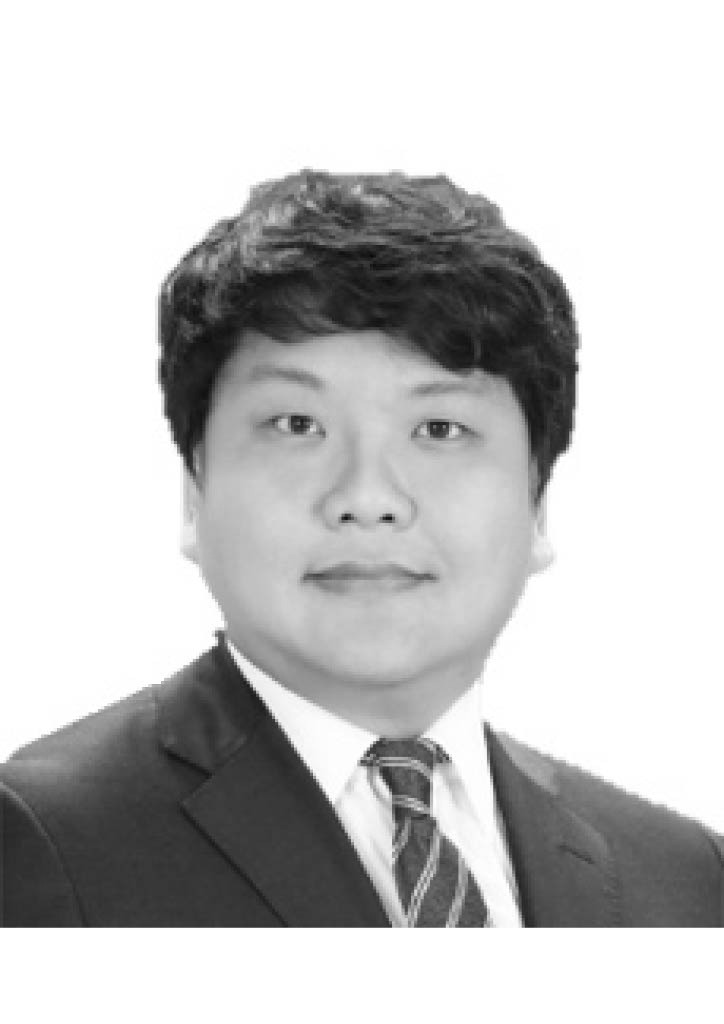}}]{Sung Ju Hwang}
(Member, IEEE) He received the B.S. degree in Computer Science and Engineering from Seoul National University. He received the M.S. and Ph.D. degrees in Computer Science from The University of Texas at Austin. From September 2013 to August 2014, he was a postdoctoral research associate at Disney Research. From September 2013 to December 2017, he was an assistant professor in the School of Electric and Computer Engineering at UNIST. Since 2017, he has been on the faculty at the Korea Advanced Institute of Science and Technology (KAIST), where he is currently a KAIST Endowed Chair Professor in the Kim Jaechul School of Artificial Intelligence and School of Computing at KAIST. 
\end{IEEEbiography}

\vspace{-13mm}

\begin{IEEEbiography}[{\includegraphics[width=1in,height=1.25in,clip,keepaspectratio]{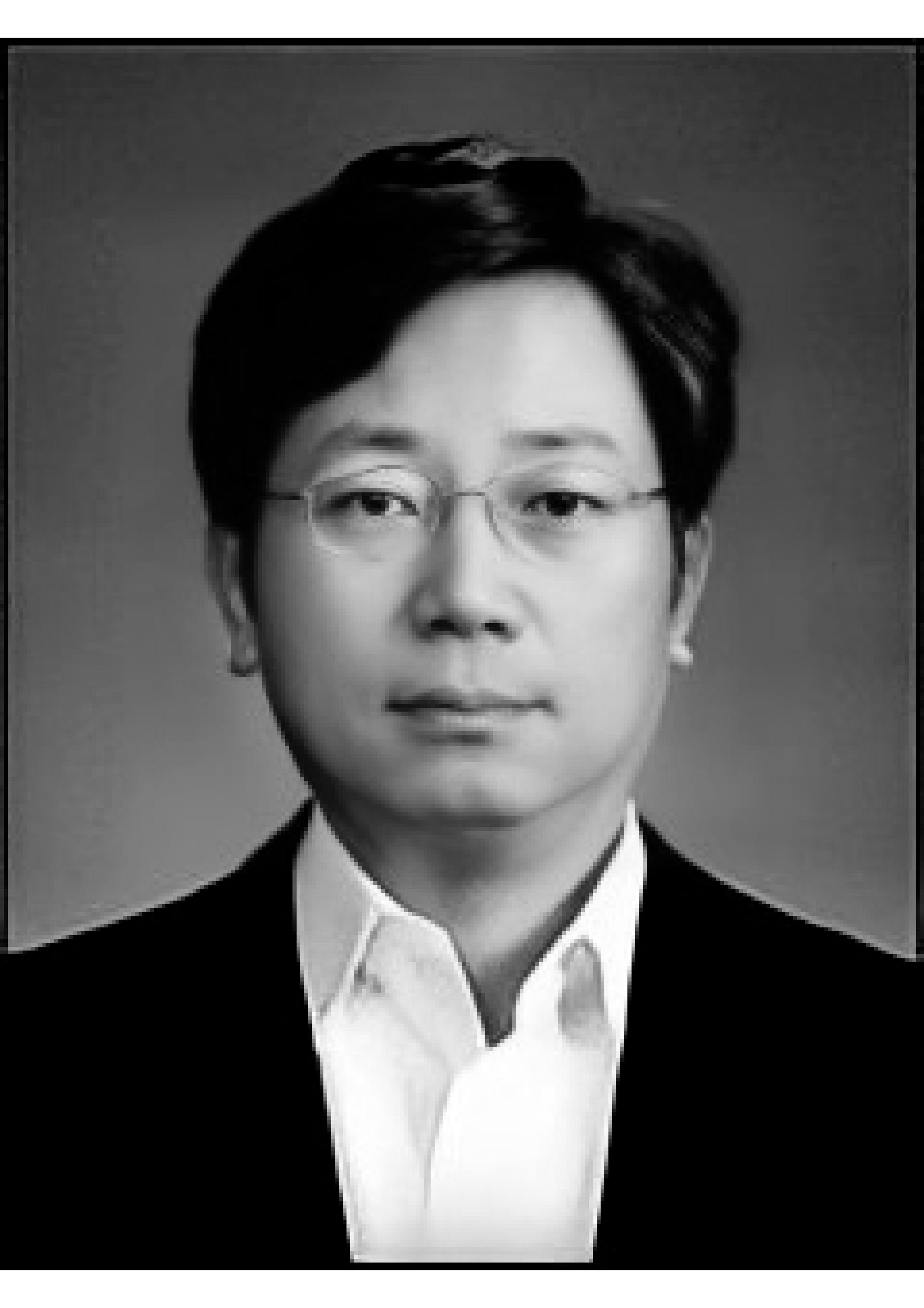}}]{Chang D. Yoo}
(Senior Member, IEEE) He received the B.S. degree in Engineering and Applied Science from the California Institute of Technology, the M.S. degree in Electrical Engineering from Cornell University, and the Ph.D. degree in Electrical Engineering from the Massachusetts Institute of Technology. From January 1997 to March 1999, he was Senior Researcher at Korea Telecom (KT). Since 1999, he has been on the faculty at the Korea Advanced Institute of Science and Technology (KAIST), where he is currently a Full Professor with tenure in the School of Electrical Engineering and an Adjunct Professor in the Department of Computer Science. He also served as Dean of the Office of Special Projects and Dean of the Office of International Relations.
\end{IEEEbiography}

\appendix 
\input{7_appendix}



\end{document}

%% file: 1_abstract.tex
\begin{abstract}
Inspired by the Lottery Ticket Hypothesis (LTH), which highlights the existence of efficient subnetworks within larger, dense networks, a high-performing Winning Subnetwork (WSN) in terms of task performance under appropriate sparsity conditions is considered for various continual learning tasks. It leverages pre-existing weights from dense networks to achieve efficient learning in Task Incremental Learning (TIL) and Task-agnostic Incremental Learning (TaIL) scenarios. In Few-Shot Class Incremental Learning (FSCIL), a variation of WSN referred to as the Soft subnetwork (SoftNet) is designed to prevent overfitting when the data samples are scarce. Furthermore, the sparse reuse of WSN weights is considered for Video Incremental Learning (VIL). The use of Fourier Subneural Operator (FSO) within WSN is considered. It enables compact encoding of videos and identifies reusable subnetworks across varying bandwidths. We have integrated FSO into different architectural frameworks for continual learning, including VIL, TIL, and FSCIL. Our comprehensive experiments demonstrate FSO's effectiveness, significantly improving task performance at various convolutional representational levels. Specifically, FSO enhances higher-layer performance in TIL and FSCIL and lower-layer performance in VIL.
\end{abstract}

%% file: 2_intro.tex
\section{Introduction}

\IEEEPARstart{C}{ontinual} Learning (CL), also known as Lifelong Learning \cite{ThrunS1995,rusu2016progressive,zenke2017continual, hassabis2017neuroscience}, is a learning paradigm where a series of tasks are learned sequentially. The principle objective of continual learning is to replicate human cognition, characterized by the ability to learn new concepts or skills incrementally throughout one's lifespan. An optimal continual learning algorithm could facilitate a positive forward and backward transfer, leveraging the knowledge gained from previous tasks to solve new ones while also updating its understanding of previous tasks with the new knowledge. However, building successful continual learning algorithms is challenging due to the occurrence of \textit{catastrophic forgetting} or \textit{catastrophic interference}~\cite{McCloskey1989}, a phenomenon where the performance of the model on previous tasks significantly deteriorates when learning new tasks. This can make it challenging to retain the knowledge acquired from previous tasks, ultimately leading to a decline in overall performance. To tackle the catastrophic forgetting problem in continual learning, numerous approaches have been proposed, which can be broadly classified as follows: (1) Regularization-based methods~\cite{Kirkpatrick2017, chaudhry2020continual, Jung2020, titsias2019functional, mirzadeh2020linear} aim to keep the learned information of past tasks during continual training aided by sophisticatedly designed regularization terms, (2) Rehearsal-based methods~\cite{rebuffi2017icarl, riemer2018learning, chaudhry2018efficient, chaudhry2019continual, Saha2021, sarfraz2023error} utilize a set of real or synthesized data from the previous tasks and revisit them, and (3) Architecture-based methods~\cite{mallya2018piggyback, Serra2018, li2019learn, wortsman2020supermasks, kang2022forget, kang2022soft} propose to minimize the inter-task interference via newly designed architectural components. 

\input{materials/3_concept}

Despite the remarkable success of recent works on rehearsal- and architecture-based continual learning, most current methods request external memory as new tasks arrive, making the model difficult to scale to larger and more complex tasks. Rehearsal-based CL requires additional storage to store the replay buffer or generative models, and architecture-based methods leverage additional model capacity to account for new tasks. These trends lead to an essential question: how can we build a memory-efficient CL model that does not exceed the backbone network's capacity or even requires a much smaller capacity? Several studies have shown that deep neural networks are over-parameterized~\cite{Denil2013, Han2016learning_both_weights_struct, Li2016pruning_convnets} and thus removing redundant/unnecessary weights can achieve on-par or even better performance than the original dense network. More recently, Lottery Ticket Hypothesis (LTH)~\cite{frankle2018lottery} demonstrates the existence of sparse subnetworks, named \emph{winning tickets}, that preserve the performance of a dense network. However, searching for optimal winning tickets during continual learning with iterative pruning methods requires repetitive pruning and retraining for each arriving task, which could be more impractical. 

To tackle the issues of external replay buffer and capacity, we suggest a novel CL method, which finds the high-performing \textit{Winning SubNetwork} referred to as WSN~\cite{kang2022forget} given tasks without the need for retraining and rewinding, as shown in \Cref{fig:concept_comparison} (d). Also, we set previous pruning-based CL approaches \cite{mallya2018piggyback, wortsman2020supermasks} (see \Cref{fig:concept_comparison} (a)) to baselines of architectures, which obtain task-specific subnetworks given a pre-trained backbone network. Our WSN incrementally learns model weights and task-adaptive binary masks (the subnetworks) within the neural network. To allow the forward transfer when a model learns on a new task, we reuse the learned subnetwork weights for the previous tasks, however selectively, as opposed to using all the weights \cite{mallya2018packnet} (see \Cref{fig:concept_comparison} (b)), that may lead to biased transfer. Further, the WSN eliminates the threat of catastrophic forgetting during continual learning by freezing the subnetwork weights for the previous tasks and does not suffer from the negative transfer, unlike \cite{YoonJ2018iclr} (see \Cref{fig:concept_comparison} (c)), that subnetwork weights for the previous tasks can be updated when training on the new sessions. Moreover, we observed that subnetworks could overfit limited task sample data, potentially reducing their effectiveness on new tasks or datasets, such as in Few-Shot Class Incremental Learning (FSCIL). To address the overfitting issue, we adopted the Regularized Lottery Ticket Hypothesis (RLTH)~\cite{kang2022soft}. This led to the discovery of regularized subnetworks characterized by smoother (soft) masks referred to as Soft-SubNetwork (SoftNet), enhancing their adaptability and performance. 

These conventional architecture-based methods, i.e., WSN and SoftNet, offer solutions to prevent forgetting or to alleviate overfitting. However, they are unsuited for sequential complex Video Incremental Learning (VIL) as they reuse a few or all adaptive parameters without finely discretized operations. To enhance neural representation incrementally on complex sequential videos, we propose a novel sequential video compilation method to identify and utilize Lottery tickets (i.e., the weights of complex signals) in frequency space. To achieve this, we define Fourier Subneural Operator (FSO), which breaks down a neural implicit representation into its sine and cosine components (real and imaginary parts) and then selectively identifies the most effective \emph{Lottery tickets} for representing complex periodic signals. Given a backbone and FSO architecture, our method continuously learns to identify input-adaptive sub-modules in Fourier space and encode videos in each sequential training session. We apply FSO to various architectures accompanied by continual learning scenarios, such as Task Incremental Learning (TIL) and Task-agnostic Incremental Learning (TaIL), Video Incremental Learning (VIL), and Few-shot Class Incremental Learning (FSCIL), to demonstrate the effectiveness of FSO representations and compensate abstracted ones.

\noindent
Our contributions can be summarized as follows:
\begin{itemize}[leftmargin=*]

    \item We introduce Fourier Subneural Operator (FSO), which breaks down a neural implicit representation into its sine and cosine components (real and imaginary parts) and then selectively, identifies the most effective \emph{Lottery tickets} for representing complex periodic signals such as sequential video compilation.

    \item We have applied the FSO to various architectures used by a variety of continual learning scenarios: Video Incremental Learning (VIL), Task Incremental Learning (TIL), Task-agnostic Incremental Learning (TaIL), and Few-Shot Class Incremental Learning (FSCIL). In our evaluations, the proposed FSO performs better than architecture-based continual learning models, such as WSN and SoftNet, in TIL, TaIL, VIL, and FSCIL, respectively, underscoring its exceptional representational power.
    
\end{itemize}

%% file: materials/3_concept.tex
\begin{figure*}[ht]
    \centering
    \setlength{\tabcolsep}{-0pt}{%
    \begin{tabular}{cccc}
    \hspace{0.0in}
    \includegraphics[height=3.7cm, trim={0.1cm 0.02cm 0.08cm 0.08cm},clip]{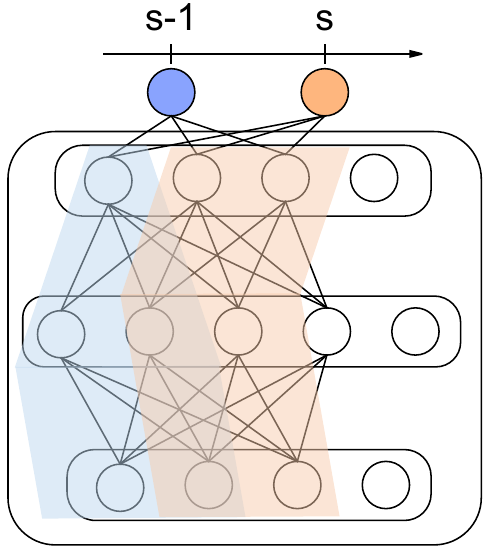} &\;\;\;\;\;\;\;\;\;\;\;
    \includegraphics[height=3.7cm, trim={0.1cm 0.02cm 0.1cm 0.08cm},clip]{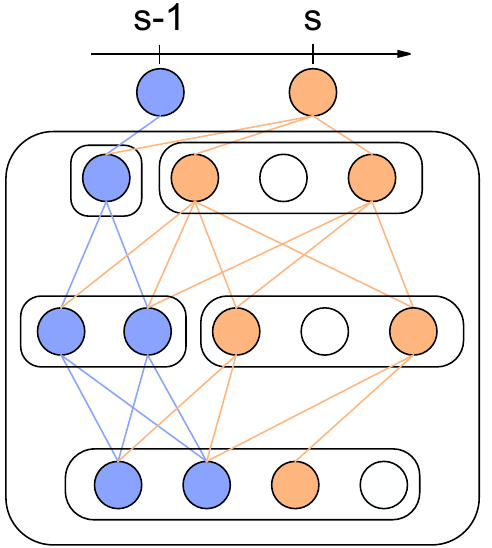} &\;\;\;\;
    \includegraphics[height=3.7cm, trim={0.1cm 0.02cm 0.1cm 0.08cm},clip]{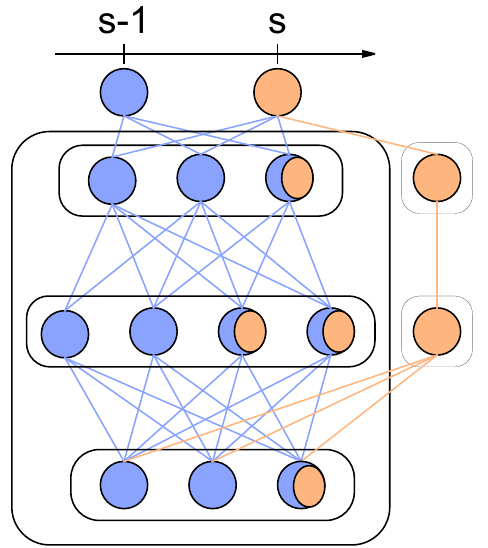} &\;
    \includegraphics[height=3.7cm, trim={0.1cm 0.02cm 0.1cm 0.08cm},clip]{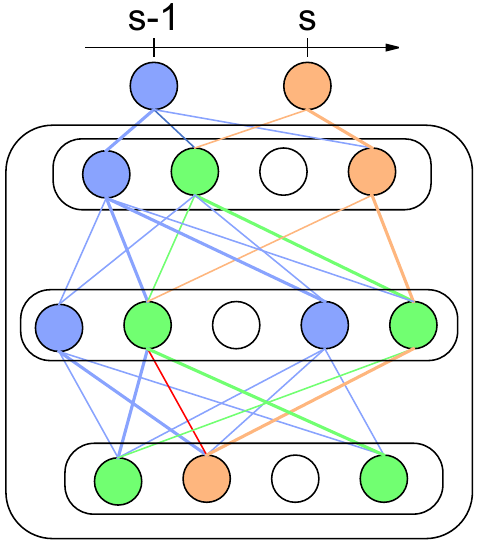} \\ 
    
    \hspace{0.0in}
    \makecell{\small (a) Fixed Backbone \\ \small (Piggyback, SupSup) } &\hspace{0.2in}
    \makecell{\small (b) Biased Transfer \\ \small (PackNet, CLNP)} &\hspace{0.2in}
    \makecell{\small (c) Selective Reuse Expansion \\ \small beyond Dense Network (APD)} &\hspace{0.2in}
    \makecell{\small (d) Selective Reuse Expansion \\ \small within Network (WSN)} \\
    \end{tabular}
    }
    \vspace{-0.05in}
    \caption{\small \textbf{Concept Comparison}: (a) Piggyback \cite{mallya2018piggyback}, and SupSup \cite{wortsman2020supermasks} find the optimal binary mask on a fixed backbone network a given task (b) PackNet \cite{mallya2018packnet} and CLNP \cite{golkar2019continual} forces the model to reuse all features and weights from previous subnetworks which cause bias in the transfer of knowledge (c) APD \cite{Yoon2020} selectively reuse and dynamically expand the dense network (d) Our WSN selectively reuse and dynamically expand subnetworks within a dense network. \textbf{Green edges} are reused weights.}
    \label{fig:concept_comparison}
    \vspace{-0.12in}
\end{figure*}

%% file: 3_related_work.tex
\section{Related Works}

\noindent
\textbf{Continual Learning}~\cite{McCloskey1989, ThrunS1995, KumarA2012icml, LiZ2016eccv}, also known as lifelong learning, is the challenge of learning a sequence of tasks continuously while utilizing and preserving previously learned knowledge to improve performance on new tasks. Four major approaches have been proposed to tackle the challenges of continual learning, such as catastrophic forgetting. One such approach is \textit{regularization-based methods}~\cite{Kirkpatrick2017,chaudhry2020continual, Jung2020, titsias2019functional, mirzadeh2020linear}, which aim to reduce catastrophic forgetting by imposing regularization constraints that inhibit changes to the weights or nodes associated with past tasks. \textit{Rehearsal-based approaches}~\cite{rebuffi2017icarl, chaudhry2018efficient, chaudhry2019continual, Saha2021, deng2021flattening, sun2023decoupling, sarfraz2023error, mai2021supervised, lin2023pcr, aljundi2019online, caccia2021new, caccia2021new, chaudhry2019tiny, liang2024loss, buzzega2020dark} store small data summaries to the past tasks and replay them during training to retain the acquired knowledge. Some methods in this line of work~\cite {ShinH2017nips, aljundi2019online} accommodate the generative model to construct the pseudo-rehearsals for previous tasks. \textit{Architecture-based approaches}~\cite{mallya2018piggyback, Serra2018, li2019learn, wortsman2020supermasks, kang2022forget, kang2022soft} use the additional capacity to expand~\cite{xu2018reinforced, YoonJ2018iclr}, dynamic representation~\cite{yan2021dynamically, singh2020calibrating} or isolate~\cite{rusu2016progressive} model parameters, preserving learned knowledge and preventing forgetting. Both rehearsal and architecture-based methods have shown remarkable efficacy in suppressing catastrophic forgetting but require additional capacity for the task-adaptive parameters~\cite{wortsman2020supermasks} or the replay buffers. Recently, \textit{Prompt-based learning}, an emerging transfer learning technique in natural language processing (NLP), harnesses a fixed function of pre-trained Transformer models. This empowers the language model to receive additional instructions for enhancing its performance on downstream tasks. Notably, while L2P~\cite{wang2022learning} stands out as the seminal work that bridges the gap between prompting and continual learning, DualPrompt~\cite{wang2022dualprompt} introduces an innovative approach to affixing complementary prompts to the pre-trained backbone, thereby enabling the acquisition of both task-invariant and task-specific instructions. Additionally, other notable contributions in this field encompass DyTox~\cite{douillard2022dytox}, S-Prompt~\cite{wang2022s}, CODA-P~\cite{Smith_2023_CVPR}, ConStruct-VL~\cite{Smith_2023_CVPR_b}, ST-Prompt~\cite{Pei_2023_ICCV}, LGCL~\cite{khan2023introducing}, PGP~\cite{qiao2024prompt}. All previous learning methods depend mainly on continual representations in real space. However, the central focus of this study is to pinpoint the most optimal winning ticket representations for convolutional operators in four continual learning scenarios in Fourier space.




\noindent
\textbf{Architecture-based Continual Learning.} 
Artificial network architecture is designed to enable the training of deeper networks. ResNets~\cite{he2016deep}, as a fundamental backbone network with a convolutional operator that shares parameters to obtain image representations in latent spaces, have been widely used in various research fields, such as image classification~\cite{tan2019efficientnet, dosovitskiy2020image}, object detection~\cite{wang2021scaled, liu2021swin}, semantic segmentation~\cite{he2017mask, yuan2020object}, image captioning~\cite{vinyals2015show, cho2020x}, image generation~\cite{ho2020denoising, blattmann2023align}, and architecture-based continual learning~\cite{golkar2019continual, mallya2018piggyback, Serra2018, Chen2021lifelonglottery}. A recent CL method, LL-Tickets~\cite{Chen2021lifelonglottery}, shows a sparse subnetwork called lifelong tickets that performs well on all tasks during continual learning. However, LL-Tickets require external data to maximize knowledge distillation with learned models for prior tasks, and the ticket expansion process involves retraining and pruning steps. WSN~\cite{kang2022forget} was an improved method that jointly learns the model and task-adaptive subnetwork associated with each task in Task Incremental Learning (TIL). Also, in Few-Shot Class Incremental Learning (FSCIL), Soft-Subnetworks (SoftNet)~\cite{kang2022soft} was proposed as another variant of WSN, consisting of major sub-networks (winning tickets) to obtain base session knowledge and minor sub-networks to alleviate overfitting few samples for new sessions. However, these methods adhere to the traditional ResNet architecture. As the layers increase in depth, the image representations become more abstract, which can result in the loss of the global structure of the input images. These drawbacks can negatively impact the effectiveness of image representation and the continuity of the representational power. To overcome these issues, we introduce a new convolutional operator referred to as the Fourier Subneural Operator (FSO), which transfers the global image representation obtained from lower layers to higher layers through Residual Blocks. This approach helps maintain the global image representation in Artificial Neural Networks at higher layers.


%
\noindent 
\textbf{Neural Implicit Representation (NIR)}~\cite{mehta2021modulated} are neural network architectures for parameterizing continuous, differentiable signals. Based on coordinate information, they provide a way to represent complex, high-dimensional data with a small set of learnable parameters that can be used for various tasks such as image reconstruction~\cite{sitzmann2020implicit, tancik2020fourier}, shape regression~\cite{chen2019learning, park2019deepsdf}, and 3D view synthesis ~\cite{mildenhall2021nerf, schwarz2020graf}. Instead of using coordinate-based methods, NeRV~\cite{chen2021nerv} proposes an image-wise implicit representation that takes frame indices as inputs, enabling faster and more accurate video compression than coordinate methods. NeRV has inspired further improvements in video regression by CNeRV~\cite{chen2022cnerv}, DNeRV~\cite{he2023towards}, E-NeRV~\cite{li2022nerv}, and NIRVANA~\cite{maiya2022nirvana}, and HNeRV~\cite{chen2023hnerv}. A few recent works have explored video continual learning (VCL) scenarios for the NIR. To tackle non-physical environments, Continual Predictive Learning (CPL)~\cite{chen2022continual} learns a mixture world model via predictive experience replay and performs test-time adaptation using non-parametric task inference. PIVOT~\cite{villa2022pivot} leverages the past knowledge present in pre-trained models from the image domain to reduce the number of trainable parameters and mitigate forgetting. CPL needs memory to replay, while PIVOT needs pre-training and fine-tuning steps. In contrast, along with the conventional progressive training techniques~\cite{rusu2016progressive, cho2022streamable}, considering the advantages of forget-free convergence speed, we set WSN as baselines, which utilizes the Lottery Ticket Hypothesis (LTH) to identify an adaptive substructure within the dense networks that are tailored to the specific video input index. However, WSN is inappropriate for sequential complex video compilation since it reuses a few adaptive but sparse learnable parameters. To overcome the weakness of WSN, We proposed a novel Fourier Subneural Operator (FSO)~\cite{kang2024progressive} for representing complex video in Fourier space \cite{li2020fourier, li2020neural, kovachki2021neural, tran2021factorized}. We have expanded the FSO of Fourier representations to encompass a variety of continual learning architectures and scenarios to validate its effectivness.

%% file: 4_approach.tex
\section{Winning SubNetworks in Fourier Space}
This section presents our pruning-based continual learning methods, \textit{Winning SubNetworks} (WSN, see \Cref{fig:wsn})~\cite{kang2022forget} and introduces the Fourier Subneural Operator (FSO, see \Cref{fig:concept_CVRNet} and \Cref{fig:residual_fso}) for better video representations. Then, we depict how we apply FSO to various architectures used in four continual learning scenarios.

\input{materials/4_concept_wsn}

\subsection{WSN \& Fourier Subneural Operator (FSO)}
The WSN searches for the task-adaptive winning tickets and updates only the weights not trained on the previous tasks, as shown in \Cref{fig:wsn}. After training on each session, the subnetwork parameters of the model are frozen to ensure that the proposed method is inherently immune to catastrophic forgetting. Moreover, we illustrate \textit{Soft-Winning SubNetworks} (SoftNet)~\cite{kang2022soft}, proposed to address the issues of forgetting previous sessions and overfitting a few samples of new sessions. These conventional architecture-based methods, i.e., WSN and SoftNet, offer solutions to prevent forgetting. However, they are unsuited for sequential complex Video Incremental Learning (VIL see \Cref{fig:concept_CVRNet}) as they reuse a few or all adaptive parameters without finely discretized operations. To enhance neural representation incrementally on complex sequential videos, we introduce Fourier Subneural Operator (FSO), which breaks down a neural implicit representation into its sine and cosine components (real and imaginary parts) and then selectively identifies the most effective \emph{Lottery tickets} for representing complex periodic signals. In practice, given a backbone and FSO architecture, our method continuously learns to identify input-adaptive subnetwork modules and encode each new video into the corresponding module during sequential training sessions. We extend Fourier representations to various continual learning scenarios, such as TIL, TaIL, VIL, and FSCIL, to demonstrate its effectiveness.

\noindent 
\textbf{Problem Statement.} Consider a supervised learning setup where $S$ sessions or tasks arrive to a learner sequentially. We denote that $\mathcal{D}_s=\{\bm{x}_{i,s}, y_{i,s}\}_{i=1}^{n_s}$ is the dataset of session $s$, composed of $n_s$ pairs of raw instances and corresponding labels. We assume a neural network $f(\cdot;\bm{\theta})$, parameterized by the model weights $\bm{\theta}$ and standard continual learning scenario aims to learn a sequence of sessions by solving the following optimization procedure at each step $s$: 
\begin{equation}
\bm{\theta}^{\ast}=\minimize_{\bm{\theta}} \frac{1}{n_s}\sum^{n_s}_{i=1}\mathcal{L}(f(\bm{x}_{i,s};\bm{\theta}), y_{i,s}),
\label{eq:task_loss}
\end{equation}
where $\mathcal{L}(\cdot, \cdot)$ is a classification objective loss such as cross-entropy loss. $\mathcal{D}_s$ for session $s$ is only accessible when learning session $s$.

Continual learners frequently use over-parameterized deep neural networks (dense network) to ensure enough capacity for learning future tasks. This approach often leads to the discovery of subnetworks that perform as well as, or better than, the dense network. Given the neural network parameters $\bm{\theta}$, the binary mask $\bm{m}^*_s$ that describes the optimal subnetwork for session $s$ such that $|\bm{m}^*_s|$ is less than the model capacity $c$, is defined as:
\begin{equation}
\begin{split}
    \bm{m}^*_s &= \underset{\bm{m}_s\in\{0,1\}^{|\bm{\theta}|}}{\minimize} \frac{1}{n_s}\sum^{n_s}_{i=1}\mathcal{L}\big(f(\bm{x}_{i,s};\bm{\theta} \odot \bm{m}_s), y_{i,s}\big) - \mathcal{J}  \\
    &\quad\quad\quad\quad\quad\quad \text{subject to~}|\bm{m}^*_s|\leq c \ll |\bm{\theta}|,
\end{split}
\label{eq:subnetwork}
\end{equation}
where the session loss $\mathcal{J}=\mathcal{L}\big(f(\bm{x}_{i,s};\bm{\theta}), y_{i,s}\big)$ and the total number of parameters in the dense network is $|\bm{\theta}|$, and $c = \frac{|\bm{m}^*_s|}{|\bm{\theta}|} \times 1e2$ is used as the selected proportion ($\%$) of model parameters in the following sections. In the optimization section, we describe how to obtain $\bm{m}^*_s$ using a single learnable weight score $\bm{\rho}$ that is subject to updates while minimizing session loss jointly for each task or session.

\subsection{Fourier Subnueral Operator (FSO)} 
Conventional continual learner (i.e., WSN) only uses a few learnable parameters in convolutional operations to represent complex sequential image streams in Video Incremental Learning. To capture more parameter-efficient and forget-free video representations (i.e., Neural Implicit Representation (NIR), see \Cref{fig:concept_CVRNet}), the NIR model requires fine discretization and video-specific sub-parameters. This motivation leads us to propose a novel subnetwork operator in Fourier space, which provides it with various bandwidths. Following the previous definition of Fourier convolutional operator~\cite{li2020fourier}, we adapt and redefine this definition to better fit the needs of the NIR framework. We use the symbol $\mathcal{F}$ to represent the Fourier transform of a function $f$, which maps from an embedding space of dimension $d_{\bm{e}}=1 \times 160$ to a frame size denoted as $d_{\bm{v}}$. The inverse of this transformation is represented by $\mathcal{F}^{-1}$. In this context, we introduce our Fourier-integral Subneural Operator (FSO), symbolized as $\mathcal{K}$, which is tailored to enhance the capabilities of our NIR system:
\begin{equation}
\left( \mathcal{K}(\phi) \tilde{\bm{v}}_t^s \right)(\bm{e}_{s,t}) = \mathcal{F}^{-1}(R_{\bm \phi} \cdot (\mathcal{F} \tilde{\bm{v}}_t^s))(\bm{e}_{s,t}),
\label{eq:FSOper}
\end{equation}
where $\tilde{\bm{v}}_t^s$ is a hidden representation; $R_{\bm \phi}$ is the Fourier transform of a periodic subnetwork function which is parameterized by its subnetwork's parameters of real $(\bm{\theta}^{real} \odot \bm{m}_s^{real})$ and imaginary $(\bm{\theta}^{imag} \odot \bm{m}_s^{imag})$. We thus parameterize $R_{\bm \phi}$ separately as complex-valued tensors of real and imaginary $\bm{\phi}_{FSO} \in \{ \bm{\theta}^{real}, \bm{\theta}^{imag} \}$. One key aspect of the FSO is that its parameters grow with the depth of the layer and the input/output size. However, through careful layer-wise inspection and adjustments for sparsity, we can find a balance that allows the FSO to describe neural implicit representations efficiently. In the experimental section, we will showcase the most efficient FSO structure and its performance. \Cref{fig:concept_CVRNet} shows one possible structure of a single FSO. We describe the optimization in the following section. 

\input{FSO-materials/outline}

For Task/Class Incremental Learnings, Convolutional Neural Networks (CNNs) take a convolutional operation, followed by a pooling operation. These iterative operations of CNNs represent more abstract features at higher layer output levels and lose global contextual representations. To compensate for low-contextual representations, we add an FSO to CNN architecture as shown in \Cref{fig:residual_fso}. The lower layer's output $\bm{x}^{l-1}$ is merged into the $l$th layer residual block through the FSO to acquire spatially ensembling features $\bm{x}^{l}$. The FSO also provides additional parameters to push the residual to zero. We show the differences between ensembling features (WSN+FSO) and single features (WSN) represented by residual blocks as shown in \Cref{fig:var_vs_freq}: WSN+FSO provides lower variances of feature maps and higher frequency components than WSN. In the following experimental settings, we investigate various CNN architectures with FSO.  

\input{FSO-materials/residual_fso}

\input{FSO-materials/plot_main_var_freq}

\subsection{SubNetworks with FSO}\label{sub_sec:sub_fso}

\subsubsection{Fourier Subneural Operator (FSO)}

\subsubsection{Winning SubNetworks (WSN) with FSO}\label{sub_sec:wsn}
Let each weight $\bm{\theta}_\ast=\{\bm{\theta}, \bm{\phi}_{FSO}\}$ be associated with a learnable parameter we call \textit{weight score} $\bm{\rho}_\ast=\{\bm{\rho}, \bm{\rho}_{FSO}\}$, which numerically determines the importance of the weight associated with it; that is, a weight with a higher weight score is seen as more important. We find a sparse subnetwork $\hat{\bm{\theta}}_s$ of the neural network and assign it as a solver of the current session $s$. We use subnetworks instead of the dense network as solvers for two reasons: (1) Lottery Ticket Hypothesis \cite{frankle2018lottery} shows the existence of a subnetwork that performs as well as the whole network, and (2) subnetwork requires less capacity than dense networks, and therefore it inherently reduces the size of the expansion of the solver.

Motivated by such benefits, we propose a novel \textit{Winning SubNetworks} (WSN\footnote{WSN code is available at \url{https://github.com/ihaeyong/WSN.git}}), which is the joint-training method for continual learning that trains on session - while selecting an important subnetwork given the session $s$ as shown in Fig. \ref{fig:wsn}. The illustration of WSN explains step-by-step how to acquire binary weights within a dense network. We find $\hat{\bm\theta}_t$ by selecting the $c$\% weights with the highest weight scores $\bm{\rho}_\ast$, where $c$ is the target layerwise capacity ratio in \%. A task-dependent binary weight represents the selection of weights $\bm{m}_s$ where a value of $1$ denotes that the weight is selected during the forward pass and $0$ otherwise. Formally, $\mathbf{m}_s$ is obtained by applying a indicator function $\mathbbm{1}_c$ on $\bm{\rho}$ where $\mathbbm{1}_c(\rho)=1$ if $\rho_\ast$ belongs to top-$c\%$ scores and $0$ otherwise. Therefore, the subnetwork $\hat{\bm{\theta}}_s$ for session $s$ is obtained by $\hat{\bm{\theta}}_s = \bm{\theta}_\ast \odot \mathbf{m}_s$.  

\subsubsection{Soft-Subnetworks (SoftNet) with FSO}
Several works have addressed overfitting issues in continual learning from different perspectives, including NCM~\cite{hou2019learning}, BiC~\cite{wu2019large}, OCS~\cite{yoon2022online}, and FSLL~\cite{mazumder2021few}. To mitigate the overfitting issue in subnetworks, we use a simple yet efficient method named \emph{SoftNet} proposed by \cite{kang2022soft}. The following new paradigm, referred to as \emph{Regularized Lottery Ticket Hypothesis}~\cite{kang2022soft} which is inspired by the \emph{Lottery Ticket Hypothesis}~\cite{frankle2018lottery} has become the cornerstone of SoftNet: 

\noindent
\textbf{Regularized Lottery Ticket Hypothesis (RLTH).} \textit{A randomly-initialized dense neural network contains a regularized subnetwork that can retain the prior class knowledge while providing room to learn the new class knowledge through isolated training of the subnetwork.}

\noindent
Based on RLTH, we propose a method, referred to as Soft-SubNetworks (SoftNet\footnote{SoftNet code is available at \url{https://github.com/ihaeyong/SoftNet-FSCIL.git}}). SoftNet jointly learns the randomly initialized dense model, and soft mask $\bm{m} \in [0, 1]^{|\bm{\theta}_\ast|}$ on Soft-subnetwork on each session training; the soft mask consists of the major part of the model parameters $m = 1$ and the minor ones $m < 1$ where $m = 1$ is obtained by the top-$c\%$ of model parameters and $m < 1$ is obtained by the remaining ones ($100 - \text{top-}c\%$) sampled from the uniform distribution $U(0, 1)$. Here, it is critical to select minor parameters $m < 1$ in a given dense network. 

\subsection{Optimization for TIL, TaIL, VIL, and FSCIL}\label{sub_sec:opt_wsn}
\subsubsection{Winning SubNetworks (WSN) for TIL and TaIL}\label{sub_sec:opt_til}
To jointly learn the model weights and task-adaptive binary masks of subnetworks associated with each session, given an objective $\mathcal{L}(\cdot)$, i.e., cross-entropy loss, we optimize $\bm{\theta}_\ast$ and $\bm{\rho}_\ast$ with:
\begin{equation}
\label{eq:optimizer}
    \minimize_{\bm{\theta}_\ast, \bm{\rho}_\ast} \mathcal{L} (\bm{\theta}_\ast \odot \bm{m}_s; \mathcal{D}_s).
\end{equation}
However, this vanilla optimization procedure presents two problems: (1) updating all $\bm{\theta}_\ast$ when training for new sessions will cause interference to the weights allocated for previous sessions, and (2) the indicator function always has a gradient value of $0$; therefore, updating the weight scores $\bm{\rho}_\ast$ with its loss gradient is not possible. To solve the first problem, we selectively update the weights by allowing updates only on the weights not selected in the previous sessions. To do that, we use an \textit{accumulate binary mask} $\bm{M}_{s-1}=\logicalor_{i=1}^{s-1} \bm{m}_i$ when learning session $s$, then for an optimizer with learning rate $\eta,$ the $\bm{\theta}_\ast$ is updated as follows: 
\begin{equation}
\label{eq:param_update}
\bm{\theta}_\ast \leftarrow \bm{\theta}_\ast - \eta \left(\frac{\partial \mathcal{L}}{\partial \bm{\theta}_\ast} \odot (\bm{1}-\bm{M}_{s-1})\right),
\end{equation} 
effectively freezing the weights of the subnetworks selected for the previous sessions. To solve the second problem, we use Straight-through Estimator \cite{Hinton2012, Bengio2013, Ramanujan2020} in the backward pass since $\bm{m}_s$ is obtained by top-$c\%$ scores. Specifically, we ignore the derivatives of the indicator function and update the weight score as follows: 
\begin{equation}
\label{eq:ste}
\quad \bm{\rho}_\ast \leftarrow \bm{\rho}_\ast - \eta\left(\frac{\partial \mathcal{L}}{\partial \bm{\rho}_\ast}\right).
\end{equation}
Our WSN optimizing procedure is summarized in \Cref{alg:algorithm1}.

\input{materials/4_algorithm}

At the inference time of TaIL, we infer task identity for arbitrary pieces of task samples. To infer the task identity, we follow SupSup's one-shot task-inference method described in \cite{wortsman2020supermasks}. In short, we assign each learned subnetwork $\bm{m}_s$ a weight $\alpha_s$ such that $\sum_s \alpha_s = 1$ and $\alpha_s \geq 0.$ Given an example data point of batch $\bm{x} \in \bm{b}$ to classify, we can compute our loss as $\mathcal{L}=\mathcal{H}(f(\bm{x}; \bm{\theta} \odot (\sum_s \alpha_s \bm{m}_s)))$ where$f(\bm{x};\bm{\theta})$ is our neural network which outputs logits and $\mathcal{H}$ is our entropy function. From here our inferred task is simply $\hat{s} = \mathrm{argmin}_s \frac{\partial \mathcal{H}}{\partial \alpha_s}.$ High entropy prediction distributions are very uncertain (close to uniform), and the lowest entropy is reached when our distribution is very certain (at a one-hot vector). As recorded in the SupSup report, the task inference performance of WSN+FSO shows $100\%$ accuracy for all tasks.

\subsubsection{Winning SubNetworks (WSN) for VIL}\label{sub_sec:opt_vil}
Let a video at $s_{th}$ session $\bm{V}_s =\{\bm{v}^s_t\}^{T_s}_{t=1} \in \mathbb{R}^{T_s \times H \times W \times 3}$ be represented by a function with the trainable parameter $\bm{\theta}_\ast$, $f_{\bm{\theta}_\ast}: \mathbb{R} \rightarrow \mathbb{R}^{H \times W \times 3}$, during Video Incremental Learning (VIL), where $T_s$ denotes the number of frames in a video at session $s$, and $s \in \{1 \dots, |S|\}$. Given a session and frame index $s$ and $t$, respectively, the neural implicit representation aims to predict a corresponding RGB image $\bm{v}^s_t \in \mathbb{R}^{H \times W \times 3}$ by fitting an encoding function to a neural network: $\bm{v}^s_t = f_{\bm{\theta}_\ast}([s;t], H_s)$ where $H_s$ is $s_{th}$ head. For the sake of simplicity, we omit $H_s$ in the following equations. Let's consider a real-world learning scenario in which $|\mathcal{S}|=N$ or more sessions arrive in the model sequentially. We denote that $\mathcal{D}_s=\{\bm{e}_{s,t}, \bm{v}_{s,t}\}_{t=1}^{T_s}$ is the dataset of session $s$, composed of $T_s$ pairs of raw embeddings $\bm{e}_{s,t} = \left[\bm{e}_s; \bm{e}_t\right] \in \mathbb{R}^{1 \times 160}$ and corresponding frames $\bm{v}^s_t$. Here, we assume that $\mathcal{D}_s$ for session $s$ is only accessible when learning session $s$ due to the limited hardware memory and privacy-preserving issues, and session identity is given in the training and testing stages. The primary training objective in this sequence of $N$ video sessions is to minimize the following optimization problem:
\begin{equation}
\minimize_{\bm{\theta}_\ast, \bm{\rho}_\ast} \frac{1}{N} \frac{1}{T_s}\sum^{N}_{s=1}\sum^{T_s}_{t=1}\mathcal{L}(f(\bm{e}_{s,t};\bm{\theta}_\ast \odot \bm{m}_s), \bm{v}^s_t),
\label{eq:sess_loss}
\end{equation}
where the loss function $\mathcal{L}(\bm{v}^s_t)$ is composed of $\ell_1$ loss and \textit{SSIM loss}. The former minimizes the pixel-wise RGB gap with the original input frames evenly, and the latter maximizes the similarity between the two entire frames based on luminance, contrast, and structure, as follows: 
\begin{equation}
\mathcal{L}(\bm{V}_s) = {1\over T_s} \sum_{t=1}^{T_s} \alpha ||\bm{v}^s_t-\hat{\bm{v}}^s_t||_1 + (1-\alpha)(1-\text{SSIM}(\bm{v}^s_t, \hat{\bm{v}}^s_t)),
\label{eq:l1_loss}
\end{equation} 
where $\hat{\bm{v}}^s_t$ is the output generated by the model $f$. For all experiments, we set the hyperparameter $\alpha$ to $0.7$, and we adapt PixelShuffle~\cite{shi2016real} for session and time positional embedding. 

\subsubsection{Soft-SubNetworks (SoftNet) for FSCIL}\label{sub_sec:opt_fscil}
Similar to WSN's optimization discussed in \Cref{sub_sec:opt_wsn}, let each weight $\bm{\theta}_\ast$ be associated with a learnable parameter we call \textit{weight score} $\bm{\rho}_\ast$, which numerically determines the importance of the associated weight. In the optimization process for FSCIL, however, we consider two main problems: (1) Catastrophic forgetting: updating all $\bm{\theta}_\ast \odot \bm{m}_{s-1}$ when training for new sessions will cause interference with the weights allocated for previous sessions; thus, we need to freeze all previously learned parameters $\bm{\theta}_\ast \odot \bm{m}_{s-1}$; (2) Overfitting: the subnetwork also encounters overfitting issues when training an incremental session on a few samples, as such, we need to update a few parameters irrelevant to previous session knowledge., i.e., $\bm{\theta}_\ast \odot (\bm{1}-\bm{m}_{s-1})$.  

To acquire the optimal subnetworks that alleviate the two issues, we define a soft-subnetwork by dividing the dense neural network into two parts-one is the major subnetwork $\bm{m}_\text{major}$, and another is the minor subnetwork $\bm{m}_\text{minor}$. The defined Soft-SubNetwork (SoftNet) follows as:
\begin{equation}
    \bm{m}_\text{soft} = \bm{m}_\text{major} \oplus \bm{m}_\text{minor},
    \label{eq:soft_mask}
\end{equation}
where $\bm{m}_\text{major}$ is a binary mask and $\bm{m}_\text{minor} \sim U(0,1)$ and $\oplus$ represents an element-wise summation. As such, a soft-mask is given as $\bm{m}_s^\ast \in [0,1]^{|\bm{\theta}_\ast|}$. In the all-experimental FSCIL setting, $\bm{m}_\text{major}$ maintains the base session knowledge $s=1$ while $\bm{m}_\text{minor}$ acquires the novel session knowledge $s \geq 2$. Then, with base session learning rate $\alpha,$ the $\bm{\theta}_\ast$ is updated as follows: $\bm{\theta}_\ast \leftarrow \bm{\theta}_\ast - \alpha \left(\frac{\partial \mathcal{L}}{\partial \bm{\theta}_\ast} \odot \bm{m}_\text{soft}\right)$ effectively regularize the weights of the subnetworks for incremental learning. Our Soft-subnetwork optimizing procedure is summarized in \Cref{alg:algorithm2}. Once a single soft-subnetwork $\bs{m}_{\text{soft}}$ is obtained in the base session, then we use the soft-subnetwork for the entire new sessions without updating.

\input{soft-materials/5_softnet/4_algorithm}

\noindent 
\textbf{Base Training} $(s = 1)$. In the base learning session, we optimize the soft-subnetwork parameter $\bm{\theta}_\ast$ (including a fully connected layer as a classifier) and weight score $\bm{\rho}_\ast$ with cross-entropy loss jointly using the training data $\mathcal{D}^1$. 

\noindent 
\textbf{Incremental Training} $(s \geq 2)$. In the incremental few-shot learning sessions $(s \geq 2)$, leveraged by $\bm{\theta}_\ast \odot \bm{m}_{\text{soft}}$, we fine-tune few minor parameters $\bm{\theta}_\ast \odot \bm{m}_{\text{minor}}$ of the soft-subnetwork to learn new classes. Since $\bm{m}_{\text{minor}} < \bm{1}$, the soft-subnetwork alleviates the overfitting of a few samples. Furthermore, instead of Euclidean distance \cite{shi2021overcoming}, we employ a metric-based classification algorithm with cosine distance to finetune the few selected parameters. In some cases, Euclidean distance fails to give the real distances between representations, especially when two points with the same distance from prototypes do not fall in the same class. In contrast, representations with a low cosine distance are located in the same direction from the origin, providing a normalized informative measurement. We define the loss function as: 
\begin{equation}
\begin{split}
    \mathcal{L}_m (\bm{x},y; \bm{\theta}_\ast \odot \bm{m}_{soft}) = ~~~~~~~~~~~~~~~~~~~~~~~~~~~~~~~~~~~~~~~~~~~~~~~~~~~~~~~~ & \\ 
    -\sum_{\bm{x},y \in \mathcal{D}} \sum_{o \in \mathcal{O}} \mathbbm{1}(y=o) \log \left( \frac{e^{-d(\bm{p}_o, f(\bm{x};\; \bm{\theta}_\ast \odot \bm{m}_{soft}))}}{\sum_{o_k \in \mathcal{O}} e^{-d(\bm{p}_{o_k}, f(\bm{x};\; \bm{\theta}_\ast \odot \bm{m}_{soft}))}} \right)
\end{split}
\label{eq:proto_loss}
\end{equation}
where $d\left(\cdot, \cdot\right)$ denotes cosine distance, $\bm{p}_o$ is the prototype of class $o$, $\mathcal{O} = \bigcup_{i=1}^s \mathcal{O}^i$ refers to all encountered classes, and $\mathcal{D} = \mathcal{D}^s \bigcup \mathcal{P}$ denotes the union of the current training data $\mathcal{D}^s$ and the exemplar set $\mathcal{P} = \left\{\bm{p}_2 \cdots, \bm{p}_{s-1}\right\}$, where $\mathcal{P}_{s_e} \left(2 \leq s_e < s\right)$ is the set of saved exemplars in session $s_e$. Note that the prototypes of new classes are computed by $\bm{p}_o = \frac{1}{N_o} \sum_i \mathbbm{1}(y_i = o) f(\bm{x}_i; \bm{\theta}_\ast \odot \bm{m}_{soft})$ and those of base classes are saved in the base session, and $N_o$ denotes the number of the training images of class $o$. We also save the prototypes of all classes in $\mathcal{O}^s$ for later evaluation.  \\

\noindent 
\textbf{Inference for Incremental Soft-Subnetwork.} In each session, the inference is also conducted by a simple nearest class mean (NCM) classification algorithm \cite{mensink2013distance, shi2021overcoming} for fair comparisons. Specifically, all the training and test samples are mapped to the embedding space of the feature extractor $f$, and Euclidean distance $d_u(\cdot, \cdot)$ is used to measure their similarity. The classifier gives the $k$th prototype index $o_k^{\ast} = \arg\min_{o \in \mathcal{O}} d_u(f(\bm{x}; \bm{\theta}_\ast \odot \bm{m}_{soft}), \bm{p}_o)$ as output.

%% file: materials/4_concept_wsn.tex
\begin{figure*}[ht]
    \centering
    \setlength{\tabcolsep}{-0pt}{%
    \begin{tabular}{cccc}
    \includegraphics[height=4.4cm, trim={0.2cm 0.1cm 0.1cm 0.3cm},clip]{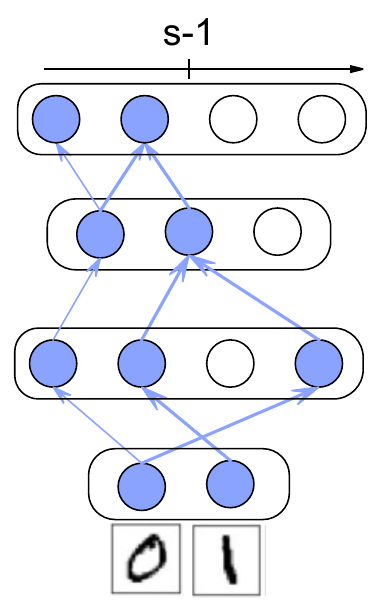} & \;\;\;\;\;\;\;\;\;
    \includegraphics[height=4.4cm, trim={0.2cm 0.1cm 0.1cm 0.3cm},clip]{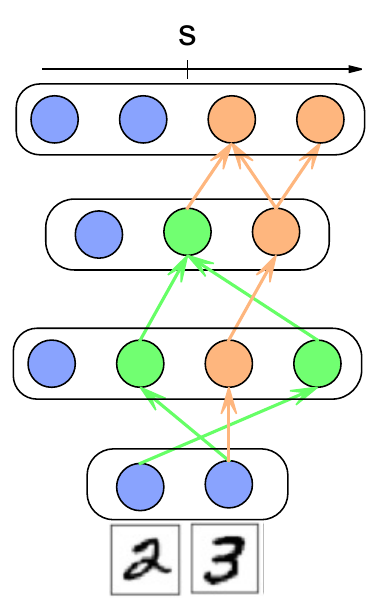} & \;\;\;\;\;\;\;\;\;
    \includegraphics[height=4.4cm, trim={0.2cm 0.1cm 0.1cm 0.3cm},clip]{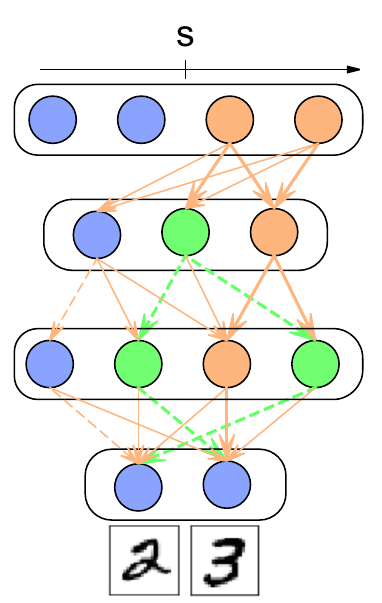} & \;\;\;\;\;\;\;\;\;
    \includegraphics[height=4.4cm, trim={0.2cm 0.1cm 0.1cm 0.3cm},clip]{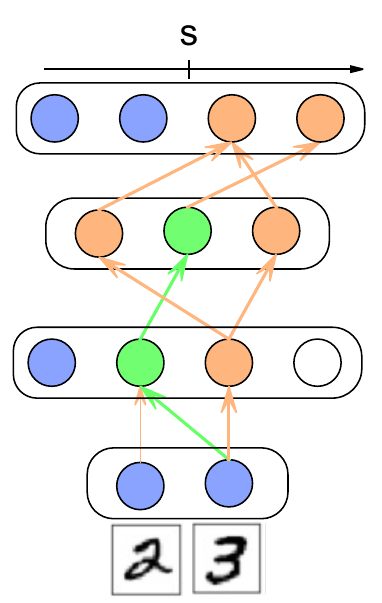} \\ 
    \hspace{0.2in}
    \makecell{(a) selected \textcolor{black}{weights} \\ at prior task} & \makecell{(b) forward pass \\ at current task} & \makecell{(c) backward pass \\ at current task} & \makecell{(d) selected \textcolor{black}{weights} \\ at current task} \\
    \end{tabular}
    }
    \vspace{-0.1in}
    \caption{\small \textbf{An illustration of Winning SubNetworks (WSN)}: (a) The top-c\% \textcolor{black}{weights} $\hat{\bm{\theta}}_{s-1}$ at prior task are obtained, (b) In the forward pass of a new task, WSN selects the top-c\% and \textcolor{black}{reuses weights} selected from prior tasks, (c) In the backward pass, WSN updates only \textcolor{black}{non-used weights}(\sampleline{}) while freezing reused weights(\sampleline{dashed}), and (d) after several iterations of (b) and (c), we acquire again the top-c\% \textcolor{black}{weights} $\hat{\bm{\theta}}_{s}$ including subsets of \textbf{reused weights (green)} for the new task.}
  \label{fig:wsn}
  \vspace{-0.1in}
\end{figure*}

%% file: FSO-materials/outline.tex
\begin{figure*}
    \small
    \centering
    \vspace{-0.12in}
    \setlength{\tabcolsep}{-2pt}{%
    \includegraphics[height=5.8cm, trim={0.1cm 0.1cm 0.1cm 0.1cm},clip]{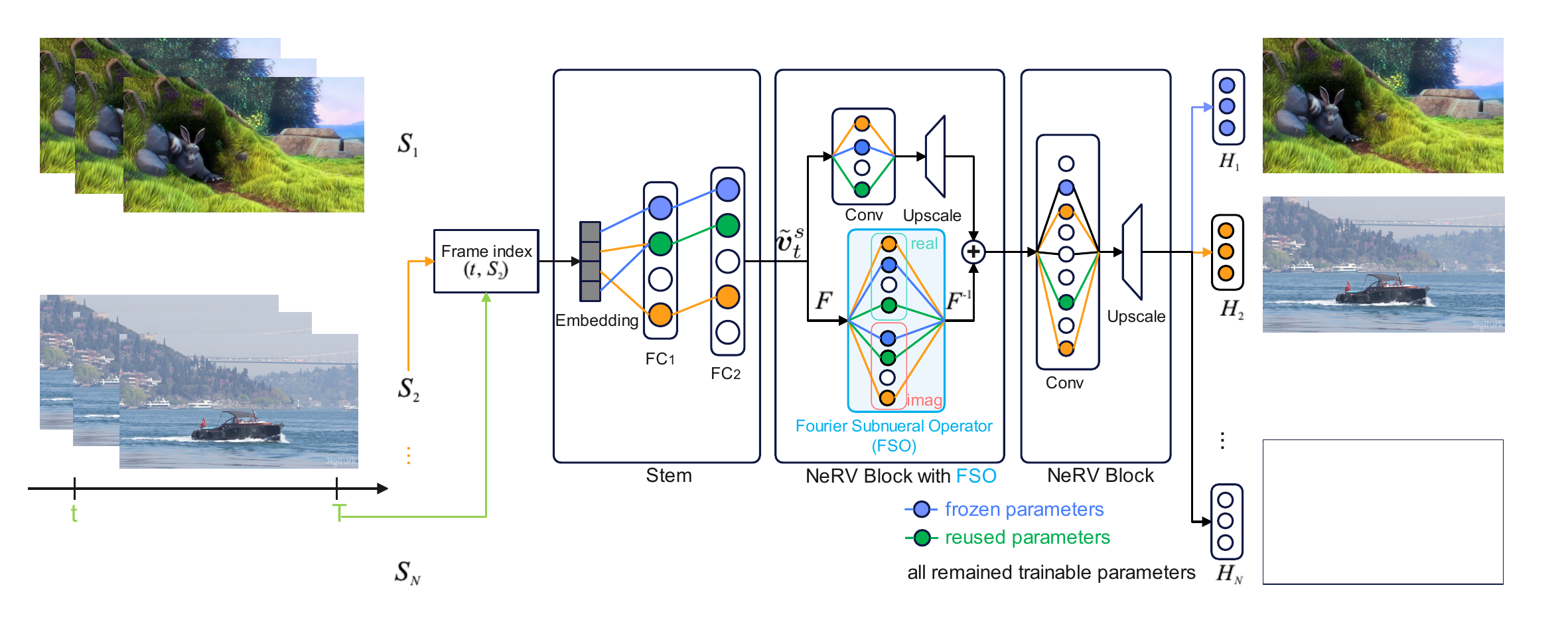}
    }
    \vspace{-0.2in}
    \caption{\small \textbf{Forget-free Neural Implicit Representation with Fourier Subneural Operator (FSO) for Video Incremental Learning}: Image-wise neural implicit representation taking frame and video (session $s$) indices as input and using a sparse Stems + NeRV Blocks with \textit{Fourier Subneural Operator} (FSO) to output the whole image through multi-heads $H_N$ where $\tilde{\bm{v}}_s^t$ is a hidden representation. We denote frozen, reused, and trainable parameters in training at session 2. Note that each video representation is colored. In inference, we only need indices of session $s$ and frame $t$ and session mask (subnetwork).}
    \label{fig:concept_CVRNet}
    \vspace{-0.1in}
\end{figure*}

%% file: FSO-materials/residual_fso.tex
\begin{figure}
    \small
    \centering
    \vspace{-0.12in}
    \setlength{\tabcolsep}{-2pt}{%
    \includegraphics[height=5.9cm, trim={1.8cm 1.8cm 1.8cm 2.0cm},clip]{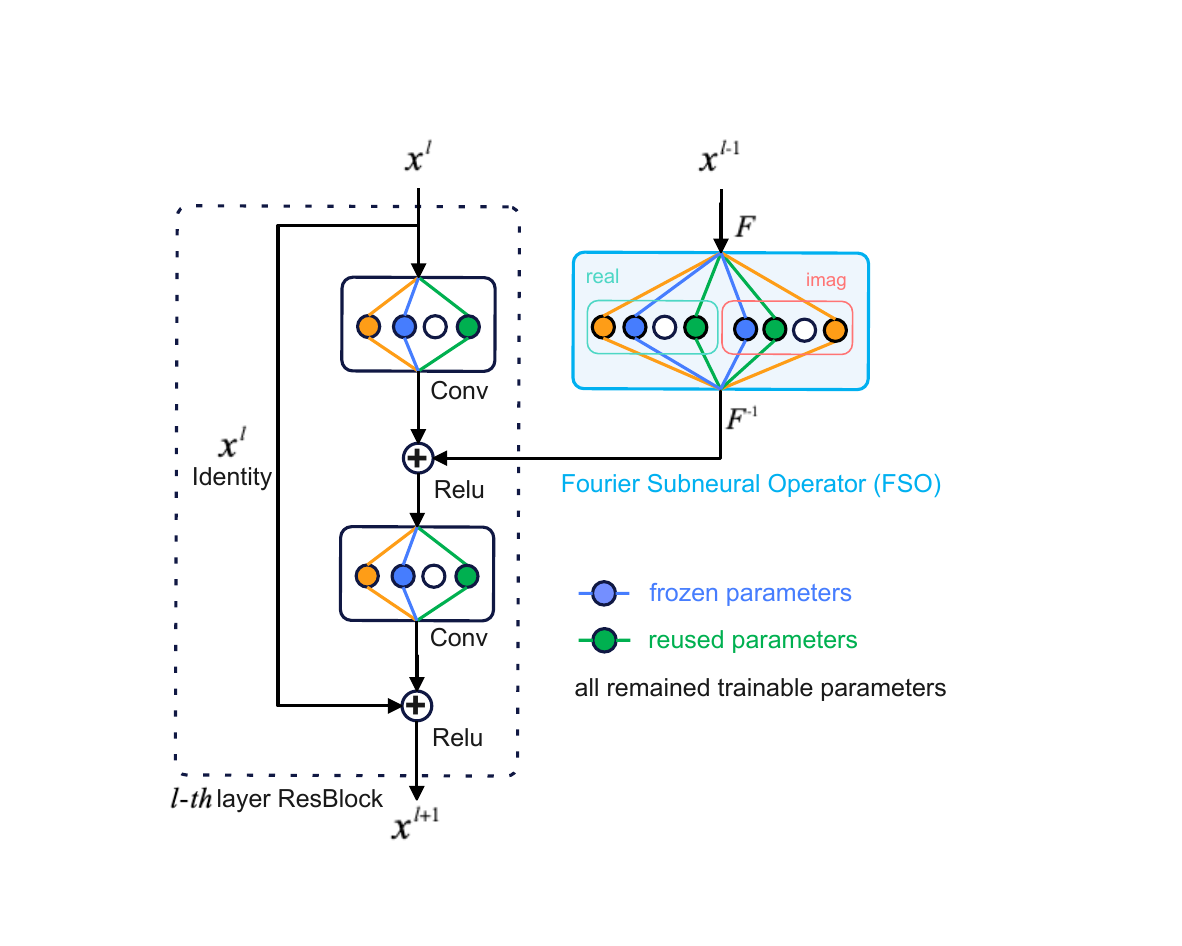}
    }
    \vspace{-0.05in}
    \caption{\small \textbf{Residual Blocks (ResBlocks) with Fourier Subnerual Operator (FSO)}.}
    \label{fig:residual_fso}
    \vspace{-0.1in}
\end{figure}

%% file: FSO-materials/plot_main_var_freq.tex
\begin{figure}[ht]
    \centering
    \vspace{-0.1in}
    \renewcommand{\arraystretch}{1}
    \setlength{\tabcolsep}{0pt}{%
    \begin{tabular}{cc}
    \includegraphics[width=0.47\columnwidth]{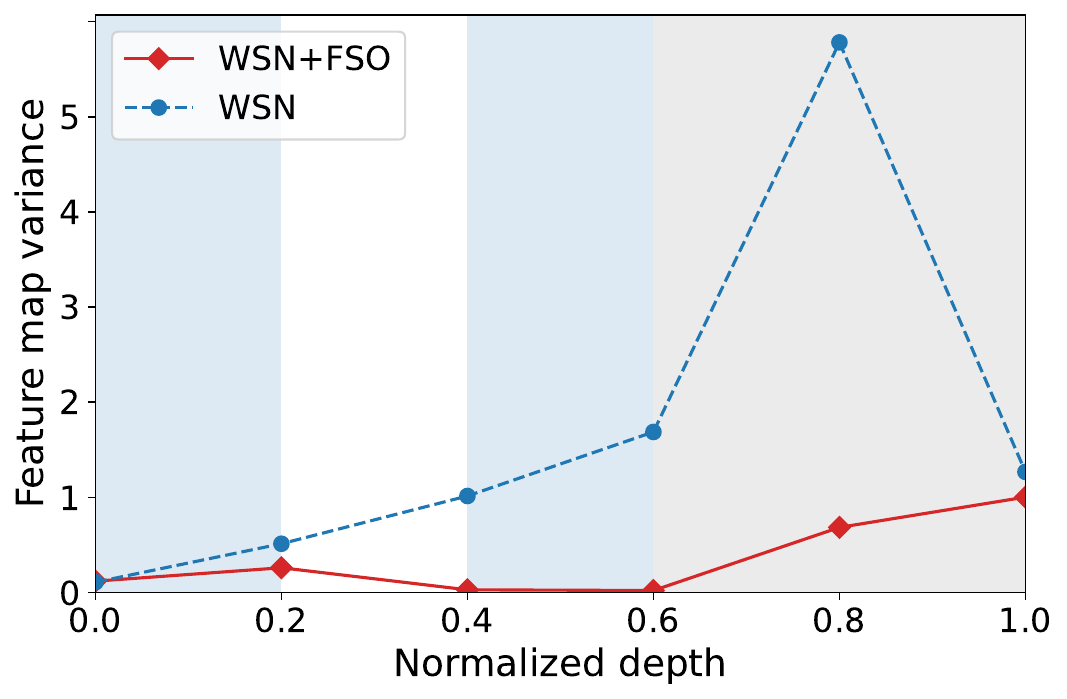} &
    \includegraphics[width=0.5\columnwidth]{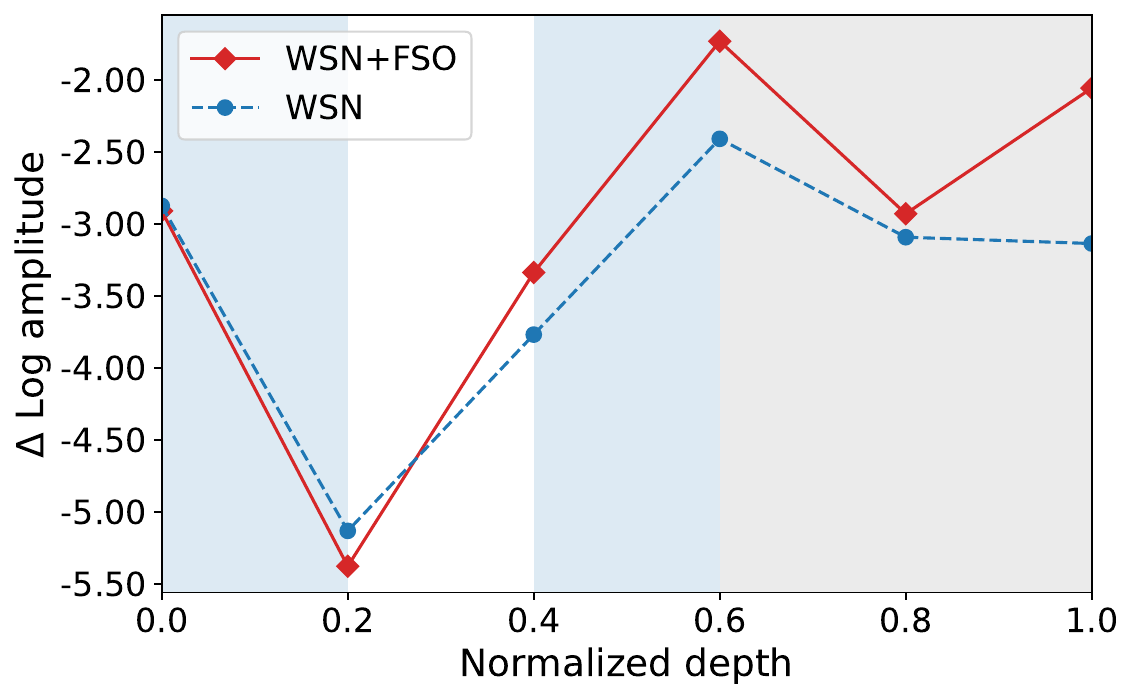} \\
    \small (a) feature map variance & \small (b) high-freq. of feature map\\

    \end{tabular}}
    \vspace{-0.05in}
    \caption{\small \textbf{The comparisons of WSN of 4 Conv (blue area) and 3 FC (gray area) with FSO (white area) in terms of Feature variances and high-frequency components}: the (a) offers the variance of the feature map and the (b) provides $\Delta \log$ amplitudes at high-frequency ($1.0 \pi$).}
    \label{fig:var_vs_freq}
\vspace{-0.1in}
\end{figure}

%% file: materials/4_algorithm.tex
\begin{algorithm}[ht]
  \caption{Training WSNs for TIL, TaIL, and VIL.}\label{alg:algorithm1}
    \small
    \begin{algorithmic}[1]
    \INPUT $\{\mathcal{D}_s\}_{s=1}^{\mathcal{S}}$, model weights $\bm{\theta}_\ast=\{\bm{\theta}, \bm{\phi}_{FSO}\}$, score weights $\bm{\rho}_\ast=\{\bm{\rho}, \bm{\rho}_{FSO}\}$, binary mask $\mathbf{M}_0=\{\bm{0}^{|\bm{\theta}|}, \bm{0}^{|\bm{\theta}_{FSO}|}\}$, layer-wise capacity $c \%$.
    \STATE Randomly initialize $\bm{\theta}_\ast$ and $\bm{\rho}_\ast$.
    \FOR{task $s = 1, \cdots , \mathcal{S}$}
        \FOR{batch $\bm{b}_s \sim \mathcal{D}_s$} 
            \STATE Obtain mask $\bm{m}_s$ of the top-$c\%$ scores $\bm{\rho}_\ast$
            \STATE Compute $\mathcal{L}\left( \bm{\theta}_\ast \odot \bm{m}_s;\bm{b}_s \right)$
            \STATE $\bm{\theta}_\ast \leftarrow \bm{\theta}_\ast - \eta \left(\frac{\partial \mathcal{L}}{\partial \bm{\theta}_\ast} \odot (\bm{1}-\bm{M}_{s-1})\right)$ \COMMENT{Weight update}     
            \STATE $\bm{\rho}_\ast \leftarrow \bm{\rho}_\ast - \eta(\frac{\partial \mathcal{L}}{\partial \bm{\rho}_\ast})$ \COMMENT{Weight score update}
        \ENDFOR
        \STATE $\bm{M}_{s} \leftarrow \bm{M}_{s-1} \newor \bm{m}_s$ \COMMENT{Accumulate binary mask}
    \ENDFOR
  \end{algorithmic}
\end{algorithm}

%% file: soft-materials/5_softnet/4_algorithm.tex
\begin{algorithm}[ht]
  \caption{Soft-Subnetworks (SoftNet) for FSCIL.}\label{alg:algorithm2}
    \small
    \begin{algorithmic}[1]
    \INPUT $\{\mathcal{D}^t\}_{t=1}^{\mathcal{T}}$, model weights $\bm{\theta}_\ast=\{\bm{\theta}, \bm{\phi}_{FSO}\}$, and score weights $\bm{\rho}_\ast=\{\bm{\rho}, \bm{\rho}_{FSO}\}$, layer-wise capacity $c$ \\
    \STATE \textcolor{blue}{// Training over base classes $s = 1$} \\ 
    \STATE Randomly initialize $\bm {\theta}_\ast$ and $\bm{\rho}_\ast$. \\
    \FOR{epoch $e = 1, 2, \cdots$}
        \STATE Obtain softmask $\bm{m}_\text{soft}$ of $\bm{m}_{major}$ and $\bm{m}_{minor} \sim U(0,1)$ at each layer \\
        \FOR{batch $\bm{b}_s\sim \mathcal{D}^s$} 
            \STATE Compute $\mathcal{L}_{base}\left( \bm{\theta}_\ast \odot \bm{m}_\text{soft};\bm{b}_s \right)$ by \Cref{eq:optimizer}
            
            \STATE $\bm{\theta}_\ast \leftarrow \bm{\theta}_\ast - \alpha \left(\frac{\partial \mathcal{L}}{\partial \bm{\theta}_\ast} \odot \bm{m}_\text{soft}\right)$ 
            \STATE $\bm{\rho}_\ast \leftarrow \bm{\rho}_\ast - \alpha \left(\frac{\partial \mathcal{L}}{\partial \bm{\rho}_\ast} \odot \bm{m}_\text{soft}\right)$ 
        \ENDFOR
    \ENDFOR \\
    \STATE \textcolor{blue}{// Incremental learning $s \geq 2$} \\
    \STATE \text{Combine the training data $\mathcal{D}^s$} \\
    \STATE \text{~~~and the exemplars saved in previous few-shot sessions} \\
    \FOR{epoch $e = 1, 2, \cdots$}
        \FOR{batch $\bm{b}_s \sim \mathcal{D}^s$} 
            \STATE Compute $\mathcal{L}_{m}\left( \bm{\theta}_\ast \odot \bm{m}_\text{soft};\bm{b}_s \right)$ by 
            \Cref{eq:proto_loss} 
            
            \STATE $\bm{\theta}_\ast \leftarrow \bm{\theta}_\ast - \beta \left(\frac{\partial \mathcal{L}}{\partial \bm{\theta}_\ast} \odot \bm{m}_{\textcolor{red}{minor}} \right)$ 
        \ENDFOR
    \ENDFOR \\
    \OUTPUT model parameters $\bm{\theta}_\ast$, $\bm{\rho}_\ast$, and $\bm{m}_{\text{soft}}$.
  \end{algorithmic}
\end{algorithm}

%% file: 5_experiment.tex
\section{Experiments}
We validate our method on several benchmark datasets against relevant continual learning baselines on Task-Incremental Learnings (TIL and TaIL, \Cref{sub_sec:opt_til}), Video Incremental Learning (VIL, \Cref{sub_sec:opt_vil}), and Few-shot Class Incremental Learning (FSCIL, \Cref{sub_sec:opt_fscil}).

\subsection{Task-incremental Learning (TIL)}

\noindent 
\textbf{Datasets and architectures.} We use three different popular sequential datasets for CL problems with three different neural network architectures as follows: {1) CIFAR-100 Split} \cite{Krizhevsky2009}{:} A visual object dataset constructed by randomly dividing 100 classes of CIFAR-100 into ten tasks with ten classes per task. {2) CIFAR-100 Superclass:} We follow the setting from \cite{Yoon2020} that divides CIFAR-100 dataset into 20 tasks according to the 20 superclasses, and each superclass contains five different but semantically related classes. {3) TinyImageNet}~\cite{Stanford}{:} A variant of ImageNet~\cite{krizhevsky2012imagenet} containing 40 of 5-way classification tasks with the image size by $64 \times 64\times 3$. 

We use variants of LeNet~\cite{LeCun1998} for the experiments on CIFAR-100 Superclass experiments, and a modified version of AlexNet similar to \cite{Serra2018, Saha2021} for the CIFAR-100 Split dataset. For TinyImageNet, we also use the same network architecture as \cite{gupta2020maml, deng2021flattening}, which consists of 4 Conv layers and 3 fully connected layers. We set WSN as the architecture-based baselines, which jointly train and find task-adaptive subnetworks of novel/prior parameters for continual learning. WSN+FSO follows the Residual Blocks as stated in \Cref{fig:residual_fso} in all three architectures. The FSO's ensemble structures follow in terms of three architectures:

\begin{itemize}[leftmargin=*]
    \item \small \textcolor{black}{In the LeNet, the Conv. layer's output $\bm{x}^{l=1}$ is merged into the $l=2$th Conv. layer output through the FSO to acquire spatially ensembling features $\bm{x}^{l=2}$.}

    \item \small \textcolor{black}{In the AlexNet, the Conv. layer's output $\bm{x}^{l=2}$ is merged into the $l=3$th Conv. layer's output through the FSO to acquire spatially ensembling features $\bm{x}^{l=3}$.}

    \item \small \textcolor{black}{In the TinyImageNet, the Conv. layer's output $\bm{x}^{l=2}$ is merged into the $l=3$th Conv. layer's output through the FSO to acquire spatially ensembling features $\bm{x}^{l=3}$.}
\end{itemize}

\noindent
\textbf{Experimental settings.} \label{exp_setting} 
As we directly implement our method from the official code of \cite{Saha2021}, we provide the values for HAT and GPM reported in \cite{Saha2021}. For Omniglot Rotation and Split CIFAR-100 Superclass, we deploy the proposed architecture in multi-head settings with hyperparameters as reported in \cite{Yoon2020}. All our experiments run on a single-GPU setup of NVIDIA V100. We evaluate all methods based on the following two metrics: 
\begin{enumerate}[itemsep=0em, topsep=-1ex, itemindent=0em, leftmargin=1.2em, partopsep=0em]
\item \small {\textit{Accuracy (ACC)}} measures the average of the final classification accuracy on all tasks: $\mathrm{ACC}=\frac{1}{\mathcal{T}} \sum_{i=1}^{\mathcal{T}} A_{\mathcal{T}, i}$, where $A_{\mathcal{T},i}$ is the test accuracy for task $i$ after training on task $\mathcal{T}$.

\item \small {\textit{Backward Transfer (BWT)}} measures the forgetting during continual learning. Negative BWT means that learning new tasks causes the forgetting of past tasks: $\mathrm{BWT}=\frac{1}{\mathcal{T}-1}\sum_{i=1}^{\mathcal{T}-1} A_{\mathcal{T}, i}-A_{i, i}$. \\
\end{enumerate}

\noindent 
\textbf{Baselines.} We compare our WSN with strong CL baselines; regularization-based methods: {HAT}~\cite{Serra2018} and {EWC}~\cite{Kirkpatrick2017}, rehearsal-based methods: {GPM}~\cite{Saha2021}, and a pruning-based method: {PackNet}~\cite{mallya2018packnet} and {SupSup}~\cite{wortsman2020supermasks}. {PackNet} and {SupSup} is set to the baseline to show the effectiveness of re-used weights. We also compare with a naive sequential training strategy, referred to as {FINETUNE}. {Multitask Learning (MTL)} and {Single-task Learning (STL)} are not a CL method. MTL trains on multiple tasks simultaneously, and STL trains on single tasks independently.

\subsection{Task-agnositic Incremental Learning (TaIL)}

\textbf{Datasets.} We evaluate our WSN+FSO on three popular datasets: Seq-CIFAR10~\cite{aljundi2019online}, Seq-CIFAR100~\cite{chrysakis2020online}, and Seq-TinyImageNet~\cite{le2015tiny}. Seq-CIFAR10 comprises 5 disjoint tasks containing 2 classes and 10k training samples. Seq-CIFAR100 consists of 5 disjoint tasks, each with 20 classes and 10k training samples. Seq-TinyImageNet includes 10 disjoint tasks, each with 20 classes and 10k training samples. Detailed statistics for these datasets can be found in~\cite{liang2024loss}. All experiments address task-agnostic problems where no task ID is provided during training and testing. \\

\noindent 
\textbf{Baseline.} We compare our FSO with baselines (replay-based methods, \textit{Finetune}, WSN, and WSN+FSO, \textit{Joint}) under the experimental setting~\cite{liang2024loss}, as shown in \Cref{table:tail_acc}. Here, \textit{Joint} (Upper-bound) denotes the method that all the tasks jointly while \textit{Finetune} (Lower-bound) denotes the method that learns all tasks sequentially without any memory buffers. Additionally, we compare WSN+FSO with replay-based continual learning methods that maintain a single learning model to perform continual learning (without keeping the extra model~\cite{liang2024loss}). Note that FSO is used at the 3th residual blocks of ResNet18 for TaIL. \\ 

\noindent
\textbf{Training and Testing.} In training time, we follow the continual learning methods~\cite{buzzega2020dark, liang2024loss} with standard ResNet18 for all the task-agnostic experiments. We also use stochastic gradient descent (SGD) to optimize the parameters of WSN+FSO. The batch size is set to 32 for fair comparisons with the prior works. For our experiment results, we report the average and standard deviation of the mean test accuracy of all the sessions across 5 runs with different seeds. At the last epoch of the $s$-th session, we obtain the current subnetwork for the current session and save the subnetwork sequentially so that we have $s$ numbers of subnetworks. Note that we follow the subnetwork's session identification algorithm in the test for TaIL, as stated in \Cref{sub_sec:opt_wsn}.

\subsection{Few-shot Class Incremental Learning (FSCIL)}
We introduce experimental setups - Few-Shot Class Incremental Learning (FSCIL) settings to provide soft-subnetworks' effectiveness. We empirically evaluate and compare our soft subnetworks with state-of-the-art methods and vanilla subnetworks in the following subsections. 

\noindent
\textbf{Datasets.} To validate the effectiveness of the soft subnetwork, we follow the standard FSCIL experimental setting. We randomly selected 60 classes as the base and 40 as new classes for CIFAR-100 and miniImageNet. In each incremental learning session, we construct 5-way 5-shot tasks by randomly picking five classes and sampling five training examples for each class; we set the first 100 classes of CUB-200-2011 as base classes and the remaining 100 classes as new categories split into 10 novel sessions (i.e., a 10-way 5-shot).

\noindent
\textbf{Baselines.} We mainly compare our SoftNet~\cite{kang2022soft} with architecture-based methods for FSCIL: FSLL~\cite{mazumder2021few} that selects important parameters for each session, and HardNet, representing a binary subnetwork. Furthermore, we compare other FSCIL methods such as iCaRL~\cite{rebuffi2017icarl}, Rebalance~\cite{hou2019learning}, TOPIC~\cite{tao2020few}, IDLVQ-C~\cite{chen2020incremental}, and F2M~\cite{shi2021overcoming}. \textcolor{black}{Fourier Subneural Operator (FSO) is used at the 3th residual blocks of ResNet18. The $3$th residual block's output $\bm{x}^{l=3}$ is merged into the $l=4$th residual block through the FSO to acquire spatially ensembling features $\bm{x}^{l=4}$.} We also include a joint training method~\cite{shi2021overcoming} that uses all previously seen data, including the base and the following few-shot tasks for training as a reference. Furthermore, we fix the classifier re-training method (cRT)~\cite{kang2019decoupling} for long-tailed classification trained with all encountered data as the approximated upper bound. 

\input{materials/5_main_table_TIL}

\noindent
\textbf{Experimental settings.} The experiments are conducted with NVIDIA GPU RTX8000 on CUDA 11.0. We also randomly split each dataset into multiple sessions. We run each algorithm ten times for each dataset and report their mean accuracy. We adopt ResNet18~\cite{he2016deep} as the backbone network. For data augmentation, we use standard random crop and horizontal flips. In the base session training stage, we select top-$c\%$ weights at each layer and acquire the optimal soft-subnetworks with the best validation accuracy. In each incremental few-shot learning session, the total number of training epochs is $6$, and the learning rate is $0.02$. We train new class session samples using a few minor weights of the soft-subnetwork (Conv4x layer of ResNet18) obtained by the base session learning.

\subsection{Video incremental Learning (VIL)}
We validate our method on video benchmark datasets against continual learning baselines on Video Incremental Learning (VIL). We consider continual video representation learning with a multi-head configuration (session id, i.e., $s$ is given in training and inference) for all experiments in the paper. We follow the experimental setups in NeRV~\cite{chen2021nerv} and HNeRV~\cite{chen2023hnerv}. 

\noindent 
\textbf{Datasets and architectures.}
We conducted an extended experiment on the UVG of 8/17 video sessions. The category index and order in UVG8 (\textit{1.bunny, 2.beauty, 3.bosphorus, 4.bee, 5.jockey, 6.setgo, 7.shake, 8.yacht}) and UVG17 (\textit{1.bunny, 2.city, 3.beauty, 4.focus, 5.bosphorus, 6.kids, 7.bee, 8.pan, 9.jockey, 10.lips, 11.setgo, 12.race, 13.shake, 14.river, 15.yacht, 16.sunbath, 17.twilight}). We employ NeRV as our baseline architecture and follow its details for a fair comparison. After the positional encoding, we apply 2 sparse MLP layers on the output of the positional encoding layer, followed by five sparse NeRV blocks with upscale factors of 5, 2, 2, 2, 2. These sparse NeRV blocks decode 1280$\times$720 frames from the 16$\times$9 feature map obtained after the sparse MLP layers. For the upscaling method in the sparse NeRV blocks, we also adopt PixelShuffle~\cite{shi2016real}. \textcolor{black}{Fourier Subneural Operator (FSO) is used at the NeRV2 or NeRV3 layer, denoted as $f$-NeRV2 and $f$-NeRV3\footnote{The $f$-NeRV3 code is available at \url{https://github.com/ihaeyong/PFNR.git}}).} The positional encoding for the video index $s$ and frame index $t$ is as follows:
\begin{equation*}
\begin{split}
\bm{\Gamma}(s, t) = &[~\sin(b^0\pi s), \cos(b^0\pi s), \cdots ,\sin(b^{l-1}\pi s), \cos(b^{l-1}\pi s), \\ & ~~\sin(b^0\pi t), \cos(b^0\pi t), \cdots ,\sin(b^{l-1}\pi t), \cos(b^{l-1}\pi t)~],
\end{split}
\label{eq:pos_embedding}
\end{equation*}
where the hyperparameters are set to $b=1.25$ and $l=80$ such that $\bm{\Gamma}(s, t) \in \mathbb{R}^{1 \times 160}$. As differences from the previous NeRV model, the first layer of the MLP has its input size expanded from 80 to 160 to incorporate both frame and video indices, and distinct head layers after the NeRV block are utilized for each video. For the loss objective in \Cref{eq:l1_loss}, $\alpha$ is set to $0.7$. We evaluate the video quality, average video session quality, and backward transfer with PSNR. 

\noindent 
\textbf{Baselines.} To show the effectiveness, we compare our WSN+FSO with strong CL baselines: Single-Task Learning (STL), which trains on single tasks independently, EWC~\cite{Kirkpatrick2017}, which is a regularized baseline, iCaRL~\cite{rebuffi2017icarl}, and ESMER~\cite{sarfraz2023error} which are current strong rehearsal-based baseline, WSN~\cite{kang2022forget} which is a current strong architecture-based baseline, and Multi-Task Learning (MTL) which trains on multiple video sessions simultaneously, showing the upper-bound of WSN. Except for STL, all models are trained and evaluated on multi-head settings where a video session and time $(s, t)$ indices are provided.

\noindent 
\textbf{Training.} In all experiments, we follow the same experimental settings as NeRV~\cite{chen2023hnerv} and HNeRV~\cite{chen2023hnerv} for fair comparisons. We train WSN+FSO, NeRV (STL), and MTL using Adam optimizer with a learning rate 5e-4. For the ablation study on UVG17, we use a cosine annealing learning rate schedule~\cite{loshchilov2016sgdr}, batch size of 1, training epochs of 150, and warmup epochs of 30 unless otherwise denoted.

\noindent 
\textbf{VIL's performance metrics} We evaluate all methods based on the following continual learning metrics: 
\begin{enumerate}[itemsep=0em, topsep=-1ex, itemindent=0em, leftmargin=1.2em, partopsep=0em]
\item \small {\textit{Average Peak signal-to-noise ratio (PSNR)}} measures the average of the final performances on all video sessions: $\mathrm{PSNR} =\frac{1}{N} \sum_{s=1}^{N} A_{N, s}$, where $A_{N,s}$ is the test PSNR for session $s$ after training on the final video session $S$.   

\item \small {\textit{Backward Transfer (BWT) of PSNR}} measures the video representation forgetting during continual learning. Negative BWT means that learning new video sessions causes the video representation forgetting of past sessions: $\mathrm{BWT}=\frac{1}{N-1}\sum_{s=1}^{N-1} A_{N, s}-A_{s, s}$. 
\end{enumerate}

\section{Results of Task-Incremental Learning}

\subsection{Comparisons with baselines in TIL}
We use a multi-head setting to evaluate our WSN algorithm under the more challenging visual classification benchmarks. The WSN+FSO's performances are compared with others in terms of two measurements on three major benchmark datasets as shown in \Cref{tab:main_sota_table}. Our WSN+FSO outperformed all state-of-the-art, achieving the best average accuracy of 79.00\%, 61.70\%, and 72.04\%. WSN+FSO is also a forget-free model (BWT = ZERO), aligned with architecture-based models such as PackNet, SupSup, and WSN in these experiments. In addition, to show the effectiveness of the large single-scale task performance, we prepare WSN+FSO trained on the ImageNet-1K dataset, as shown in \Cref{table:imagnet1k}. WSN+FSO outperformed all baselines.

\subsection{Statistics of WSN+FSO's Representations in TIL}
In the TinyImageNet task, the TinyImageNet takes 4 convolutional operators, followed by 3 fully connected layers. These operations represent more abstract features pooled by high-frequency components while losing low-frequency ones. To compensate for low-frequency components, we add an FSO to CNN architecture as shown in \Cref{fig:residual_fso}. \textcolor{black}{The lower layer's output $\bm{x}^{l=2}$ is merged into the $l=3$th layer residual block through the FSO to acquire spatially ensembling features $\bm{x}^{l=3}$.} The FSO also provides additional parameters to push the residual to zero. We will show the differences between ensembling features (WSN+FSO) and single features (WSN) represented by residual blocks as shown in \Cref{fig:var_vs_freq}: WSN+FSO provides lower variances of feature maps and higher frequency components than WSN. The ensemble of representations led to better performances.

\section{Results On Task-agnostic IL (TaIL)}
\textbf{Performances.} We set a baseline as WSN to compare with other SOTA methods in the Task-agnostic Incremental Learning (TaIL) scenario. In seq-CIFAR10 and seq-CIFAR100, WSN+FSO (without any buffer sample) outperformed all baselines and the upper bound (MTL), as shown in \Cref{table:tail_acc}. Concretely, FSO's global abstraction power in the residual block led to the best performances across all agnostic tasks, as shown in \Cref{fig:TaIL_cifar}. In seq-TinyImageNet, the performance of WSN+FSO is lower than that of LODE (DER++). The following reasons could explain the results. First, the loss decoupling (LODE) successfully adjusted the predictions in the TaIL setting, as shown in the performances of the WSN + FSO + LODE. Second, replay buffer samples play an essential role in the TaIL setting. As the buffer size increases, the performances of replay buffer-based methods also increase dramatically, as shown in the performances of DER++. Lastly, replay buffer samples with loss decoupling, i.e., LODE(DER++), show an ideal condition for the best results in the TynyImageNet of TaIL setting. In contrast,  the WSN framework focuses only on forward learning, leveraged by reusing previously learned parameters. Such that updating previous knowledge is not allowed by replaying samples. To overcome the issue, we focus on updating previously learned parameters through minimal sets of replay samples in future works.

\input{supples/materials/table_task_agnostic}

\input{supples/materials/plot_task_agnostic}

\section{Results On Few-shot CIL (FSCIL)}

\input{soft-materials/6_exper/6_main_cifar100_resnet18_5way_5shot_split}
\input{soft-materials/6_exper/6_main_miniImageNet_5way_5shot_split}
\input{soft-materials/6_exper/5_main_cub_10way_5shot}

\noindent 
\textbf{Comparisons with SOTA}. We compare SoftNet+FSO with the following state-of-art-methods on TOPIC class split~\cite{tao2020few} of three benchmark datasets - CIFAR100 (\Cref{tab:main_cifar100_5way_5shot_resnet18_split}), miniImageNet (\Cref{tab:miniImageNet_5way_5shot_baseline_split}), and CUB-200-2011 (\Cref{tab:main_cub200_10way_5shot}). Leveraged by regularized backbone ResNet, SoftNet+FSO outperformed all existing architecture baselines FSLL~\cite{mazumder2021few}, FACT~\cite{zhou2022forward}, and WaRP~\cite{kim2023warping} on CIFAR100, miniImageNet. On CUB-200-201, the performances of SoftNet+FSO are comparable with those of ALICE and LIMIT, considering that ALICE used class/data augmentations and LIMIT added an extra multi-head attention layer. Note NC-FSCIL~\cite{yang2023neural} could not be comparable with SoftNet+FSO since it focuses mainly on replaying prototype-based classifiers rather than backbone representations to obtain balanced categorical prototypes. Lastly, before finetuning SoftNet+FSO on CUB-200-2011, we pre-trained WSN+FSO on the ImageNet-1K dataset, as shown in \Cref{table:imagnet1k}. The WSN+FSO outperforms WSN and baselines.

\input{FSO-materials/main_table_imagenet}

\input{FSO-materials/main_table_uvg17_psnr}

\section{Results of Video Incremental Learning}
\subsection{Comparisons with Baselines} %
\textbf{Video Representations.} To compare WSN+FSO with conventional representative continual learning methods such as EWC, iCaRL, ESMER, and WSN+FSO, we prepare the reproduced results, as shown in \Cref{table:uvg17_fso_psnr}. The architecture-based WSN outperformed the regularized method and replay method. The sparseness of WSN does not significantly affect sequential video representation results on two sequential benchmark datasets. Our WSN+FSO outperforms all conventional baselines including WSN and MTL (upper-bound of WSN) on the UVG17 benchmark datasets. Moreover, our performances of WSN with $f$-NeRV3 are better than those of $f$-NeRV2 since $f$-NeRV3 tends to represent local textures, stated in the following \Cref{sec:video_reps}. Note that the number of parameters of MLT is precisely the same as those of WSN.

\input{FSO-materials/plot_main_video_mtl}

\noindent 
\textbf{Compression.} We follow NeRV's video quantization and compression pipeline~\cite{chen2021nerv}, except for the model pruning step, to evaluate performance drops and backward transfer in the video sequential learning, as shown in \Cref{fig:uvg17_fso_psnr}. Once sequential training is done, our WSN+FSO doesn't need any extra pruning and finetuning steps, unlike NeRV. This point is our key advantage of WSN+FSO over NeRV. \Cref{fig:uvg17_fso_psnr} (a) shows the results of various sparsity and bit-quantization on the UVG17 datasets: the 8bit WSN+FSO's performances are comparable with 32bit ones without a significant video quality drop. From our observations, the 8-bit subnetwork seems to be enough for video implicit representation. \Cref{fig:uvg17_fso_psnr} (b) shows the rate-distortion curves. We compare WSN+FSO with WSN and NeRV (STL). For a fair comparison, we take steps of pruning, fine-tuning, quantizing, and encoding NeRV. Our WSN+FSO outperforms all baselines.

\noindent 
\textbf{Performance and Capacity.} Our WSN+FSO outperforms WSN and MTL, as stated in \Cref{fig:psnr_cap} (a). This result might suggest that properly selected weights in Fourier space lead to generalization more than others in VIL. Moreover, to show the behavior of FSO, We prepare a progressive WSN+FSO's capacity and investigate how FSO reuses weights over sequential video sessions, as shown in \Cref{fig:psnr_cap} (b). WSN+FSO tends to progressively transfer weights used for a prior session to weights for new ones, but the proposition of reused weights gets smaller as video sessions increase.

\subsection{WSN+FSO's Video Representations}\label{sec:video_reps}
We prepare the results of video generation as shown in \Cref{fig:video_reinit_mtl}. We demonstrate that a sparse solution (WSN with $c=30.0 \%$, $f$-NeRV3) generates video representations sequentially without significant performance drops. Compared with WSN, WSN+FSO provides more precise representations. To find out the results, we inspect the layer-wise representations as shown in \Cref{fig:fmap_uvg17}, which offers essential observations that WSN+FSO tends to capture local textures broadly at the NeRV3 layer while WSN focuses on local objects. This WSN+FSO behavior could lead to more generalized performances. Moreover, we conduct an ablation study to inspect the sparsity-wise performances of $f$-NeRV3 while holding the remaining parameters' sparsity (c=50.0 \%), as shown in \Cref{fig:fmap_sparsity_uvg17}. We could observe that as the sparsity of $f$-NeRV3 increases, its performances drop. This leads us to the importance of $f$-NeRV3's representations.

\input{FSO-materials/plot_main_fmap_uvg17}

\input{FSO-materials/plot_fso_fmap_uvg17}

\input{FSO-materials/plot_main_pfnr_var_freq}

\input{FSO-materials/plot_main_psnr_bpp}

\input{FSO-materials/plot_capacity}

\subsection{Statistics of WSN+FSO's Representations in VIL}
We provide the statistics of WSN+FSO's video representations, as shown in \Cref{fig:pfnr_var_vs_freq}. In the video incremental task, the NeRV architecture takes stems of 2 fully connected layers, followed by 5 convolutional operators. These operations represent an image up-scaled by multiple scalars while losing high-frequency representations (see WSN's representations of \Cref{fig:fmap_uvg17}). To compensate for high-frequency components, we add an FSO to NeRV blocks as shown in \Cref{fig:concept_CVRNet}. The Conv. layer's output is merged into the outputs of FSO to acquire spatially ensembling features. We will show the differences between ensembling features (WSN+FSO) and single features (WSN), as shown in \Cref{fig:pfnr_var_vs_freq}: WSN+FSO provides a high variance of representations and higher frequency components than WSN at NeRV block 3. The ensemble of representations led to better performances in VIL, supporting the representations of \Cref{fig:fmap_uvg17}.

\subsection{Ablation Studies of WSN+FSO}
\textbf{Variations of FSO.} We prepared several ablation studies to prove the effectiveness of FSO. First, we show the performances of only real part (ignore an imaginary part) in f-NeRV2/3 as shown in \Cref{table:uvg8_fso_real}. The PSNR performances of only real part were lower than those of both real and imaginary parts in f-NeRV2/3. We infer that the imaginary part of the winning ticket improves the implicit neural representations. Second, we also investigate the effectiveness of only FSO without Conv. Layer in f-NeRV2/3, as shown in \Cref{table:uvg8_fso_conv}. The PSNR performances were lower than FSO with Conv block. Therefore, the ensemble of FSO and Conv improves the implicit representations. Lastly, we investigate the effectiveness of sparse FSO in STL, as shown in \Cref{table:uvg8_stl}. The sparse FSO boots the PSNR performances in STL. These ablation studies further strengthen the effectiveness of FSO for sequential neural implicit representations.

\input{FSO-materials/main_table_uvg8_psnr_fso}

\input{FSO-materials/main_table_uvg8_psnr_conv}

\input{FSO-materials/main_table_uvg8_psnr_stl}

\noindent 
\textbf{Forget-free Transfer Matrix.} We prepare the transfer matrix to prove our WSN+FSO's forget-freeness and to show video correlation among other videos, as shown in \Cref{fig:transf_matrix} on the UVG17 dataset; lower triangular estimated by each session subnetwork denotes that our WSN+FSO is a forget-free method and upper triangular calculated by current session subnetwork denotes the video similarity between source and target. The WSN+FSO proves the effectiveness from the lower triangular of \Cref{fig:transf_matrix} (a) and (b). Nothing special is observable from the upper triangular since they are not correlated, however, there might be some shared representations.

\input{FSO-materials/plot_transfer_matrix}

%% file: materials/5_main_table_TIL.tex
\begin{table*}[ht]
\begin{center}
\caption{\small \textbf{(TIL)}, {Performance comparisons of the proposed method and other baselines} - PackNet~\cite{mallya2018packnet} and SupSup~\cite{wortsman2020supermasks} - on various benchmark datasets. We report the mean and standard deviation of the average accuracy (ACC) and average backward transfer (BWT) across $5$ independent runs with five seeds under the same experimental setup \cite{deng2021flattening}. $\dagger$ $~$ denotes results reported from \cite{deng2021flattening}.}
\vspace{-0.1in}
\resizebox{0.8\textwidth}{!}{
\begin{tabular}{lcccccc}
\toprule
\multicolumn{1}{c}{{\textbf{Method}}}&\multicolumn{2}{c}{\textbf{CIFAR-100 Split}}&
\multicolumn{2}{c}{\textbf{CIFAR-100 Superclass}}&
\multicolumn{2}{c}{\textbf{TinyImageNet}} \\

\midrule
& ACC (\%) & BWT (\%) & ACC (\%) & BWT (\%) & ACC (\%) & BWT (\%) \\
\midrule

La-MaML \cite{gupta2020maml} & 71.37~\scriptsize($\pm$ 0.7)$^\dagger$ & -5.39~\scriptsize($\pm$ 0.5)$^\dagger$  & 54.44~\scriptsize($\pm$ 1.4)$^\dagger$  & -6.65~\scriptsize($\pm$ 0.9)$^\dagger$  & 66.90~\scriptsize($\pm$ 1.7)$^\dagger$  & -9.13~\scriptsize($\pm$ 0.9)$^\dagger$  \\

GPM \cite{Saha2021} & 73.18~\scriptsize($\pm$ 0.5)$^\dagger$ & -1.17~\scriptsize($\pm$ 0.3)$^\dagger$ & 57.33~\scriptsize($\pm$ 0.4)$^\dagger$  & -0.37~\scriptsize($\pm$ 0.1)$^\dagger$ & 67.39~\scriptsize($\pm$ 0.5)$^\dagger$  & ~\textbf{1.45}~\scriptsize($\pm$ \textbf{0.2})$^\dagger$  \\

FS-DGPM \cite{deng2021flattening} & 74.33~\scriptsize($\pm$ 0.3)$^\dagger$ & -2.71~\scriptsize($\pm$ 0.2)$^\dagger$ & 58.81~\scriptsize($\pm$ 0.3)$^\dagger$  & -2.97~\scriptsize($\pm$ 0.4)$^\dagger$ & 70.41~\scriptsize($\pm$ 1.3)$^\dagger$  & -2.11~\scriptsize($\pm$ 0.9)$^\dagger$ \\
 
\midrule

PackNet~\cite{mallya2018packnet} & 72.39~\scriptsize($\pm$ 0.4) & \textbf{0.0} & 58.78~\scriptsize($\pm$ 0.5)  & \textbf{0.0} & 55.46~\scriptsize($\pm$ 1.2) & \textbf{0.0} \\

SupSup~\cite{wortsman2020supermasks} & 75.47~\scriptsize($\pm$ 0.3) & \textbf{0.0} & 61.70~\scriptsize($\pm$ 0.3)  &  \textbf{0.0} & 59.60~\scriptsize($\pm$ 1.1)  & \textbf{0.0} \\
\midrule

WSN$^\ast$, $c=50\%$ & 77.67~\scriptsize($\pm$ 0.1) & \textbf{0.0} & 
{61.58}~\scriptsize($\pm$ {0.0}) & \textbf{0.0} & 
69.88~\scriptsize($\pm$ 1.7) & \textbf{0.0}  \\

WSN,~~ $c=50\%$ + \textbf{FSO} & \textbf{79.00}~\scriptsize($\pm$ 0.3) &  \textbf{0.0} & 
\textbf{61.70}~\scriptsize($\pm$ \textbf{0.2}) & \textbf{0.0} & 
\textbf{72.04}~\scriptsize($\pm$ 0.7) & \textbf{0.0}  \\

\midrule

MTL (Upper-bound) & ~~79.75~\scriptsize($\pm$ 0.4)$^\dagger$ & - & ~~61.00~\scriptsize($\pm$ 0.2)$^\dagger$ & - & ~~77.10~\scriptsize($\pm$ 1.1)$^\dagger$ & - \\

\bottomrule
\end{tabular}}
\label{tab:main_sota_table}
\end{center}
\vspace{-0.15in}
\end{table*}

%% file: supples/materials/table_task_agnostic.tex
\begin{table}[!ht]
\small
\centering
\caption{\small \textbf{(TaIL)}, Incremental Classification results (without keeping the extra model~\cite{liang2024loss}), which are averaged across 5 runs with the different seeds.}
\resizebox{0.5\textwidth}{!}{
\begin{tabular}{lcccccc}
\toprule

& \multicolumn{2}{l}{~~~~~~\textbf{Seq-CIFAR10}} & \multicolumn{2}{l}{~~~~~~\textbf{Seq-CIFAR100}} & \multicolumn{2}{l}{~~~~~~\textbf{Seq-TinyImageNet}} \\
\midrule 

\textbf{Buffer Size} & 500 & 5120 & 500 & 5120 & 500 & 5120   \\ \midrule 




                   
SCR~\cite{mai2021supervised} & 57.95 \small{$\pm 1.57$}  & 82.47 \small{$\pm 0.44$} & 23.06 \small{$\pm 0.22$}  & 45.02 \small{$\pm 0.67$} & 8.37 \small{$\pm 0.26$} & 18.20 \small{$\pm 0.48$}   \\
PCR~\cite{lin2023pcr} & 65.74 \small{$\pm 3.29$}  & 82.58 \small{$\pm 0.42$} & 28.38 \small{$\pm 0.46$}  & 52.51 \small{$\pm 1.61$} & 11.88 \small{$\pm 1.61$} & 26.39 \small{$\pm 1.64$}   \\
MIR~\cite{aljundi2019online} & 63.93 \small{$\pm 0.39$}  & 83.73 \small{$\pm 0.97$} & 27.80 \small{$\pm 0.52$}  & 53.73 \small{$\pm 0.82$} & 11.22 \small{$\pm 0.43$} & 30.60 \small{$\pm 0.40$}   \\
ER-ACE~\cite{caccia2021new} & 68.45 \small{$\pm 1.78$}  & 83.49 \small{$\pm 0.40$} & 40.67 \small{$\pm 0.06$}  & 58.56 \small{$\pm 0.91$} & 17.73 \small{$\pm 0.56$} & 37.99 \small{$\pm 0.17$}   \\
ER~\cite{chaudhry2019tiny} & 61.78 \small{$\pm 0.72$}  & 83.64 \small{$\pm 0.95$} & 27.69 \small{$\pm 0.58$}  & 53.86 \small{$\pm 0.57$} & 10.36 \small{$\pm 0.11$} & 27.54 \small{$\pm 0.30$}   \\
LODE (ER)~\cite{liang2024loss} & 68.87 \small{$\pm 0.71$}  & 83.73 \small{$\pm 0.48$} & 41.52 \small{$\pm 1.22$}  & 58.59 \small{$\pm 0.48$} & 17.77 \small{$\pm 1.03$} & 38.34 \small{$\pm 0.04$}   \\
    DER++~\cite{buzzega2020dark} & 73.29 \small{$\pm 0.96$}  & 85.66 \small{$\pm 0.14$} & 42.08 \small{$\pm 1.71$}  & 62.73 \small{$\pm 0.58$} & 19.28 \small{$\pm 0.61$} & 39.72 \small{$\pm 0.47$}   \\
LODE (DER++)~\cite{liang2024loss} & 75.45 \small{$\pm 0.90$}  & 85.78 \small{$\pm 0.40$} & 46.31 \small{$\pm 1.01$}  & 64.00 \small{$\pm 0.48$} & 21.15 \small{$\pm 0.68$} & 40.31 \small{$\pm 0.03$}   \\ \midrule

Finetune (Lower-bound) & \multicolumn{2}{l}{~~~~~~~~~~19.65 \small{$\pm 0.03$}} & \multicolumn{2}{l}{~~~~~~~~~~17.41 \small{$\pm 0.09$}}   &  \multicolumn{2}{l}{~~~~~~~~~~~8.13 \small{$\pm 0.04$}}     \\ \midrule

WSN, c=70.0\%   & \multicolumn{2}{l}{~~~~~~~~~ 94.67 \small{$\pm 0.90$}} & \multicolumn{2}{l}{~~~~~~~~~ 46.24 \small{$\pm 0.60$}}   &  \multicolumn{2}{l}{~~~~~~~~~ 18.98 \small{$\pm 0.17$}} \\

WSN, c=70.0\% + \textbf{FSO} & \multicolumn{2}{l}{~~~~~~~~~ {94.90} \small{$\pm 0.50$}} & \multicolumn{2}{l}{~~~~~~~~~ {77.12} \small{$\pm 0.33$}}   &  \multicolumn{2}{l}{~~~~~~~~~ 20.90 \small{$\pm 0.21$}}     \\ 

WSN, c=70.0\% + \textbf{FSO} + LODE & \multicolumn{2}{l}{~~~~~~~~~ \textbf{96.13} \small{$\pm 0.43$}} & \multicolumn{2}{l}{~~~~~~~~~ \textbf{78.25} \small{$\pm 0.31$}}   &  \multicolumn{2}{l}{~~~~~~~~~ \textbf{21.42} \small{$\pm 0.22$}}     \\ \midrule

MTL (Upper-bound)   & \multicolumn{2}{l}{~~~~~~~~~ 91.86 \small{$\pm 0.26$}} & \multicolumn{2}{l}{~~~~~~~~~ 70.10 \small{$\pm 0.60$}}   &  \multicolumn{2}{l}{~~~~~~~~~ \textbf{59.82} \small{$\pm 0.31$}}     \\

\bottomrule
\end{tabular}
}
\label{table:tail_acc}
\vspace{-0.1in}
\end{table}

%% file: supples/materials/plot_task_agnostic.tex
\begin{figure}[h]
    \centering
    \small
    \setlength{\tabcolsep}{0pt}{%
    \begin{tabular}{cc}
    \includegraphics[width=0.5\columnwidth]{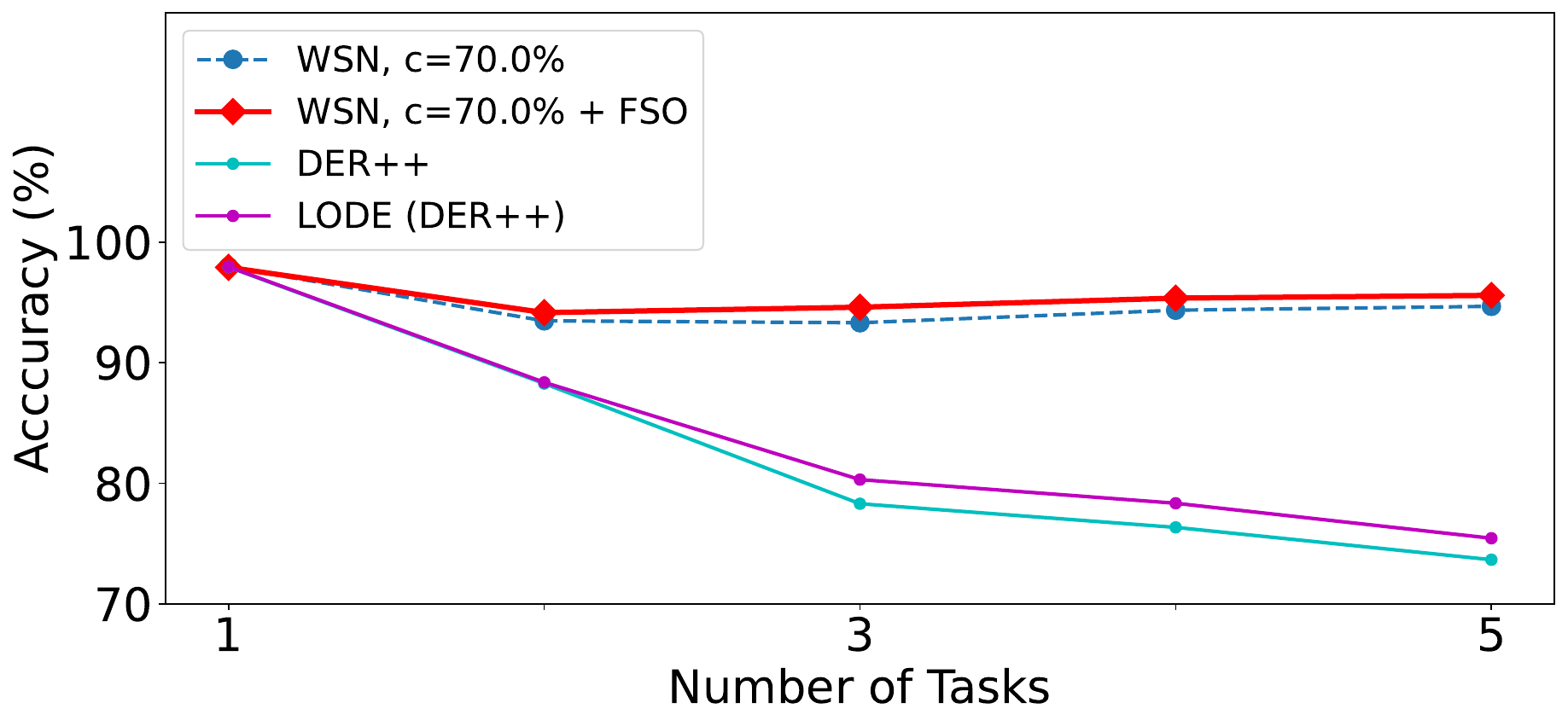} &
    \includegraphics[width=0.5\columnwidth]{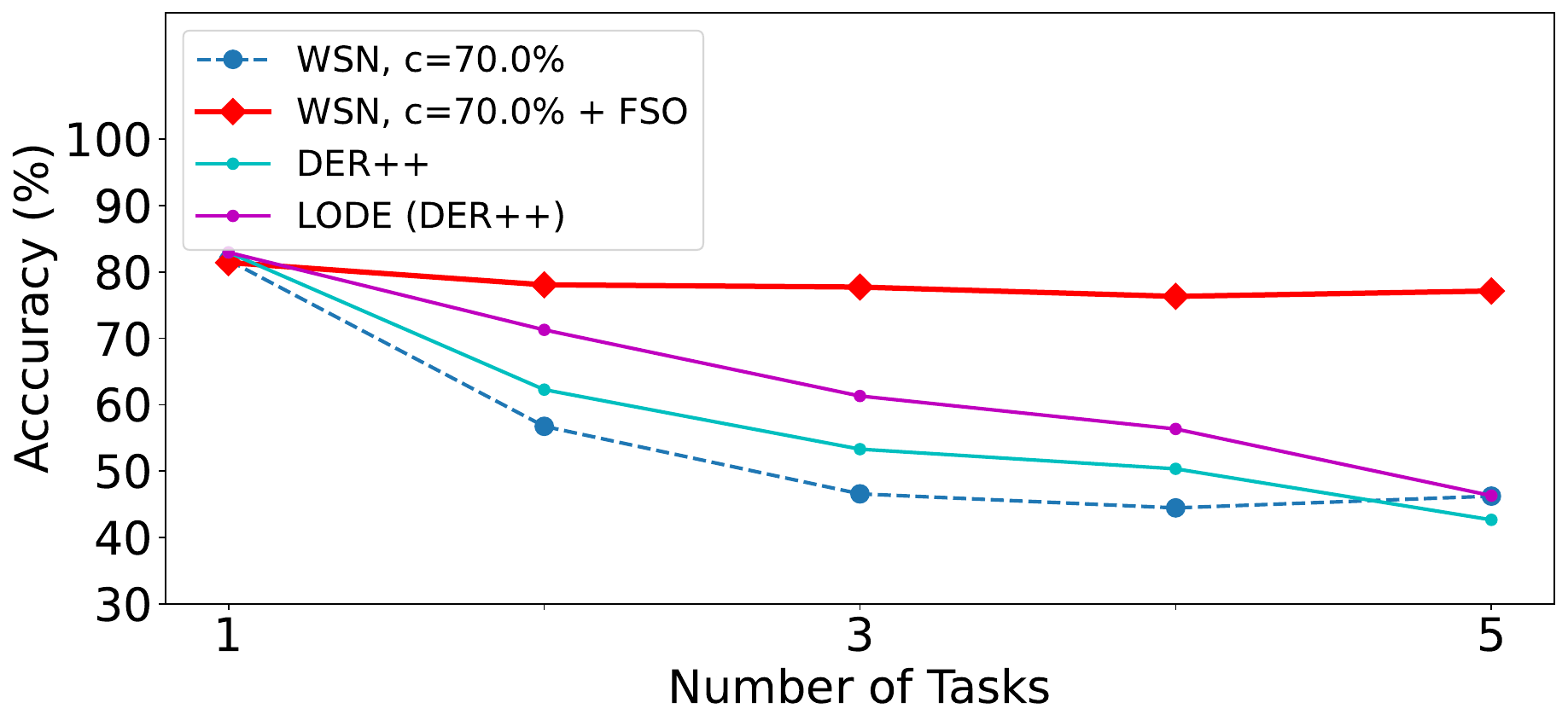} \\
    \vspace{-0.05in}
    \small (a) Seq-CIFAR10 & \small (b) Seq-CIFAR100
    \end{tabular}
    }
    \vspace{-0.05in}
    \caption{\small \textbf{(TaIL), the accuracy on Seq-CIFAR10 and Seq-CIFAR100.} }
    \label{fig:TaIL_cifar}
    \vspace{-0.2in}
\end{figure}

%% file: soft-materials/6_exper/6_main_cifar100_resnet18_5way_5shot_split.tex
\begin{table*}[ht]
\begin{center}
\caption{{\textbf{(FSCIL)}, Classification accuracy of ResNet18 on CIFAR-100 for 5-way 5-shot} incremental learning with the same class split as in TOPIC~\cite{cheraghian2021semantic}. $^\ast$ denotes the results reported from \cite{shi2021overcoming}.}
\vspace{-0.12in}
\resizebox{0.7\textwidth}{!}{
\begin{tabular}{lcccccccccc}
\toprule
\multicolumn{1}{c}{\multirow{2}{*}{{Method}}}&\multicolumn{9}{c}{{sessions}}& \multicolumn{1}{c}{\multirow{2}{*}{\makecell{{The gap} \\ {with cRT}}}} \\ 
\cline{2-10}
& 1 & 2 & 3 & 4 & 5 & 6 & 7 & 8 & 9 &  \\
\midrule
cRT \cite{shi2021overcoming}$^\ast$ & 72.28 & 69.58 & 65.16 & 61.41 & 58.83 & 55.87 & 53.28 & 51.38 & 49.51 \\
\midrule
TOPIC \cite{cheraghian2021semantic} & 64.10 & 55.88 & 47.07 & 45.16 & 40.11 & 36.38 & 33.96 & 31.55 & 29.37 & -20.14 \\
CEC \cite{zhang2021few} & 73.07 & 68.88 & 65.26 & 61.19 & 58.09 & 55.57 & 53.22 & 51.34 & 49.14 & -0.37  \\
F2M \cite{shi2021overcoming} & 71.45 & 68.10 & 64.43 & 60.80 & 57.76 & 55.26 & 53.53 & 51.57 & 49.35 & -0.16 \\ 
LIMIT \cite{zhou2022few} & 73.81 & 72.09 & 67.87 & 63.89 & 60.70 & 57.77 & 55.67 & 53.52 & 51.23 & +1.72\\
MetaFSCIL \cite{chi2022metafscil} & 74.50 & 70.10 & 66.84 & 62.77 & 59.48 & 56.52 &  54.36 & 52.56 & 49.97 & +0.46 \\
ALICE \cite{peng2022few} & 79.00 & 70.50 & 67.10 & 63.40 & 61.20 & 59.20 & 58.10 & 56.30 & 54.10 & +4.59 \\
Entropy-Reg \cite{liu2022few} & 74.40 & 70.20 & 66.54 & 62.51 & 59.71 & 56.58 & 54.52 & 52.39 & 50.14 & +0.63 \\
C-FSCIL \cite{hersche2022constrained} & 77.50 & 72.45 & 67.94 & 63.80 & 60.24 & 57.34 & 54.61 & 52.41 & 50.23 & +0.72 \\
\midrule
FSLL \cite{mazumder2021few} & 64.10 & 55.85 & 51.71 & 48.59 & 45.34 & 43.25 & 41.52 & 39.81 & 38.16 & -11.35 \\
FACT \cite{zhou2022forward} & 74.60 & 72.09 & 67.56 & 63.52 & 61.38 & 58.36 & 56.28 & 54.24 & 52.10 & +2.59 \\

WaRP ~\cite{kim2023warping} & 80.31 & 75.86 & 71.87 & 67.58 & 64.39 & 61.34 & 59.15 & 57.10 & 54.74 & +5.23 \\









\midrule








WSN,~~~~~~$c=90\%$ & 79.22	& 74.77 & 70.89 & 66.41 & 62.90 & 59.48 & 58.10 & 56.13 & 53.92 & +4.41 \\

\textcolor{black}{SoftNet}, ~~$c=90\%$    
& 79.97 & 75.75  & 71.76  & 67.36  & 64.09  & 60.91 &  59.07 & 56.94  & 54.76  & +5.25 \\







\textcolor{black}{SoftNet}, ~~$c=90\%$ + \textbf{FSO}    
& \textbf{80.40} & \textbf{76.06}  & \textbf{72.43}  & \textbf{68.43}  & \textbf{65.54}  & \textbf{62.27} &  \textbf{60.13} & \textbf{58.15}  & \textbf{56.00}  & +\textbf{6.49} \\

\bottomrule
\end{tabular}
}
\vspace{-0.15in}
\label{tab:main_cifar100_5way_5shot_resnet18_split}
\end{center}
\end{table*}

%% file: soft-materials/6_exper/6_main_miniImageNet_5way_5shot_split.tex
\begin{table*}[ht]
\begin{center}
\caption{\textbf{(FSCIL)}, {Classification accuracy of ResNet18 on miniImageNet for 5-way 5-shot} incremental learning with the same class split as in TOPIC~\cite{cheraghian2021semantic}. $^\ast$ denotes results reported from \cite{shi2021overcoming}.}
\vspace{-0.12in}
\resizebox{0.7\textwidth}{!}{
\begin{tabular}{lcccccccccc}
\toprule
\multicolumn{1}{c}{\multirow{2}{*}{{Method}}}&\multicolumn{9}{c}{{sessions}}& \multicolumn{1}{c}{\multirow{2}{*}{\makecell{{The gap} \\ {with cRT}}}} \\ 
\cline{2-10}
& 1 & 2 & 3 & 4 & 5 & 6 & 7 & 8 & 9 &  \\
\midrule
cRT \cite{shi2021overcoming}$^\ast$ & 72.08 & 68.15 & 63.06 & 61.12 & 56.57 & 54.47 & 51.81 & 49.86 & 48.31 & -\\
\midrule
TOPIC \cite{cheraghian2021semantic} & 61.31 & 50.09 & 45.17 & 41.16 & 37.48 & 35.52 & 32.19 & 29.46 & 24.42 & -23.89 \\
IDLVQ-C \cite{chen2020incremental} & 64.77 & 59.87 & 55.93 & 52.62 & 49.88 & 47.55 & 44.83 & 43.14 & 41.84 & -6.47 \\
CEC \cite{zhang2021few} &  72.00 & 66.83 & 62.97 & 59.43 & 56.70 & 53.73 & 51.19 & 49.24 & 47.63 & -0.68 \\
F2M \cite{shi2021overcoming} & 72.05 & 67.47 & 63.16 & 59.70 & 56.71 & 53.77 & 51.11 & 49.21 & 47.84 & -0.43 \\ 
LIMIT \cite{zhou2022few} & 73.81 & 72.09 & 67.87 & 63.89 & 60.70 & 57.77 & 55.67 & 53.52 & 51.23 & +2.92  \\
MetaFSCIL \cite{chi2022metafscil} & 72.04 & 67.94 & 63.77 & 60.29 & 57.58 & 55.16 & 52.90 & 50.79 & 49.19 & +0.88 \\
ALICE \cite{peng2022few} & \textbf{80.60} & 70.60 & 67.40 & 64.50 & 62.50 & 60.00 & 57.80 & \textbf{56.80} & \textbf{55.70} & +\textbf{7.39}  \\
C-FSCIL \cite{hersche2022constrained} & 76.40 & 71.14 & 66.46 & 63.29 & 60.42 & 57.46 & 54.78 & 53.11 & 51.41 & +3.10 \\
Entropy-Reg \cite{liu2022few} & 71.84 & 67.12 & 63.21 & 59.77 & 57.01 & 53.95 & 51.55 & 49.52 & 48.21 & -0.10 \\
Subspace Reg. \cite{akyurek2021subspace} &  80.37 & 71.69 & 66.94 & 62.53 & 58.90 & 55.00 & 51.94 & 49.76 & 46.79 & -1.52\\
\midrule 
FSLL \cite{mazumder2021few} & 66.48 & 61.75 & 58.16 & 54.16 & 51.10 & 48.53 & 46.54 & 44.20 & 42.28 & -6.03 \\
FACT \cite{zhou2022forward} & 72.56 & 69.63 & 66.38 & 62.77 & 60.60 & 57.33 & 54.34 & 52.16 & 50.49 & +2.18 \\
WaRP \cite{kim2023warping} & 72.99 & 68.10 & 64.31 & 61.30 & 58.64 & 56.08 & 53.40 & 51.72 & 50.65 & +2.34 \\

\midrule 



WSN, ~~~~ $c=85\%$ & 78.60 & 72.55 & 68.26 & 64.45 & 61.74 & 58.93 & 55.99 & 54.09 & 52.74 & +4.43 \\

\textcolor{black}{SoftNet}, ~~$c=85\%$  
& 79.50 & 74.54 & 70.29  & 66.39 & 63.35 & 60.38 & 57.32  & 55.22  & 53.92  & +5.61 \\



\textcolor{black}{SoftNet}, ~~$c=85\%$ + \textbf{FSO}  
& 79.72 & \textbf{74.72}  & \textbf{70.73}  & \textbf{66.88} &  \textbf{64.05}  & \textbf{61.82}  & \textbf{58.03}  & {56.01}  & {54.80}  & +{6.49} \\





\bottomrule
\end{tabular}}
\vspace{-0.15in}
\label{tab:miniImageNet_5way_5shot_baseline_split}
\end{center}
\end{table*}

%% file: soft-materials/6_exper/5_main_cub_10way_5shot.tex
\begin{table*}[h]
\begin{center}
\caption{\textbf{(FSCIL)}, {Classification accuracy of ResNet18 on CUB-200-2011 for 10-way 5-shot} incremental learning (TOPIC class split~\cite{tao2020few}). $^\ast$ denotes results reported from \cite{shi2021overcoming}.}
\vspace{-0.1in}
\resizebox{0.8\textwidth}{!}{
\begin{tabular}{lcccccccccccc}
\toprule
\multicolumn{1}{c}{\multirow{2}{*}{{Method}}}&\multicolumn{11}{c}{{sessions}}& \multicolumn{1}{c}{\multirow{2}{*}{\thead{{The gap} \\ {with cRT}}}} \\ 
\cline{2-12}
& 1 & 2 & 3 & 4 & 5 & 6 & 7 & 8 & 9 & 10 & 11   \\
\midrule
cRT \cite{shi2021overcoming}$^\ast$ & 77.16 & 74.41 & 71.31 & 68.08 & 65.57 & 63.08 & 62.44 & 61.29 & 60.12 & 59.85 & 59.30 & - \\
\midrule
TOPIC \cite{cheraghian2021semantic} & 68.68 & 62.49 & 54.81 & 49.99 & 45.25 & 41.40 & 38.35 & 35.36 & 32.22 & 28.31 & 26.28 & -34.80 \\
SPPR \cite{zhu2021self} &  68.68 & 61.85 & 57.43 & 52.68 & 50.19 & 46.88 & 44.65 & 43.07 & 40.17 & 39.63 & 37.33 & -21.97 \\
CEC \cite{zhang2021few} & 75.85 & 71.94 & 68.50 & 63.50 & 62.43 & 58.27 & 57.73 & 55.81 & 54.83 & 53.52 & 52.28 & -7.02 \\
F2M \cite{shi2021overcoming} & 77.13 & 73.92 & 70.27 & 66.37 & 64.34 & 61.69 & 60.52 & 59.38 & 57.15 & 56.94 & 55.89 & -3.41 \\ 
LIMIT \cite{zhou2022few}  & 75.89 & 73.55 & \textbf{71.99} & \textbf{68.14} & \textbf{67.42} & \textbf{63.61} & 62.40 & 61.35 & 59.91 & 58.66 & 57.41 & -1.89 \\
MetaFSCIL \cite{chi2022metafscil} & 75.90 & 72.41 & 68.78 & 64.78 & 62.96 & 59.99 & 58.30 & 56.85 & 54.78 & 53.82 & 52.64 & -6.66 \\

ALICE \cite{peng2022few} & 77.40 & 72.70 & 70.60 & 67.20 & 65.90 & 63.40 & \textbf{62.90} & \textbf{61.90} & \textbf{60.50} & \textbf{60.60} & \textbf{60.10} & \textbf{-0.02} \\
Entropy-Reg \cite{liu2022few} & 75.90 & 72.14 & 68.64 & 63.76 & 62.58 & 59.11 & 57.82 & 55.89 & 54.92 & 53.58 & 52.39 & -6.91 \\
\midrule
FSLL \cite{mazumder2021few} & 72.77 & 69.33 & 65.51 & 62.66 & 61.10 & 58.65 & 57.78 & 57.26 & 55.59 & 55.39 & 54.21 & -6.87 \\

FACT \cite{zhou2022forward} & 75.90 & 73.23 & 70.84 & 66.13 & 65.56 & 62.15 & 61.74 & 59.83 & 58.41 & 57.89 & 56.94 & -2.36 \\

WaRP \cite{kim2023warping} & 77.74 & 74.15 & 70.82 & 66.90 & 65.01 & 62.64 & 61.40 & 59.86 & 57.95 & 57.77 & 57.01 & -2.29 \\

\midrule


WSN,~~~~~$c=90\%$ & 77.23 & 73.62 & 70.20 & 66.36 &	64.32 & 61.40 &	59.86 & 58.28 &	56.36 &	55.88 &	55.30 &	-4.00 \\

\textcolor{black}{SoftNet}, ~$c=90\%$ & 78.07 & 74.58 & 71.37 & 67.54 & 65.37 & 62.60 & 61.07 &	59.37 &	57.53 &	57.21 &	56.75 &	-2.55 \\

\textcolor{black}{SoftNet}, ~$c=90 \%$ + \textbf{FSO} & \textbf{78.24} & \textbf{74.73} & 71.37 & 67.54 & 65.54 & 62.80 & 61.92 & 59.54 & 57.86 & 57.72 & 56.84 & -2.46 \\
\bottomrule
\end{tabular}
}
\label{tab:main_cub200_10way_5shot}
\end{center}
\vspace{-0.15in}
\end{table*}

%% file: FSO-materials/main_table_imagenet.tex
\begin{table}[!ht]
\small
\centering
\vspace{-0.05in}
\caption{\small {Image Classification Performances on ImageNet-1K}.}
\resizebox{0.28\textwidth}{!}{
\begin{tabular}{lcc}
\toprule 

\textbf{Method} & Acc@1 & Acc@5 \\  \midrule 
ResNet18~\cite{he2016deep}     &  69.75           & 89.07            \\ \midrule 
WSN, c=99.0 \%                 &  69.46           & 89.05            \\
WSN, c=99.0 \% + \textbf{FSO}   &  \textbf{70.63}  & \textbf{89.84}   \\ 

\bottomrule
\end{tabular}
}
\label{table:imagnet1k}
\vspace{-0.15in}
\end{table}

%% file: FSO-materials/main_table_uvg17_psnr.tex
\begin{table*}[h]
\small\centering
\caption{\small \textbf{(VIL)}, PSNR results with Fourier Subnueral Operator (FSO) layer (\textbf{$f$-NeRV$\ast$}) on UVG17 Video Sessions with average PSNR and Backward Transfer (BWT) of PSNR. Note that $\ast$ denotes our reproduced results.}
\vspace{-0.1in}
\resizebox{\textwidth}{!}{
\renewcommand{\arraystretch}{1.2}
\begin{tabular}{lccccccccccccccccccc}
\toprule 

\multicolumn{1}{c}{\multirow{2}{*}{\textbf{Method}}} & \multicolumn{17}{c}{\textbf{Video Sessions}} & \multirow{2}{*}{\thead{\textbf{Avg. PSNR} \\ / \textbf{BWT}}} \\ 

\cline{2-18}

& \textbf{1} & \textbf{2} & \textbf{3} & \textbf{4} & \textbf{5} & \textbf{6} & \textbf{7} & \textbf{8} & \textbf{9} & \textbf{10} & \textbf{11} & \textbf{12} & \textbf{13} & \textbf{14} & \textbf{15} & \textbf{16} & \textbf{17} \\ \midrule 
STL, NeRV~\cite{chen2023hnerv} & 39.63 & - & 36.06 & - & 37.35 & - & 41.23 & - & 38.14 & - & 31.86 & - & 37.22 & - & 32.45 & - & - & - / - \\ 

STL, NeRV$^{\ast}$             & 39.66 & 44.89 & 36.28 & 41.13 & 38.14 & 31.53 & 42.03 & 34.74 & 36.58 & 36.85 & 29.22 & 31.81 & 37.27 & 34.18 & 31.45 & 38.41 & 43.86 & 36.94 / - \\ 
\midrule 

EWC~\cite{Kirkpatrick2017}$^{\ast}$ & 11.15 & 9.21  & 12.71 & 11.40 & 15.58  & 9.25 & 7.06 & 12.96 & 6.34 & 10.31 & 9.55 & 13.39 & 5.76 & 8.67 & 10.93 & 10.92 & 28.29 & 11.38 / -16.13 \\ 
iCaRL~\cite{rebuffi2017icarl}$^{\ast}$ & 24.31 & 28.25  & 22.19 & 22.74 & 22.84  & 16.55 & 29.37 & 17.92 & 16.65 & 27.43 & 13.64 & 16.42 & 24.02 & 21.60 & 19.40 & 18.60 & 26.46 & 21.67 / ~~-6.23 \\ 
ESMER~\cite{sarfraz2023error}$^{\ast}$ &  30.77 & 26.33 & 22.79 & 21.35 & 23.76 & 13.64 & 28.25 & 15.22 & 16.71 & 23.78 &	13.35 & 15.23 & 18.21 & 19.22 & 24.59 & 20.61 & 22.42 &  20.95 / -15.23 \\ 
\midrule

WSN$^{\ast}$, c = 30.0 \% & 31.50 & 34.37 & 31.00 & {{32.38}} & {{29.26}} & {{23.08}} & {{31.96}} & {{22.64}} & {{22.07}} & {{33.48}} & {{18.34}} & {{20.45}} & {{27.21}} & {{24.33}} & {{23.09}} & {{21.23}} & {{29.13}} & {{26.80}} / ~0.0 \\ 
WSN$^{\ast}$, c = 50.0 \% & {{34.02}} & {{34.93}} & {{31.04}} & 31.74 & 28.95 & 23.07 & 31.26 & 22.32 & 21.93 & 33.35 & 18.22 & 20.34 & 26.88 & 24.22 & 22.72 & 21.30 & 28.86 & 26.77 / ~0.0 \\ 


\midrule

WSN, c = 30.0 \% + \textbf{$f$-NeRV2} & 32.01 &	35.84 &	32.97 &	35.17 &	31.24 &	24.82 &	36.01 &	25.85 &	24.83 &	35.76 &	20.50 &	22.79 &	30.40 &	27.37 &	25.52 &	25.40 &	32.70 &	29.36 / ~0.0 \\ 


WSN, c = 30.0 \% + \textbf{$f$-NeRV3} &  \textbf{33.64} & \textbf{39.24} & \textbf{34.21} & \textbf{37.79} & \textbf{34.05} & \textbf{27.17} & \textbf{38.17} & \textbf{29.79} & \textbf{26.56} & \textbf{36.18} & \textbf{22.97} & \textbf{24.36} & \textbf{32.50} & \textbf{30.22} &	\textbf{27.62} & \textbf{29.15} & \textbf{35.68} &	\textbf{31.72} / ~\textbf{0.0} \\ 

\midrule

MTL (Upper-bound) & 32.39 & 34.35  & 31.45 & 34.03 & 30.70  & 24.53 & 37.13 & 27.83 & 23.80 & 34.69 & 20.77 & 22.37 & 32.71 & 28.00 & 25.89 & 26.40 & 33.16 & 29.42 / - ~~~ \\ 
\bottomrule
\end{tabular}
}
\vspace{-0.1in}
\label{table:uvg17_fso_psnr}
\end{table*}

%% file: FSO-materials/plot_main_video_mtl.tex
\begin{figure}[h]
    \centering 
    \vspace{-0.1in}
    \setlength{\tabcolsep}{0pt}{%
    \begin{tabular}{cccc}
    
    t=0 & t=1 & t=2 &t=3 \\

    \includegraphics[width=0.25\columnwidth]{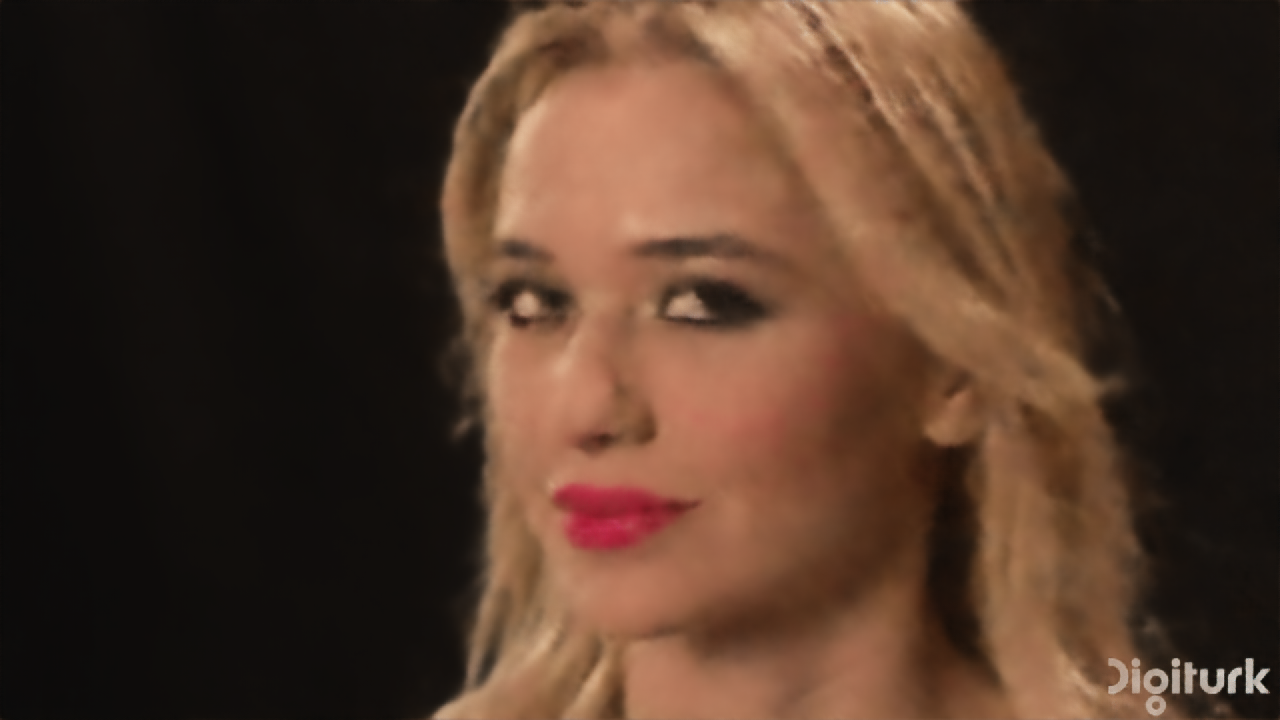} & 
    \includegraphics[width=0.25\columnwidth]{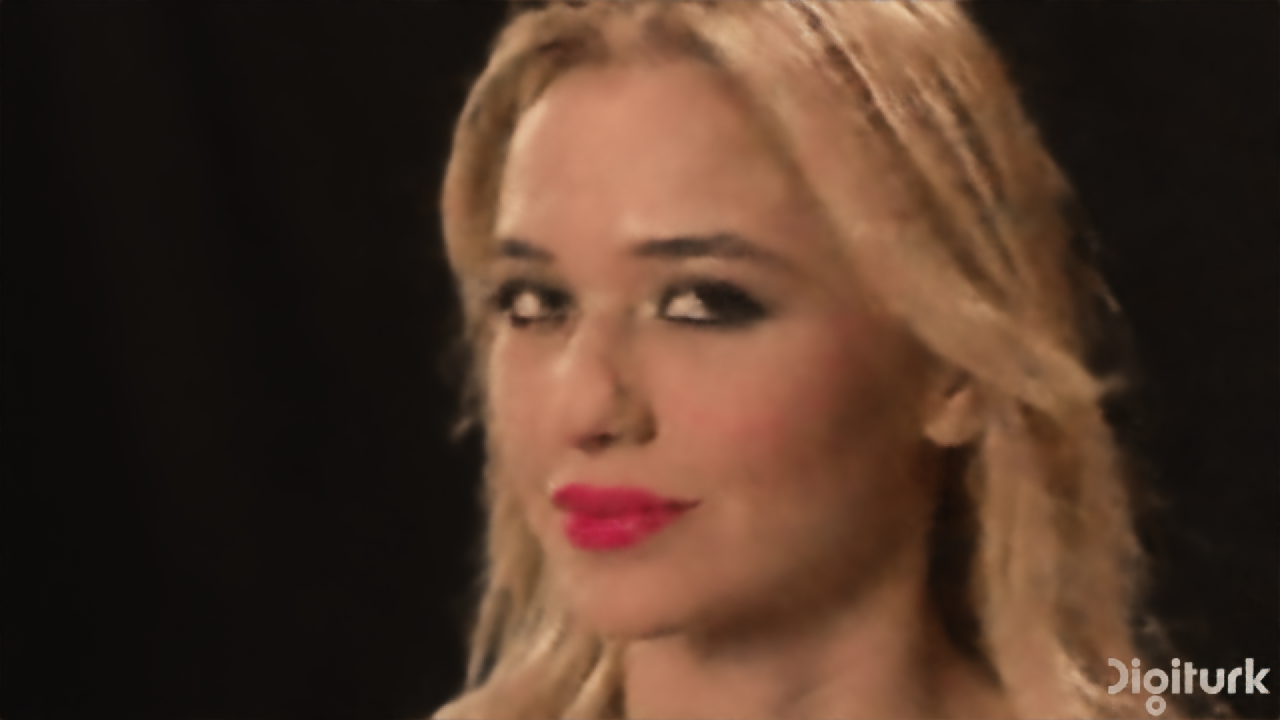} &
    \includegraphics[width=0.25\columnwidth]{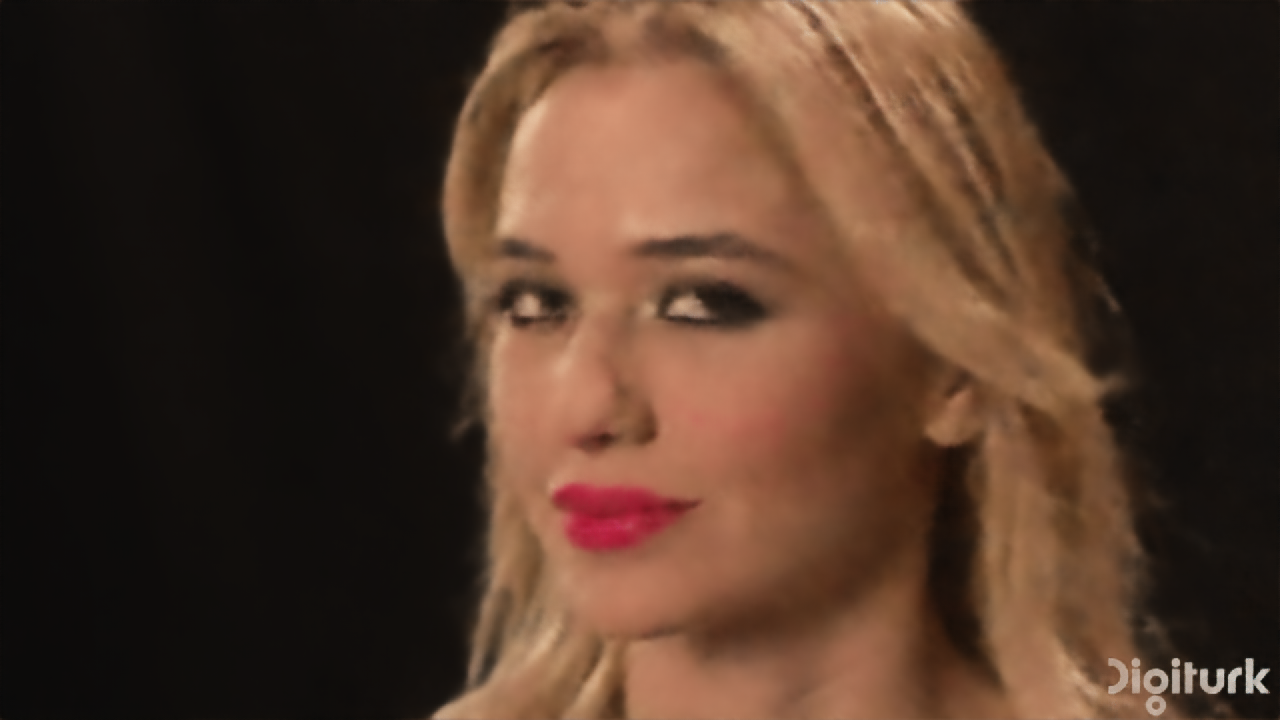} &
    \includegraphics[width=0.25\columnwidth]{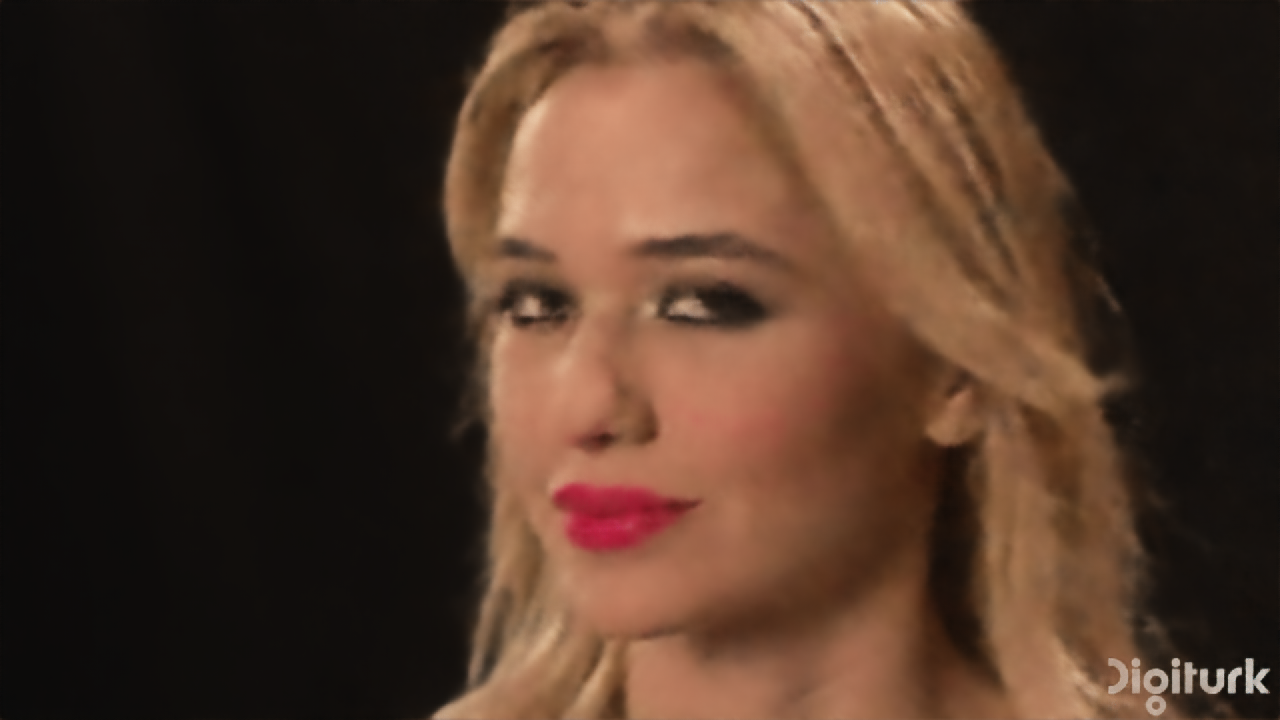} \\ 
    \multicolumn{4}{l}{\small WSN (31.00, PSNR)}\\
    

    \includegraphics[width=0.25\columnwidth]{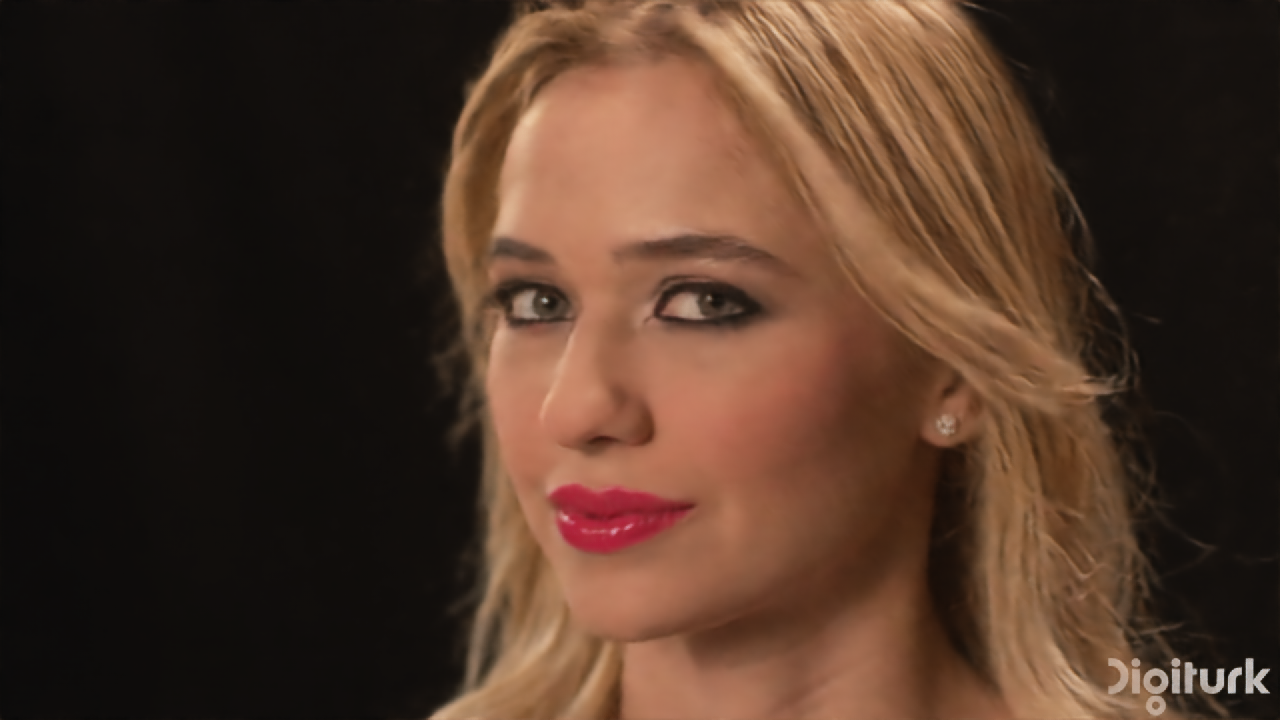} & 
    \includegraphics[width=0.25\columnwidth]{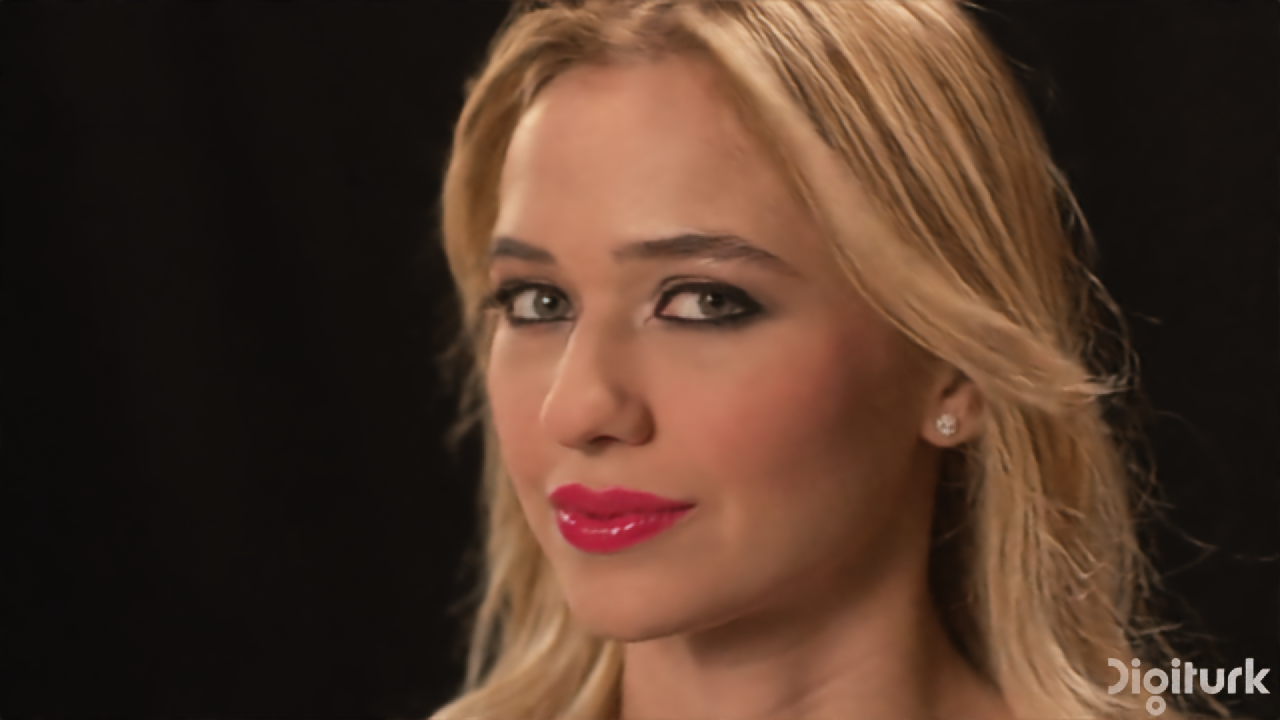} &
    \includegraphics[width=0.25\columnwidth]{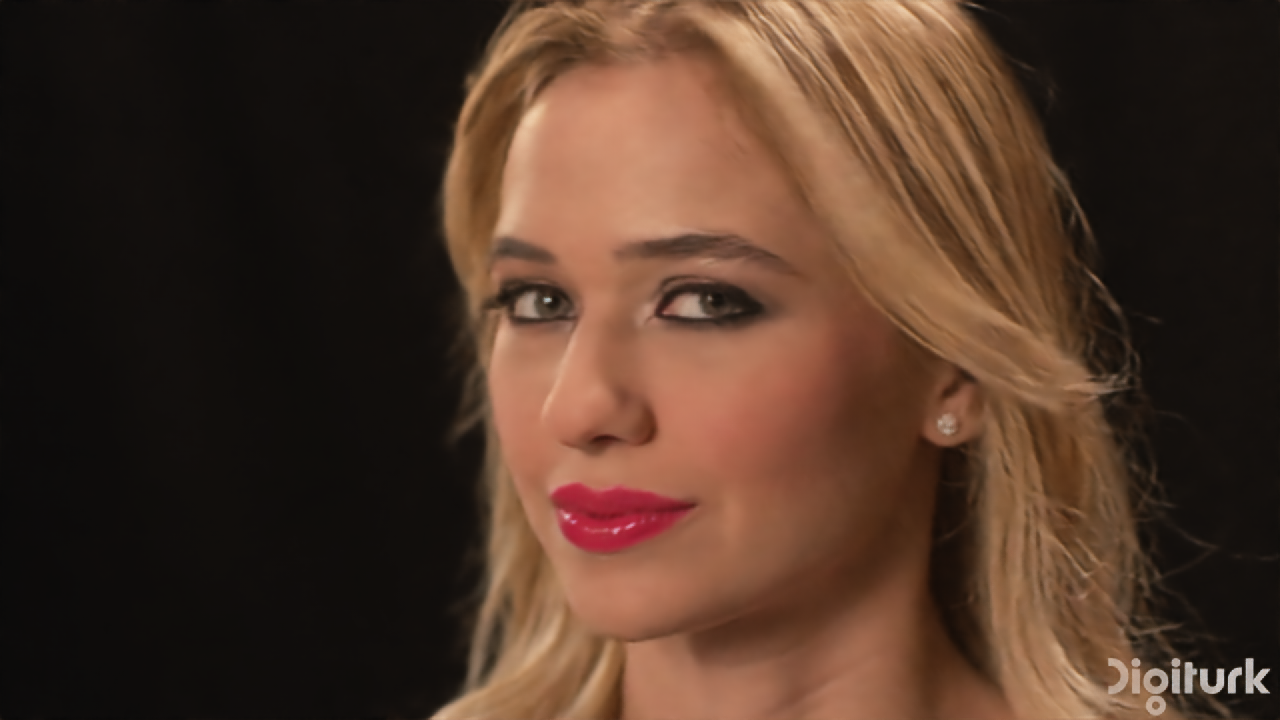} &
    \includegraphics[width=0.25\columnwidth]{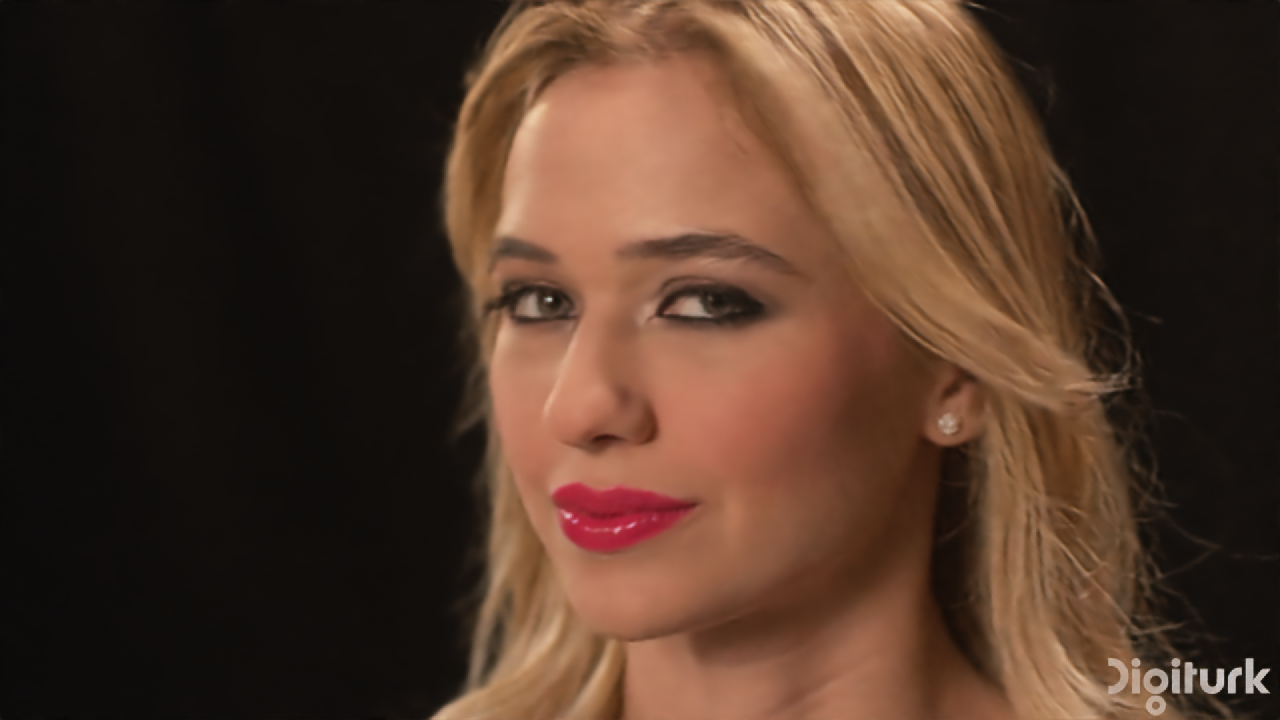} \\ 
    \multicolumn{4}{l}{\small WSN, \textbf{$f$-NeRV3 (34.21, PSNR)}}\\


    \includegraphics[width=0.25\columnwidth]{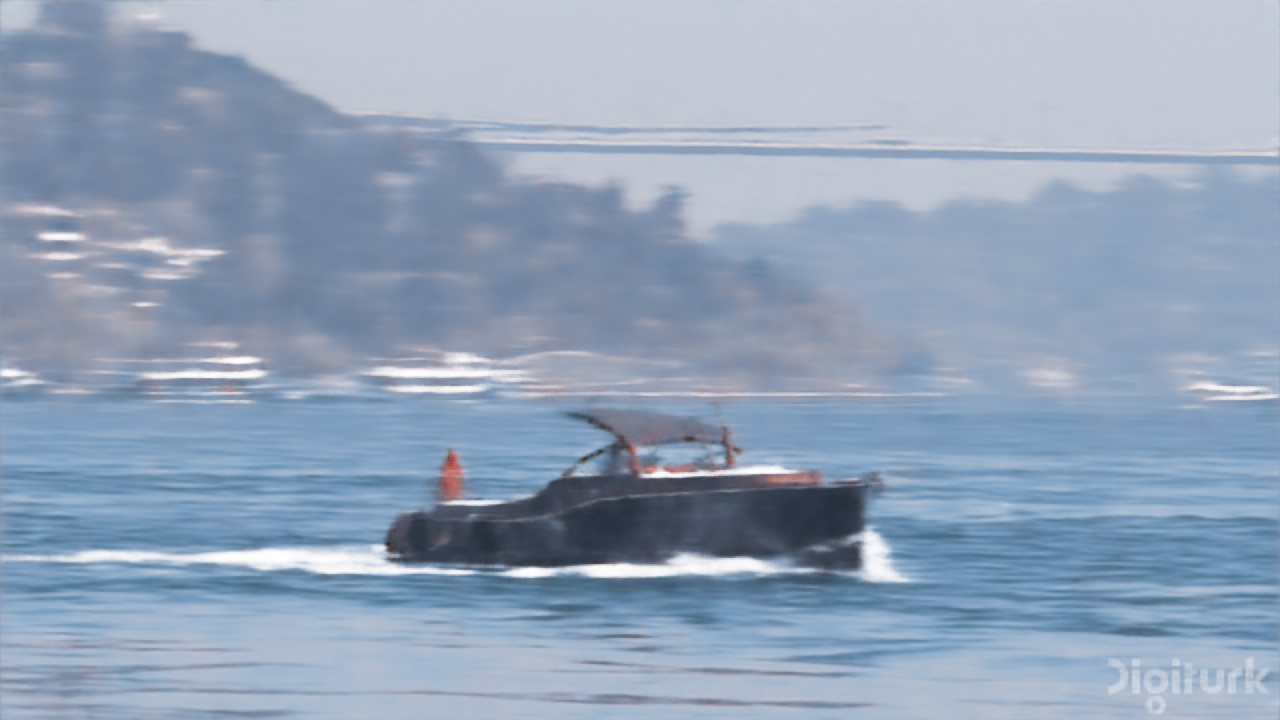} & 
    \includegraphics[width=0.25\columnwidth]{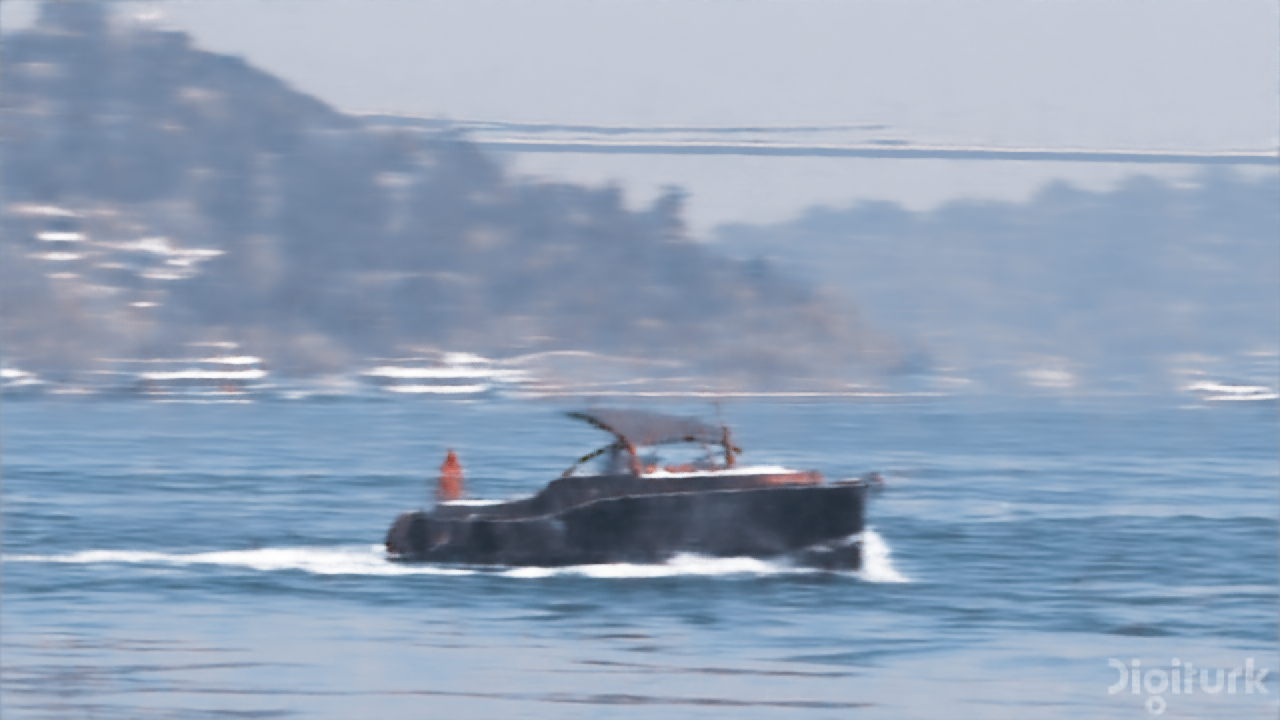} &
    \includegraphics[width=0.25\columnwidth]{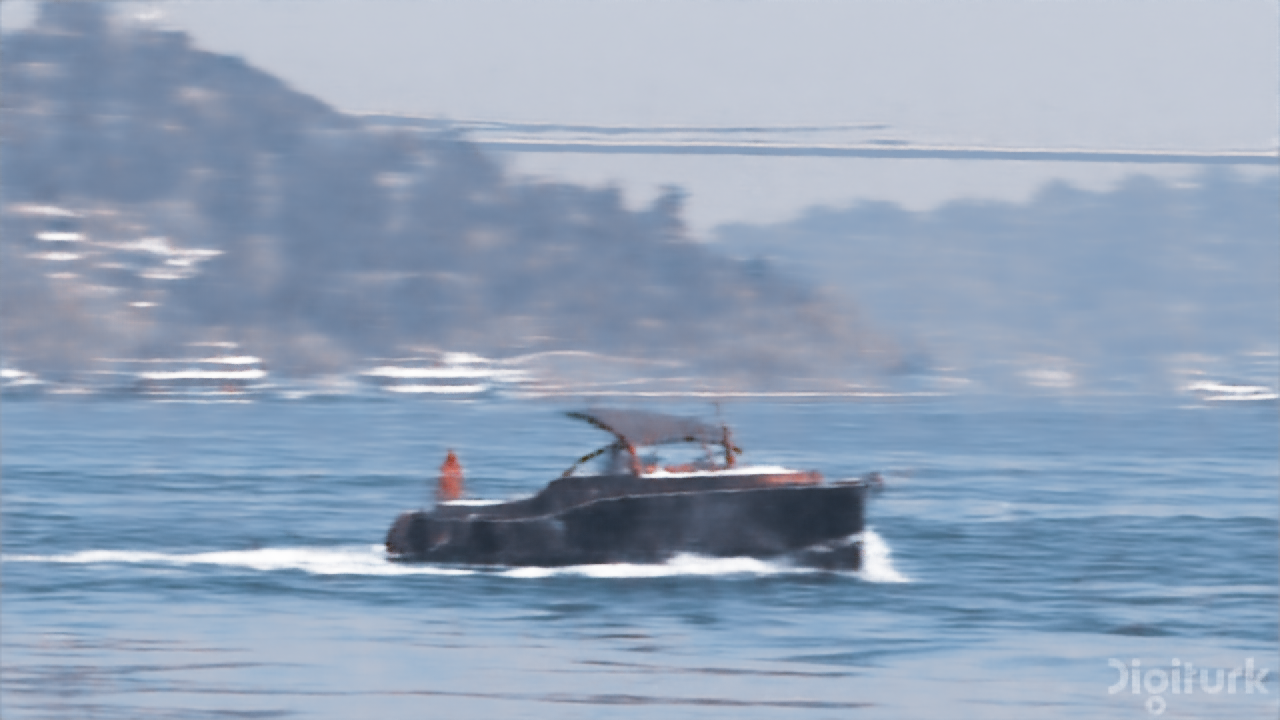} &
    \includegraphics[width=0.25\columnwidth]{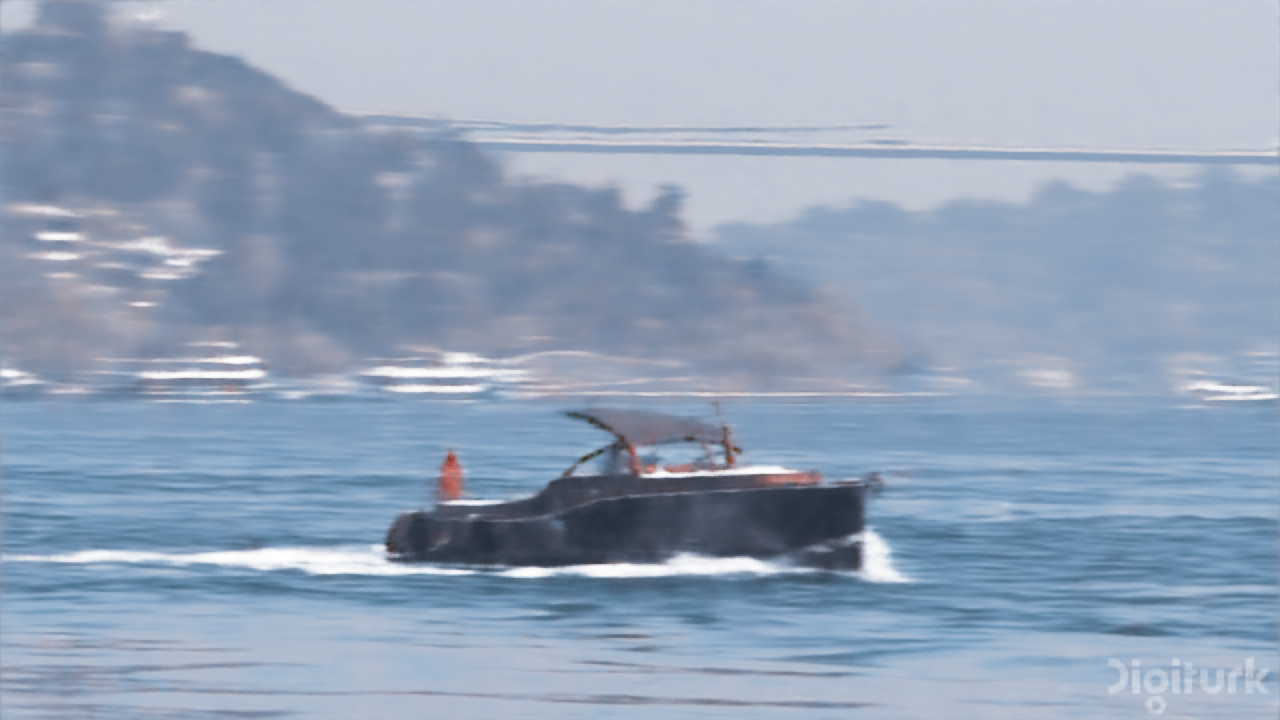} \\ 
     \multicolumn{4}{l}{\small WSN (29.26, PSNR)}\\
    

    \includegraphics[width=0.25\columnwidth]{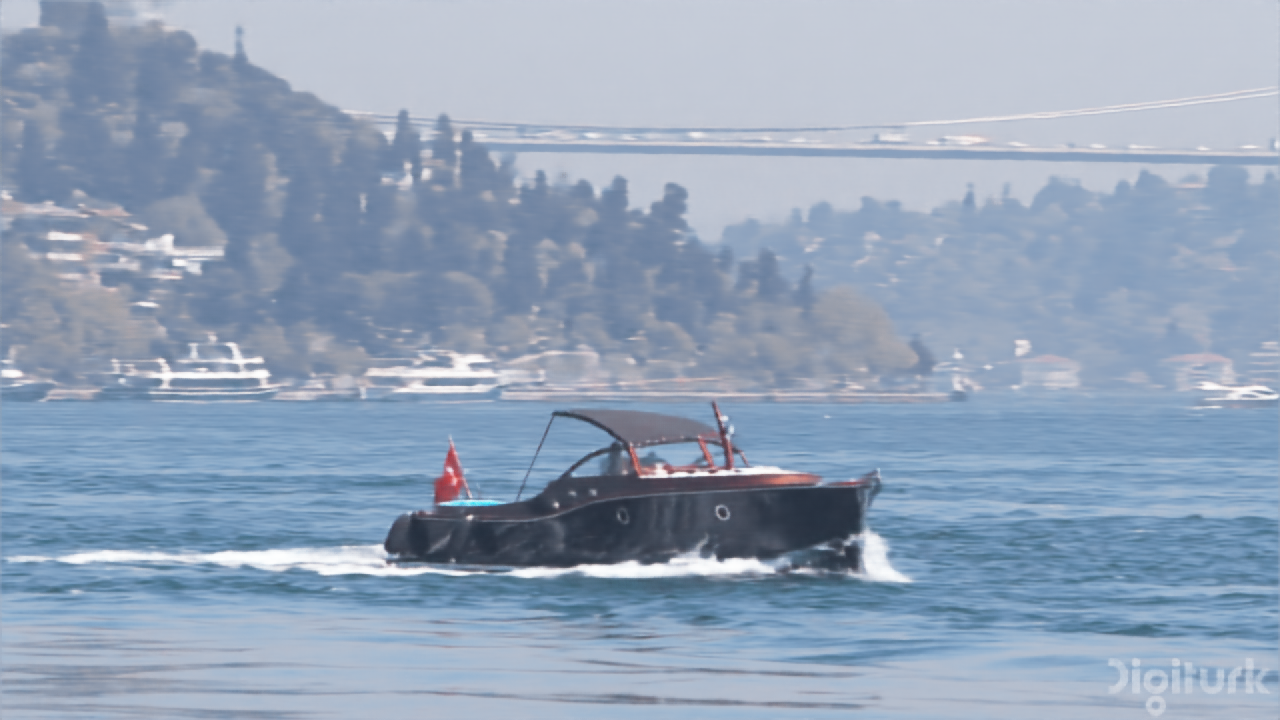} & 
    \includegraphics[width=0.25\columnwidth]{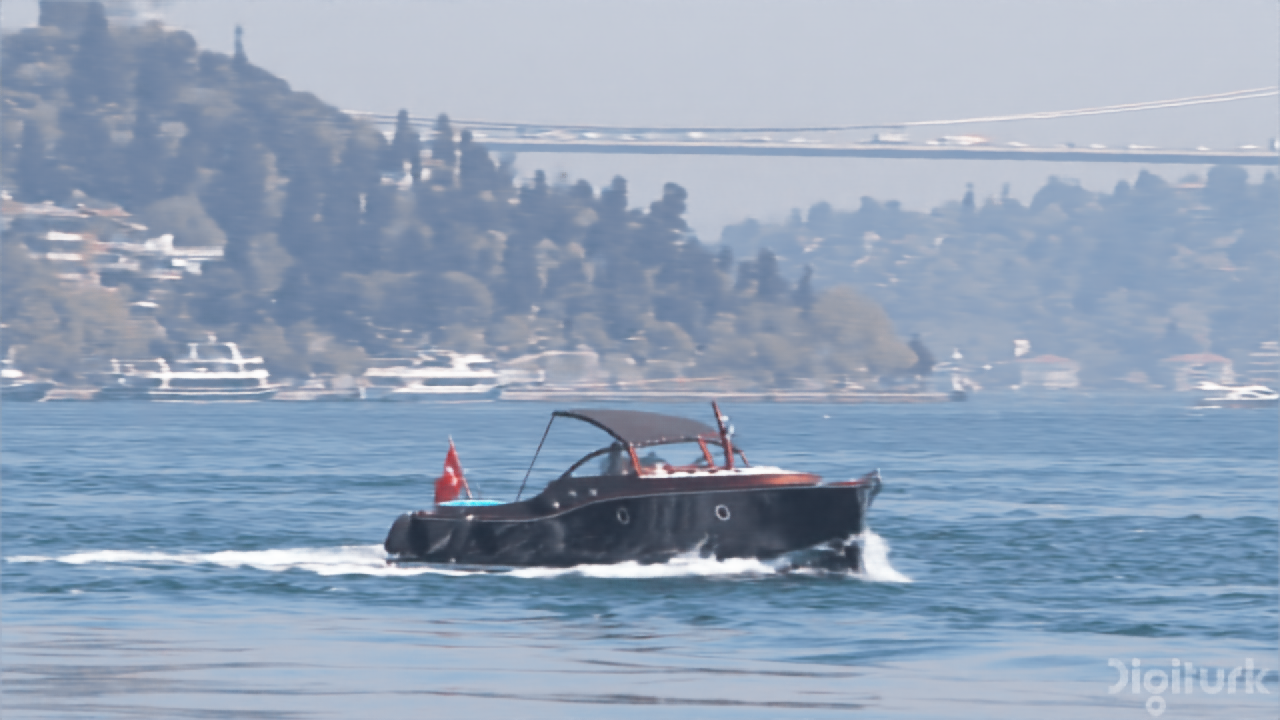} &
    \includegraphics[width=0.25\columnwidth]{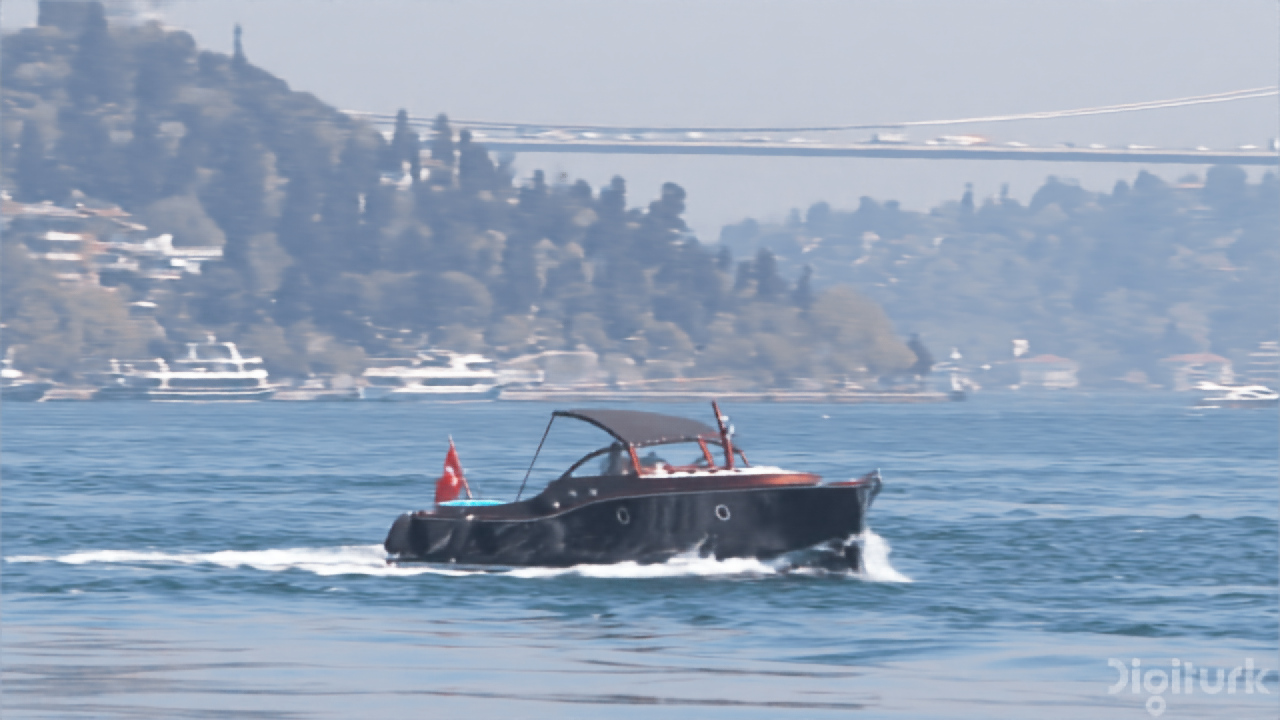} &
    \includegraphics[width=0.25\columnwidth]{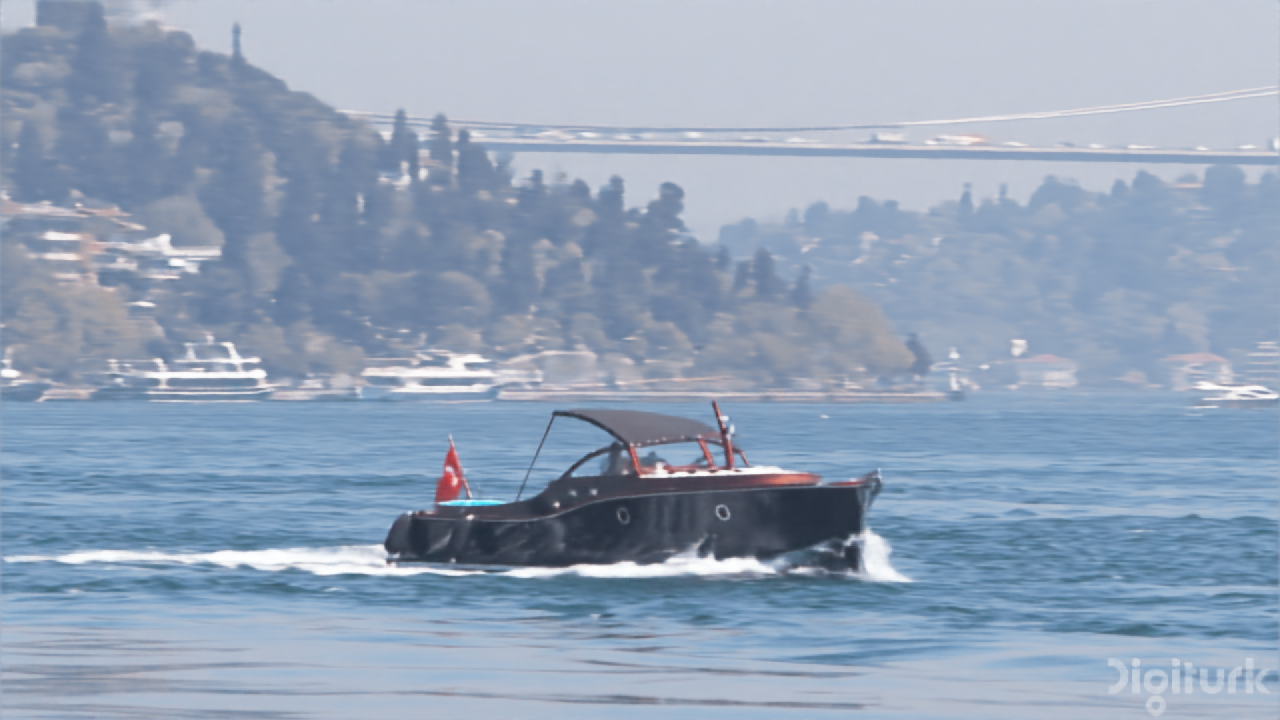} \\ 
    \multicolumn{4}{l}{\small WSN, \textbf{$f$-NeRV3 (34.05, PSNR)}}\\
    
    \end{tabular}
    }
    \vspace{-0.05in}
    \caption{\small \textbf{(VIL), Video Generation} (from t=0 to t=3) with $c = 30.0 \%$ on the UVG17 dataset.}
    \label{fig:video_reinit_mtl}
    \vspace{-0.1in}
\end{figure}

%% file: FSO-materials/plot_main_fmap_uvg17.tex
\begin{figure}[h]
    \centering 
    \vspace{-0.05in}
    \setlength{\tabcolsep}{0pt}{%
    \begin{tabular}{ccccc}
    
     NeRV3 & NeRV4 & NeRV5 & NeRV6 & Head \\ 

    \includegraphics[width=0.20\columnwidth, trim={0.1cm 0.4cm 0.1cm 0.1cm}, clip]{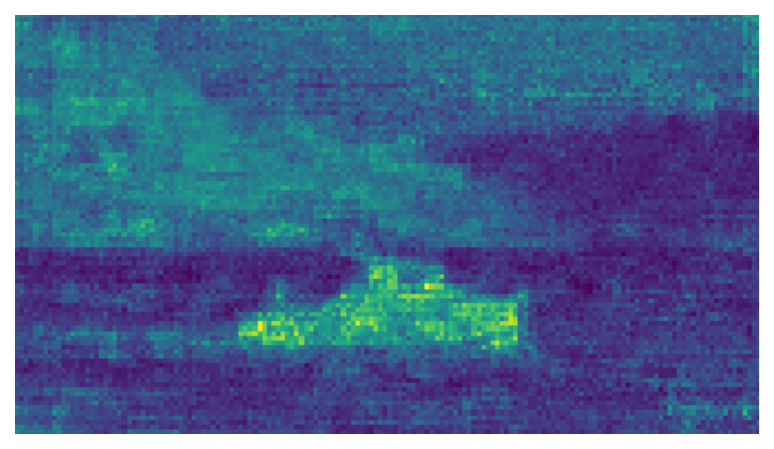} & 
    \includegraphics[width=0.20\columnwidth, trim={0.1cm 0.4cm 0.1cm 0.1cm}, clip]{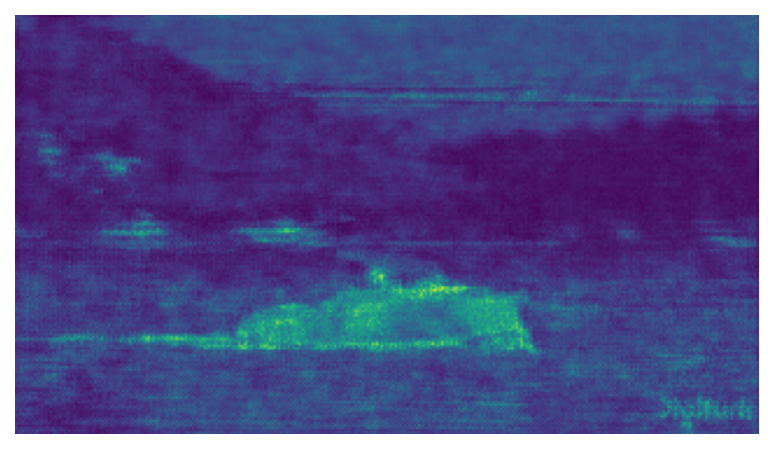} & 
    \includegraphics[width=0.20\columnwidth, trim={0.1cm 0.4cm 0.1cm 0.1cm}, clip]{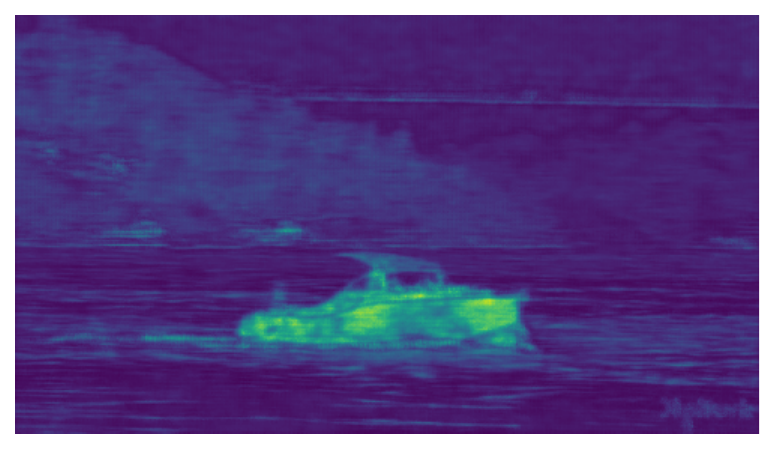} & 
    \includegraphics[width=0.20\columnwidth, trim={0.1cm 0.4cm 0.1cm 0.1cm}, clip]{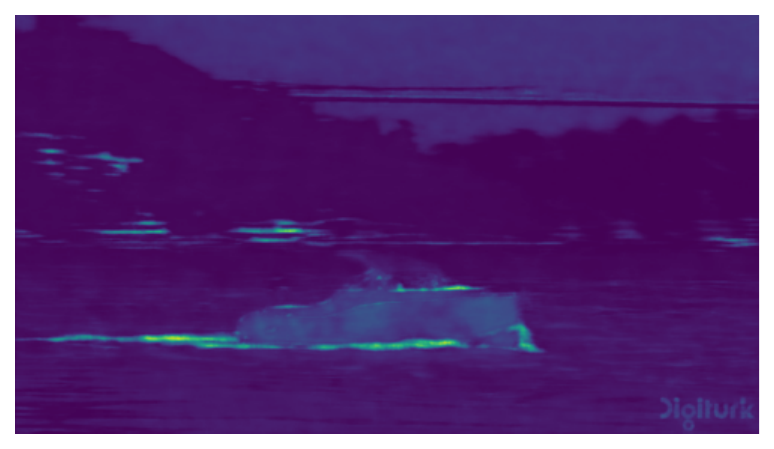} & 
    \includegraphics[width=0.19\columnwidth]{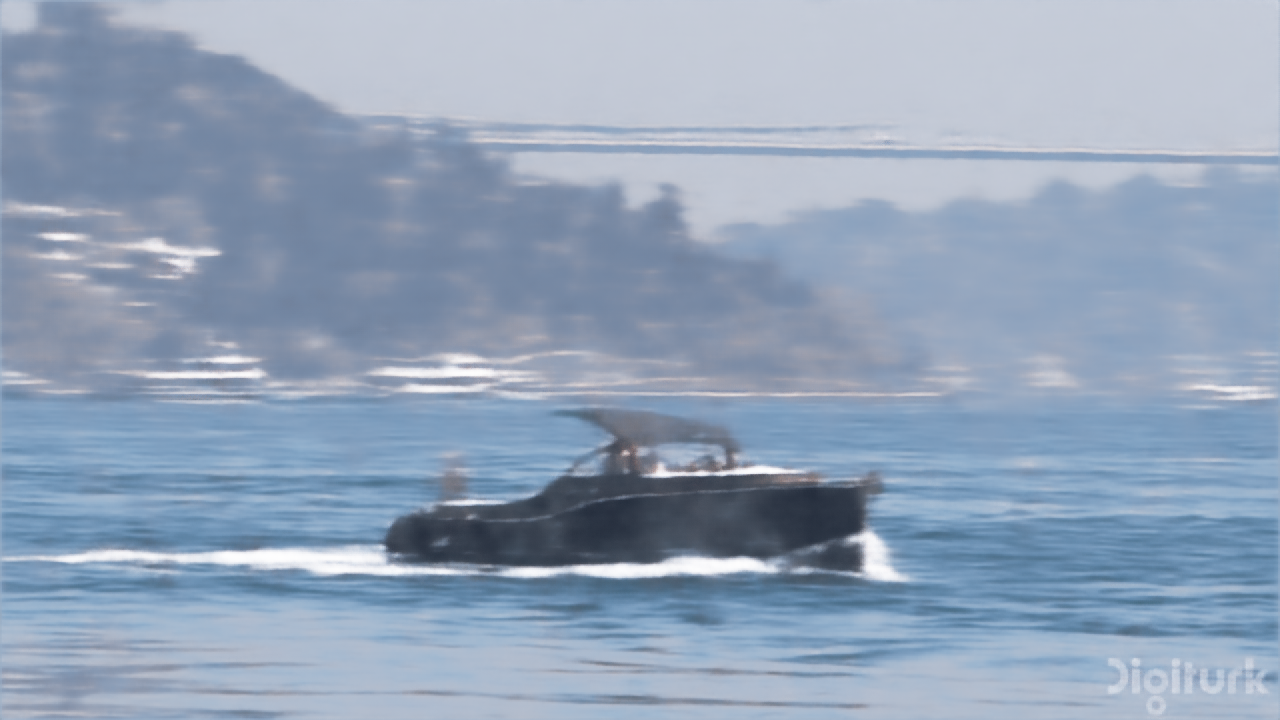} \\

    \multicolumn{5}{l}{\small WSN, c=50.0\% (28.95, PSNR)} \\
    
    \includegraphics[width=0.20\columnwidth, trim={0.1cm 0.4cm 0.1cm 0.1cm}, clip]{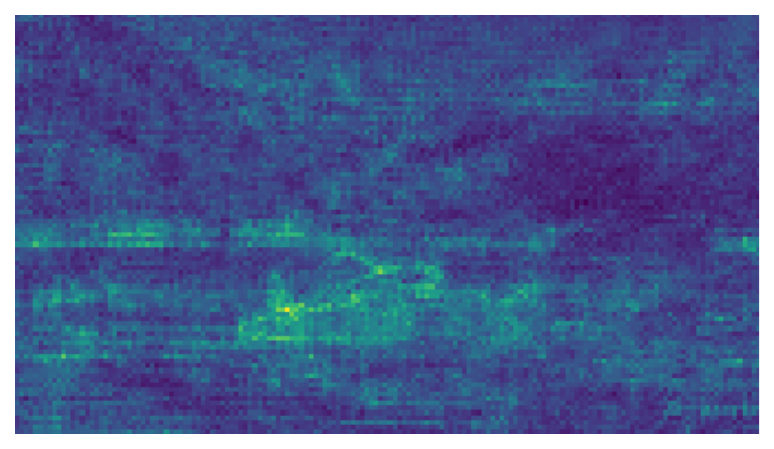} & 
    \includegraphics[width=0.20\columnwidth, trim={0.1cm 0.4cm 0.1cm 0.1cm}, clip]{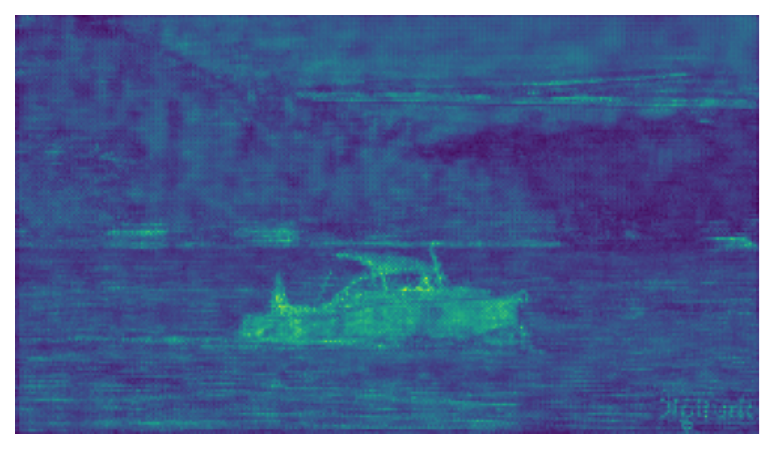} & 
    \includegraphics[width=0.20\columnwidth, trim={0.1cm 0.4cm 0.1cm 0.1cm}, clip]{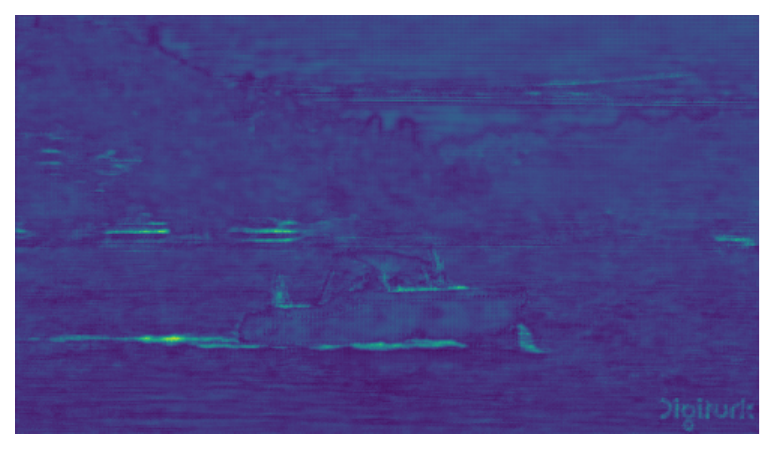} & 
    \includegraphics[width=0.20\columnwidth, trim={0.1cm 0.4cm 0.1cm 0.1cm}, clip]{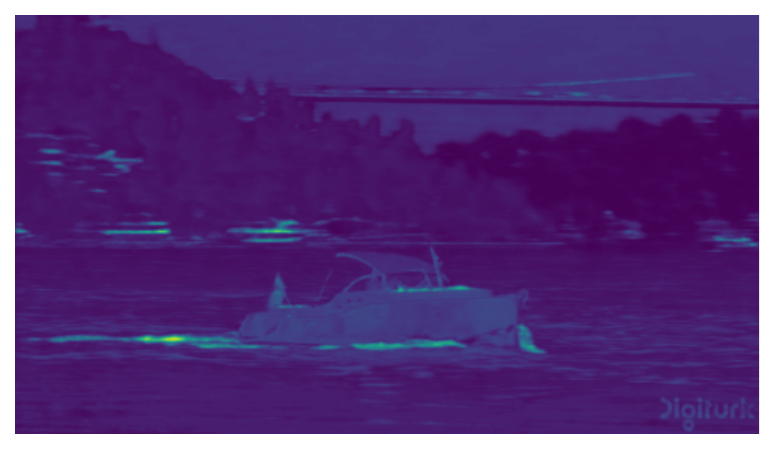} & 
    \includegraphics[width=0.19\columnwidth]{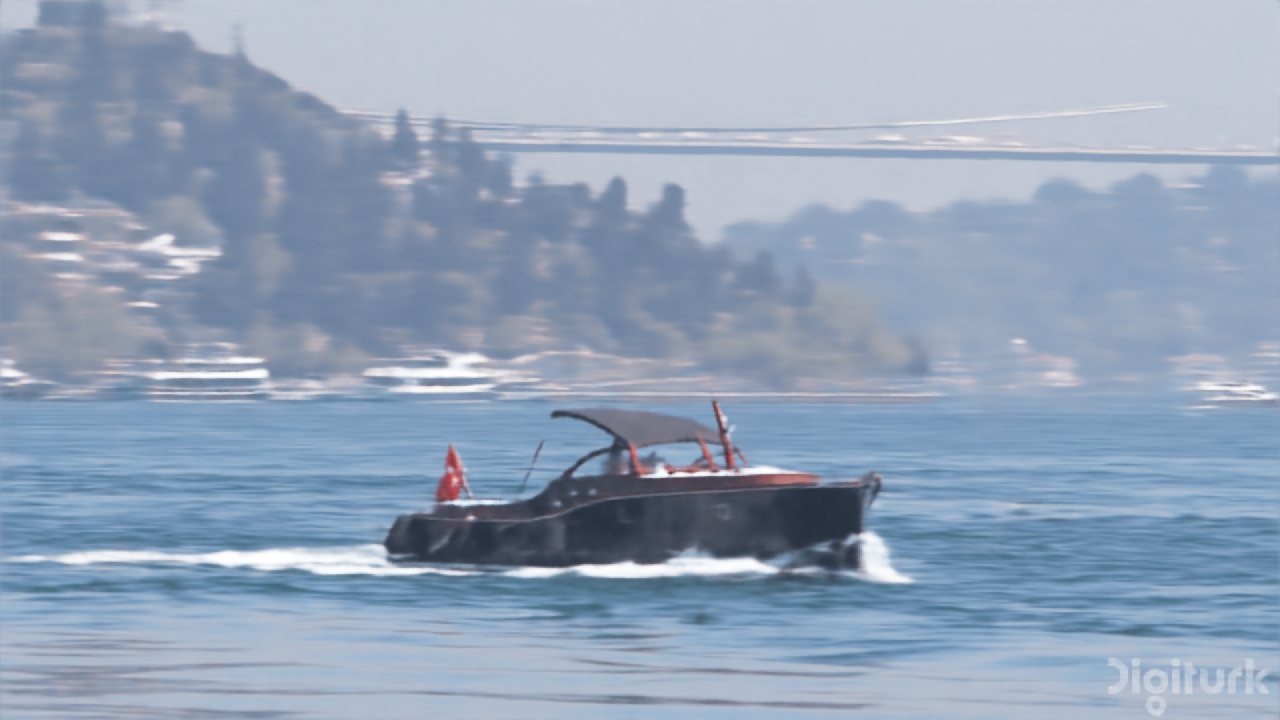} \\

    \multicolumn{5}{l}{\small WSN, c=50.0\%, {$f$-NeRV2 (31.24, PSNR)}} \\

    \includegraphics[width=0.20\columnwidth, trim={0.1cm 0.4cm 0.1cm 0.1cm}, clip]{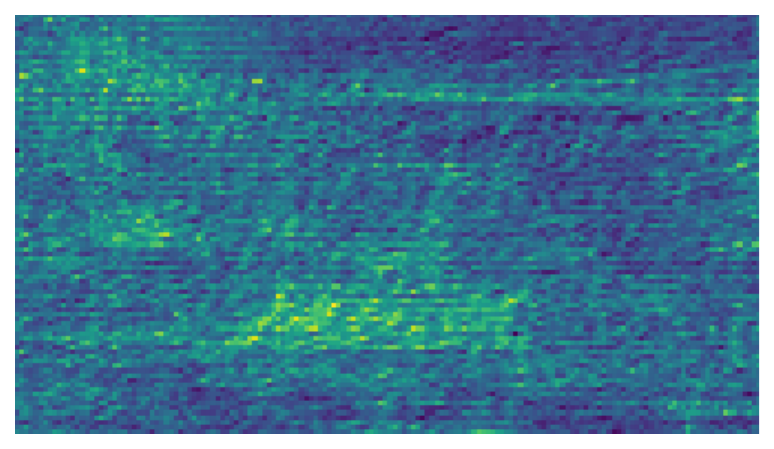} & 
    \includegraphics[width=0.20\columnwidth, trim={0.1cm 0.4cm 0.1cm 0.1cm}, clip]{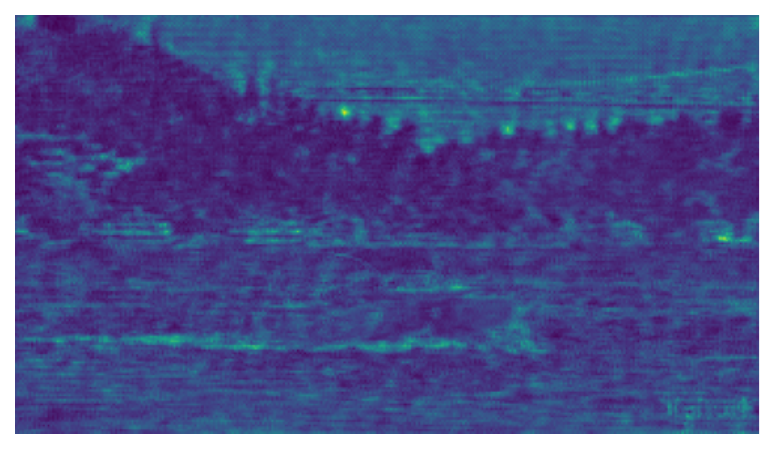} & 
    \includegraphics[width=0.20\columnwidth, trim={0.1cm 0.4cm 0.1cm 0.1cm}, clip]{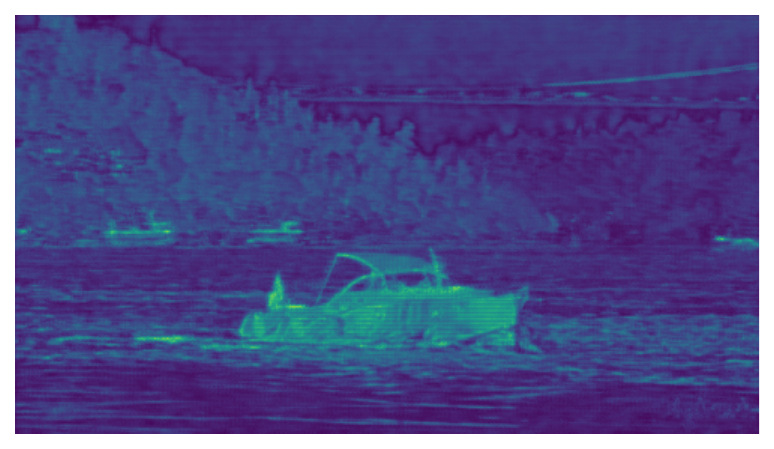} & 
    \includegraphics[width=0.20\columnwidth, trim={0.1cm 0.4cm 0.1cm 0.1cm}, clip]{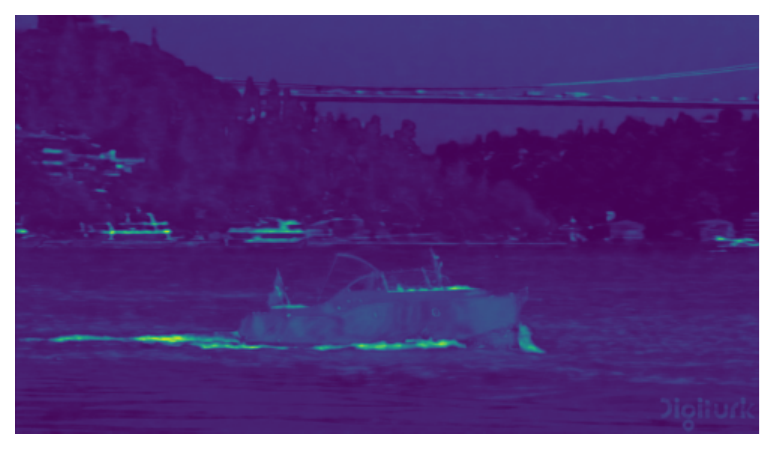} & 
    \includegraphics[width=0.19\columnwidth]{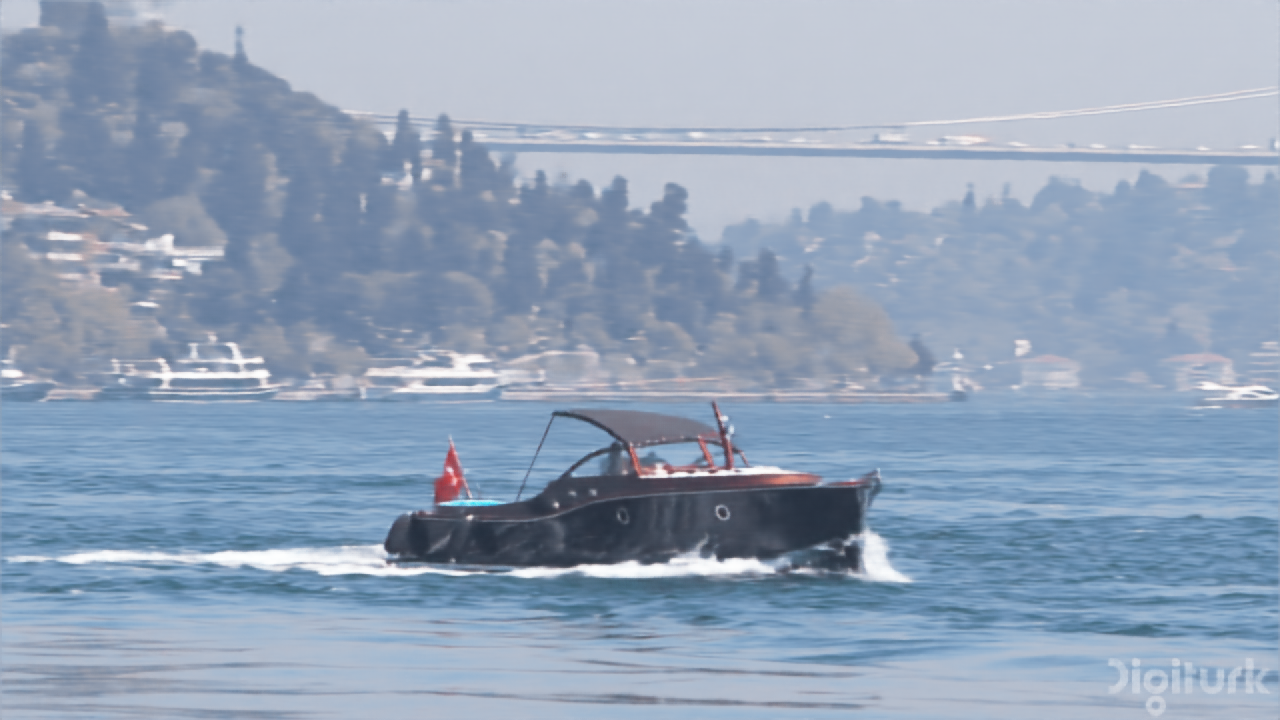} \\

    \multicolumn{5}{l}{\small WSN, c=50.0\%, \textbf{$f$-NeRV3 (34.05, PSNR)}} \\
   
    \end{tabular}
    }
    \vspace{-0.05in}
    \caption{\small \textbf{(VIL), WSN's Representations of NeRV Blocks with $c = 50.0 \%$ on the UVG17 dataset.}}
    \label{fig:fmap_uvg17}
    \vspace{-0.1in}
    
\end{figure}

%% file: FSO-materials/plot_fso_fmap_uvg17.tex
\begin{figure}[ht]
    \centering 
    \setlength{\tabcolsep}{0pt}{%
    \begin{tabular}{ccccc}
    
     NeRV3 & NeRV4 & NeRV5 & NeRV6 & Head \\ 

    \includegraphics[width=0.20\columnwidth, trim={0.1cm 0.4cm 0.1cm 0.1cm}, clip]{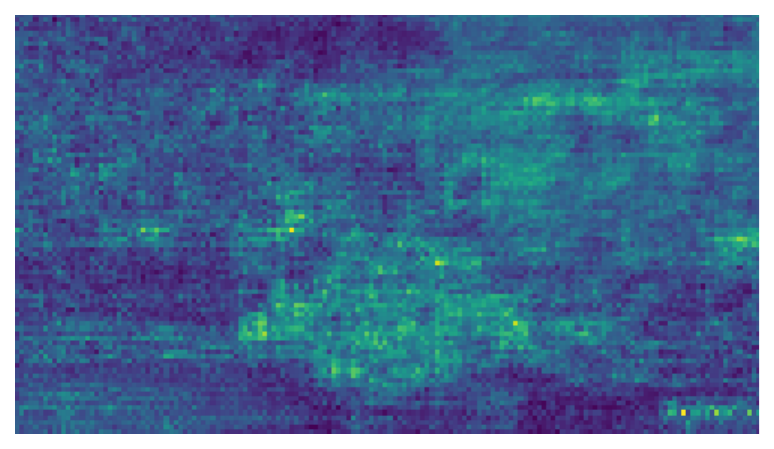} & 
    \includegraphics[width=0.20\columnwidth, trim={0.1cm 0.4cm 0.1cm 0.1cm}, clip]{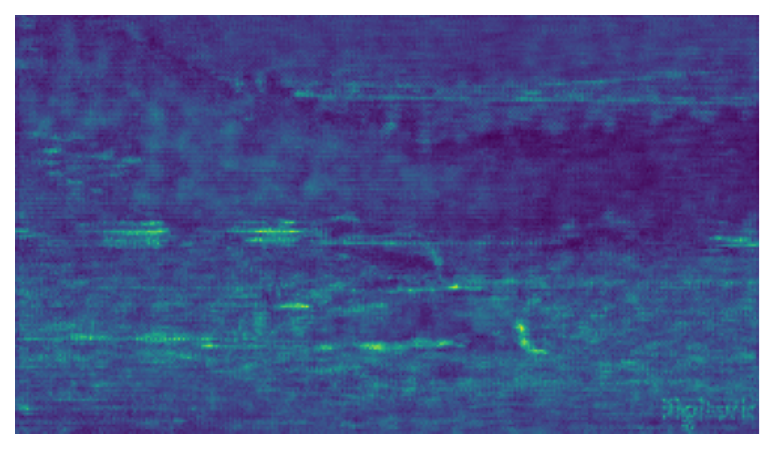} & 
    \includegraphics[width=0.20\columnwidth, trim={0.1cm 0.4cm 0.1cm 0.1cm}, clip]{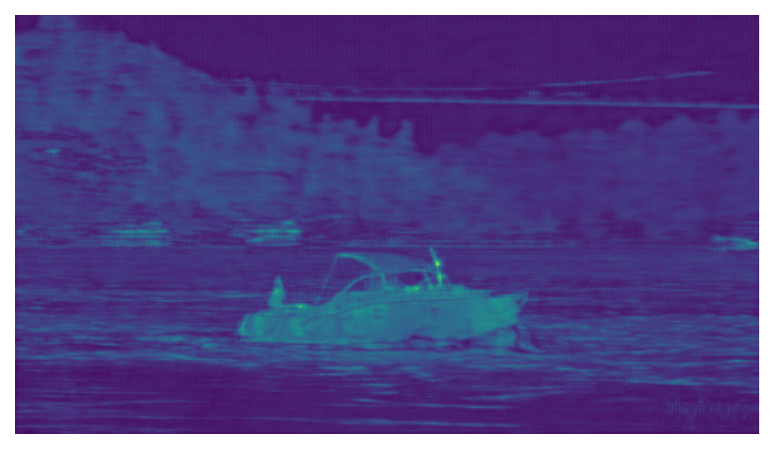} & 
    \includegraphics[width=0.20\columnwidth, trim={0.1cm 0.4cm 0.1cm 0.1cm}, clip]{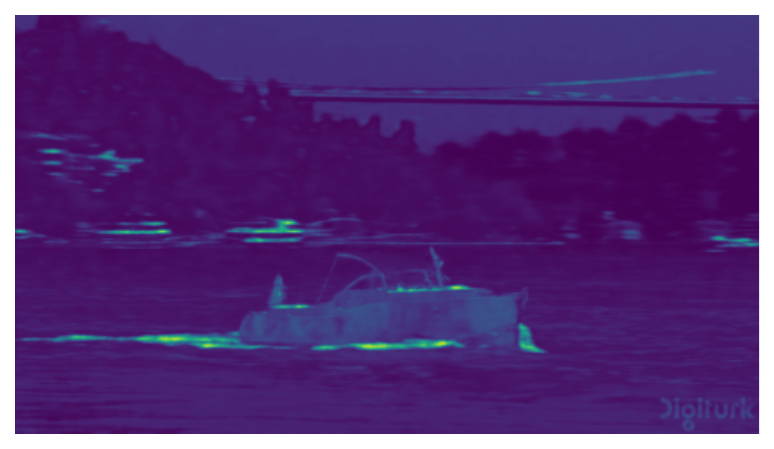} & 
    \includegraphics[width=0.19\columnwidth]{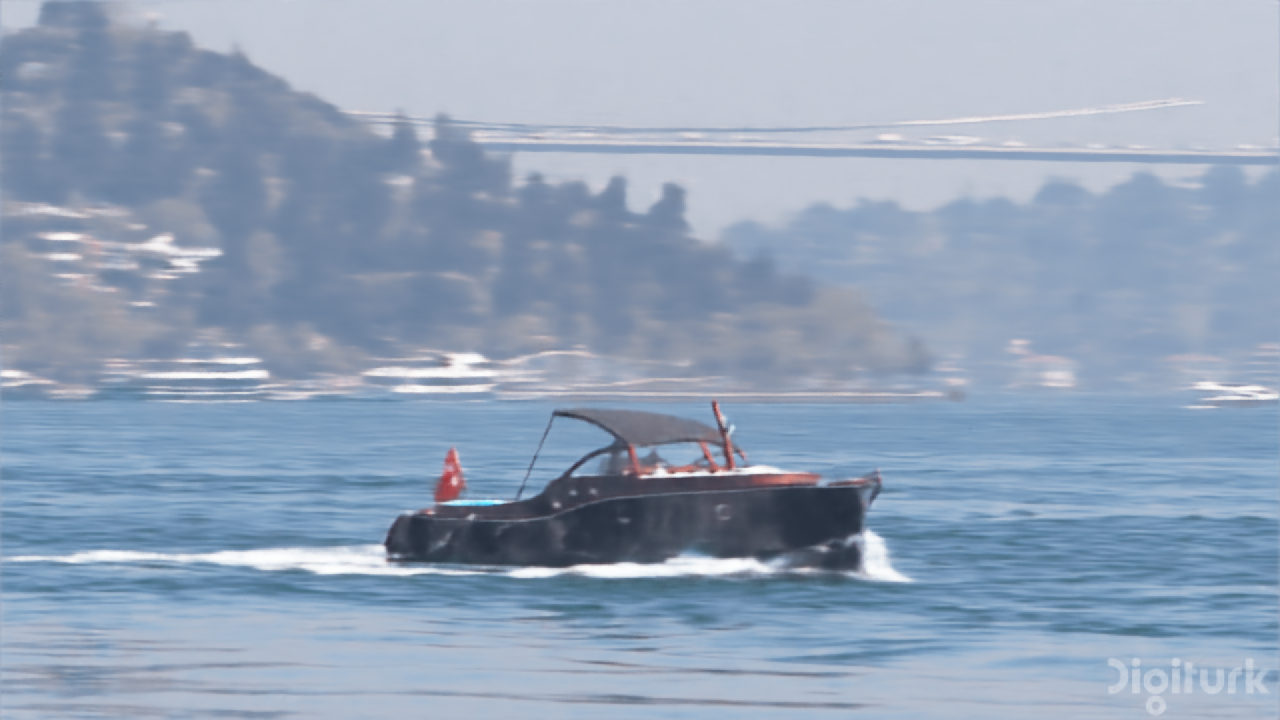} \\
    \multicolumn{5}{l}{\small PFNR, c=50.0\%, sparsity of {$f$-NeRV3=0.5\% (31.06, PSNR)}} \\


    \includegraphics[width=0.20\columnwidth, trim={0.1cm 0.4cm 0.1cm 0.1cm}, clip]{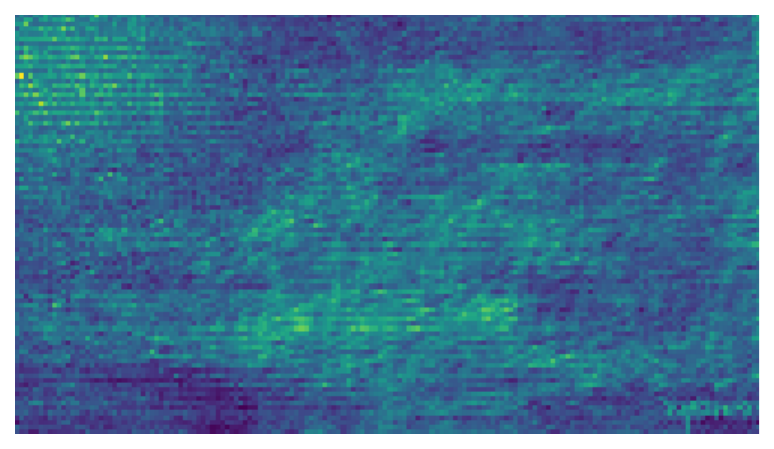} & 
    \includegraphics[width=0.20\columnwidth, trim={0.1cm 0.4cm 0.1cm 0.1cm}, clip]{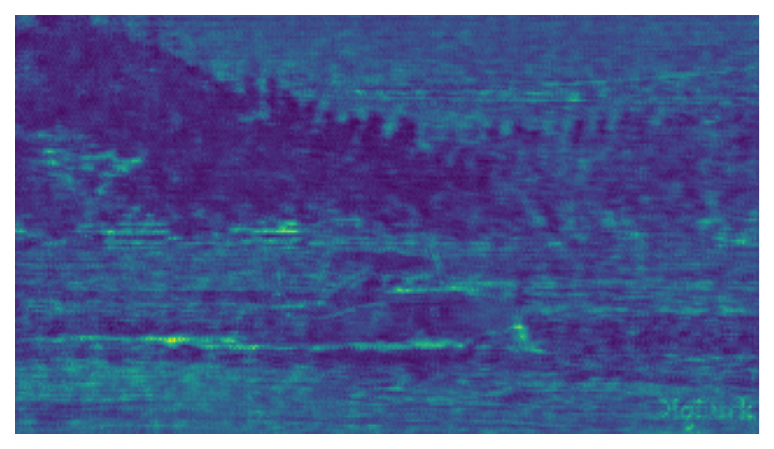} & 
    \includegraphics[width=0.20\columnwidth, trim={0.1cm 0.4cm 0.1cm 0.1cm}, clip]{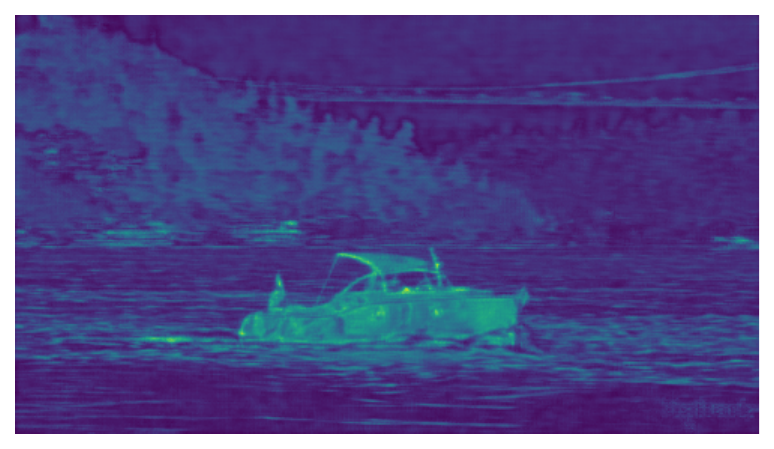} & 
    \includegraphics[width=0.20\columnwidth, trim={0.1cm 0.4cm 0.1cm 0.1cm}, clip]{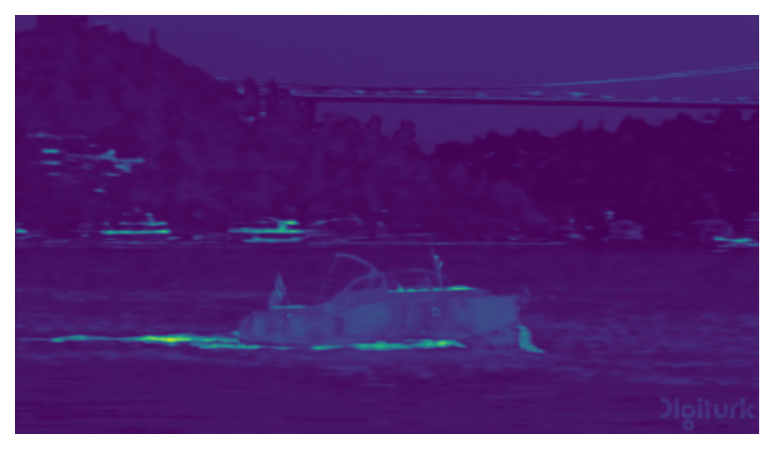} & 
    \includegraphics[width=0.19\columnwidth]{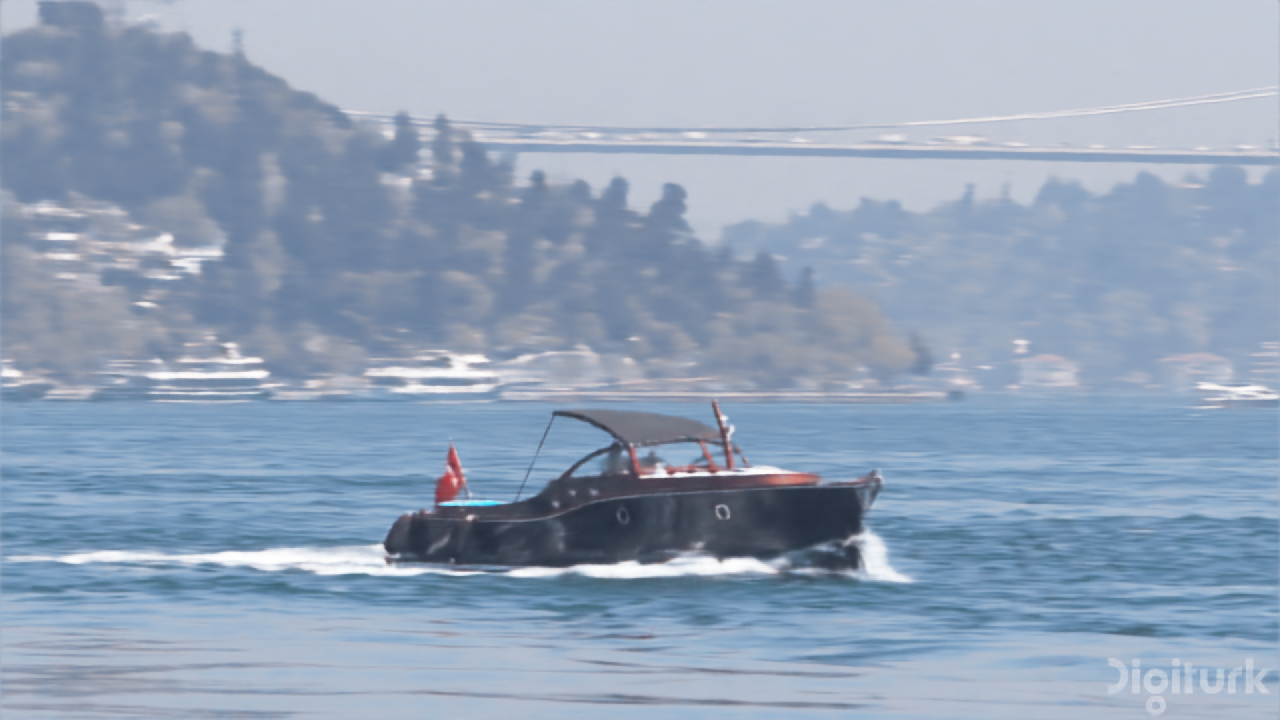} \\
    \multicolumn{5}{l}{\small WSM, c=50.0\%, sparsity of {$f$-NeRV3=5.0\% (32.38, PSNR)}} \\



    \includegraphics[width=0.20\columnwidth, trim={0.1cm 0.4cm 0.1cm 0.1cm}, clip]{FSO-materials/images/fmaps/5/nerv3/pred_0_l2.pdf} & 
    \includegraphics[width=0.20\columnwidth, trim={0.1cm 0.4cm 0.1cm 0.1cm}, clip]{FSO-materials/images/fmaps/5/nerv3/pred_0_l3.pdf} & 
    \includegraphics[width=0.20\columnwidth, trim={0.1cm 0.4cm 0.1cm 0.1cm}, clip]{FSO-materials/images/fmaps/5/nerv3/pred_0_l4.pdf} & 
    \includegraphics[width=0.20\columnwidth, trim={0.1cm 0.4cm 0.1cm 0.1cm}, clip]{FSO-materials/images/fmaps/5/nerv3/pred_0_l5.pdf} & 
    \includegraphics[width=0.19\columnwidth]{FSO-materials/images/fmaps/5/nerv3/pred_0.png} \\
    
    \multicolumn{5}{l}{\small WSN, c=50.0\%, sparsity of \textbf{$f$-NeRV3=50.0\% (34.05, PSNR)}} \\
   
    \end{tabular}
    }
    \vspace{-0.05in}
    \caption{\small \textbf{(VIL), Various sparsity of $f$-NeRV3 ranging from 0.05 \% (top row) to 50.0 \% (bottom row) on the UVG17 dataset.}}
    \label{fig:fmap_sparsity_uvg17}
\end{figure}

%% file: FSO-materials/plot_main_pfnr_var_freq.tex
\begin{figure}[ht]
    \centering
    \vspace{-0.1in}
    \renewcommand{\arraystretch}{1}
    \setlength{\tabcolsep}{0pt}{%
    \begin{tabular}{cc}

    \includegraphics[width=0.5\columnwidth]{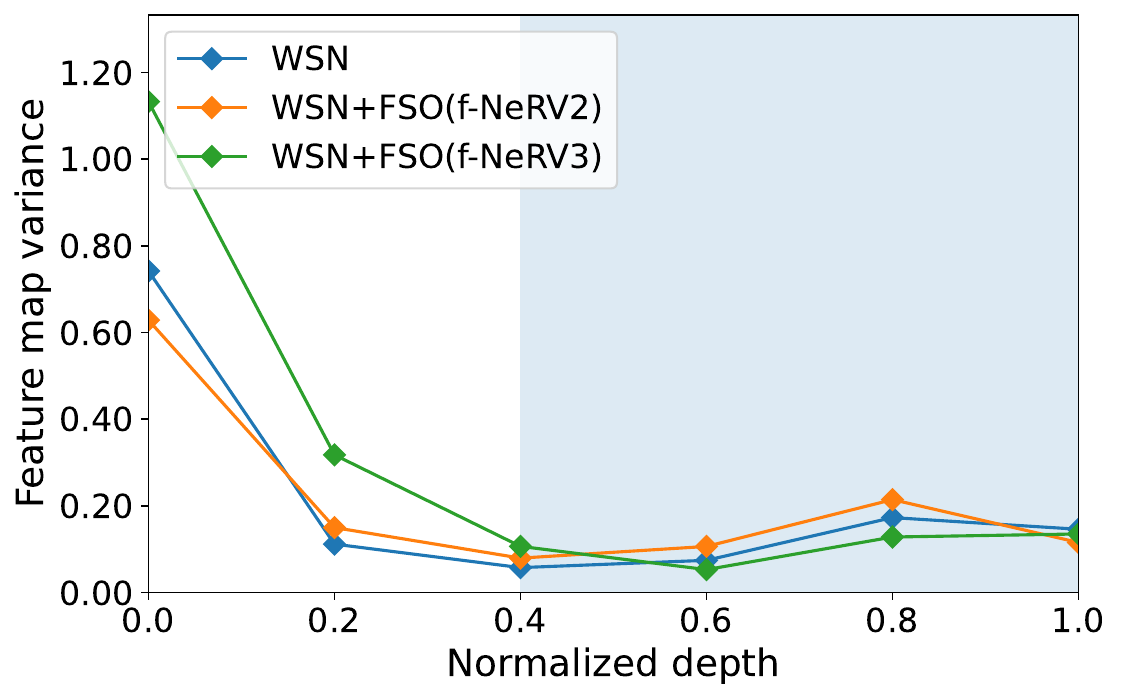} &
    \includegraphics[width=0.5\columnwidth]{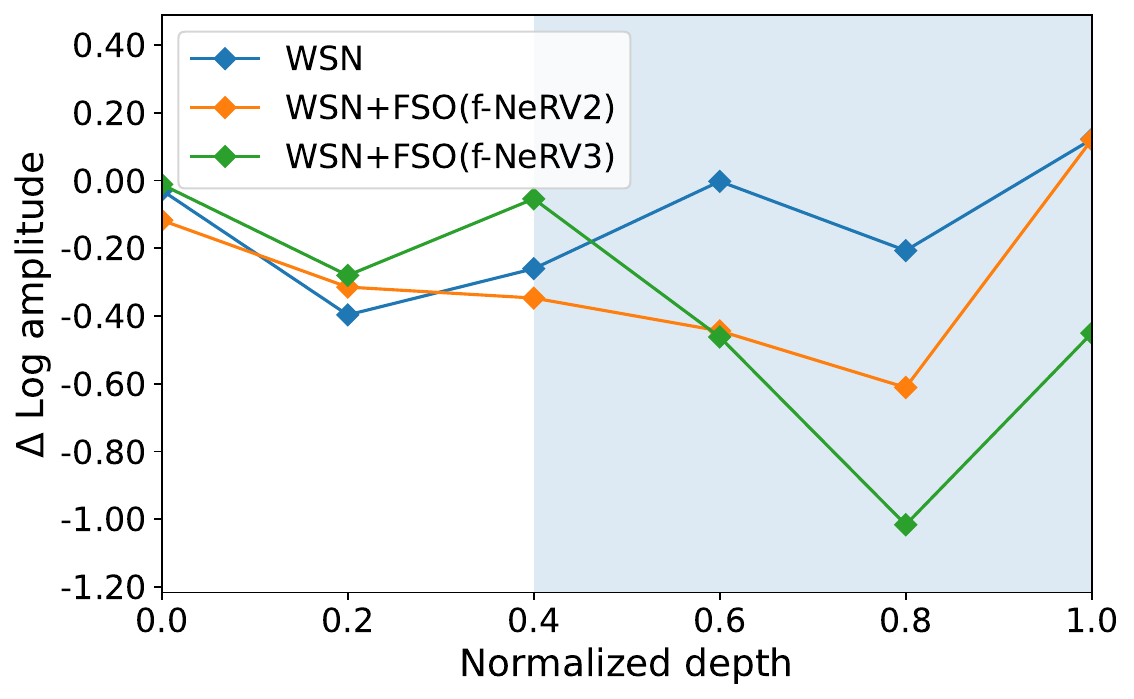} \\
    \vspace{-0.1in}
    \small (a) feature map variance & \small (b) high-freq. of feature map \\



    \end{tabular}}
    \vspace{-0.01in}
    \caption{\small \textbf{(VIL), Comparisons of WSN of NeRVs (blue area) with FSO (white area) in terms of Feature variances and high-frequency components}: (a) offers the variance of the feature map and (b) provides $\Delta \log$ amplitudes at high-frequency ($1.0 \pi$).}
    
    \label{fig:pfnr_var_vs_freq}
\vspace{-0.1in}
\end{figure}

%% file: FSO-materials/plot_main_psnr_bpp.tex
\begin{figure}[h]
    \centering
    \small
    \setlength{\tabcolsep}{0pt}{%
    \begin{tabular}{cc}
    \includegraphics[width=0.5\columnwidth]{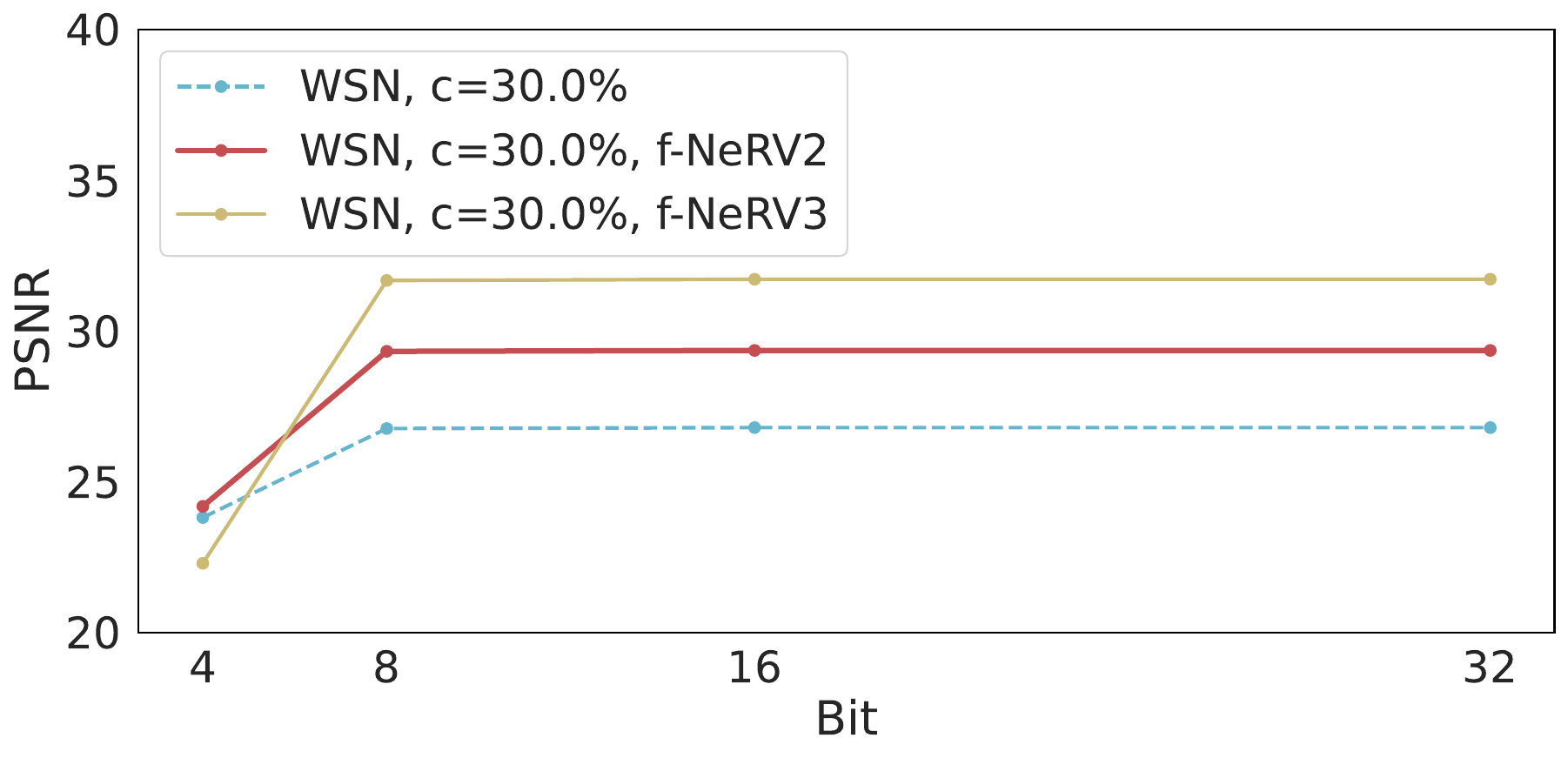} &
    \includegraphics[width=0.5\columnwidth]{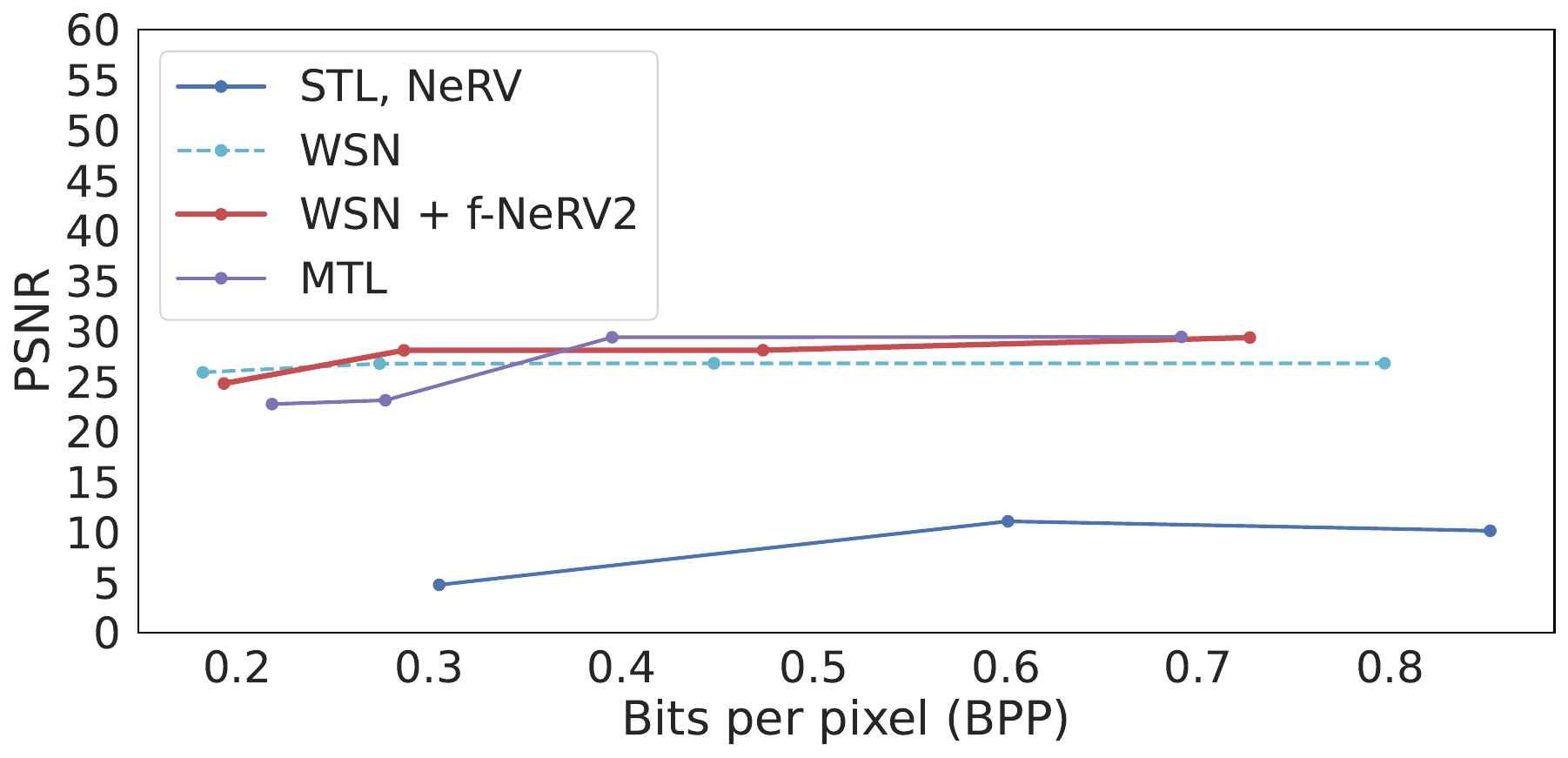} \\
    \vspace{-0.05in}
    \small (a) Quan. and Comp. & \small (b) PSNR v.s. BPP
    \end{tabular}
    }
    \vspace{-0.05in}
    \caption{\small \textbf{(VIL), PSNR v.s. Bits-per-pixel (BPP) on the UVG17 datasets.} }
    \label{fig:uvg17_fso_psnr}
    \vspace{-0.12in}
\end{figure}

%% file: FSO-materials/plot_capacity.tex
\begin{figure}[ht]
    \centering
    \small
    \setlength{\tabcolsep}{0pt}{%
    \begin{tabular}{cc}
    \includegraphics[width=0.5\columnwidth]{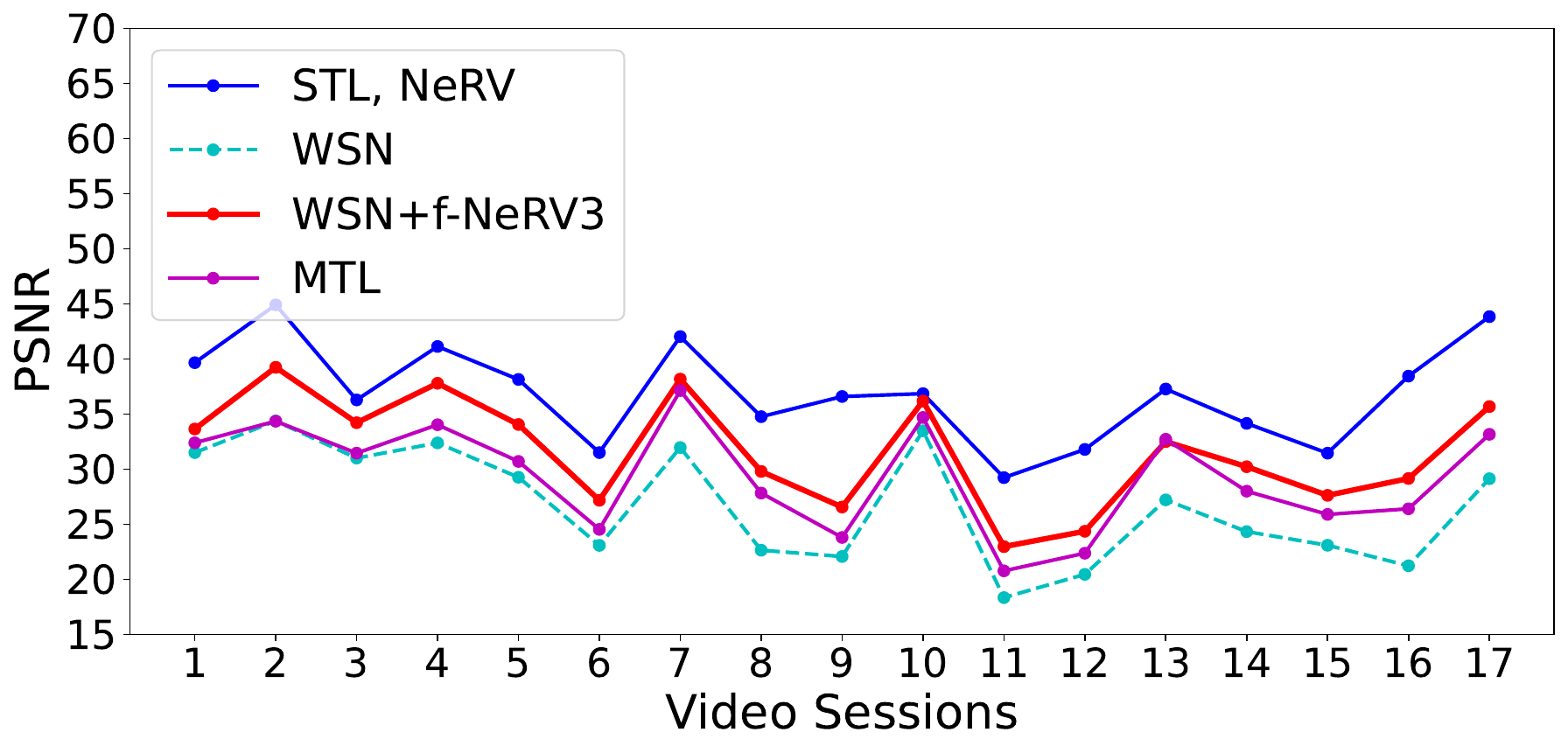} &
    \includegraphics[width=0.5\columnwidth]{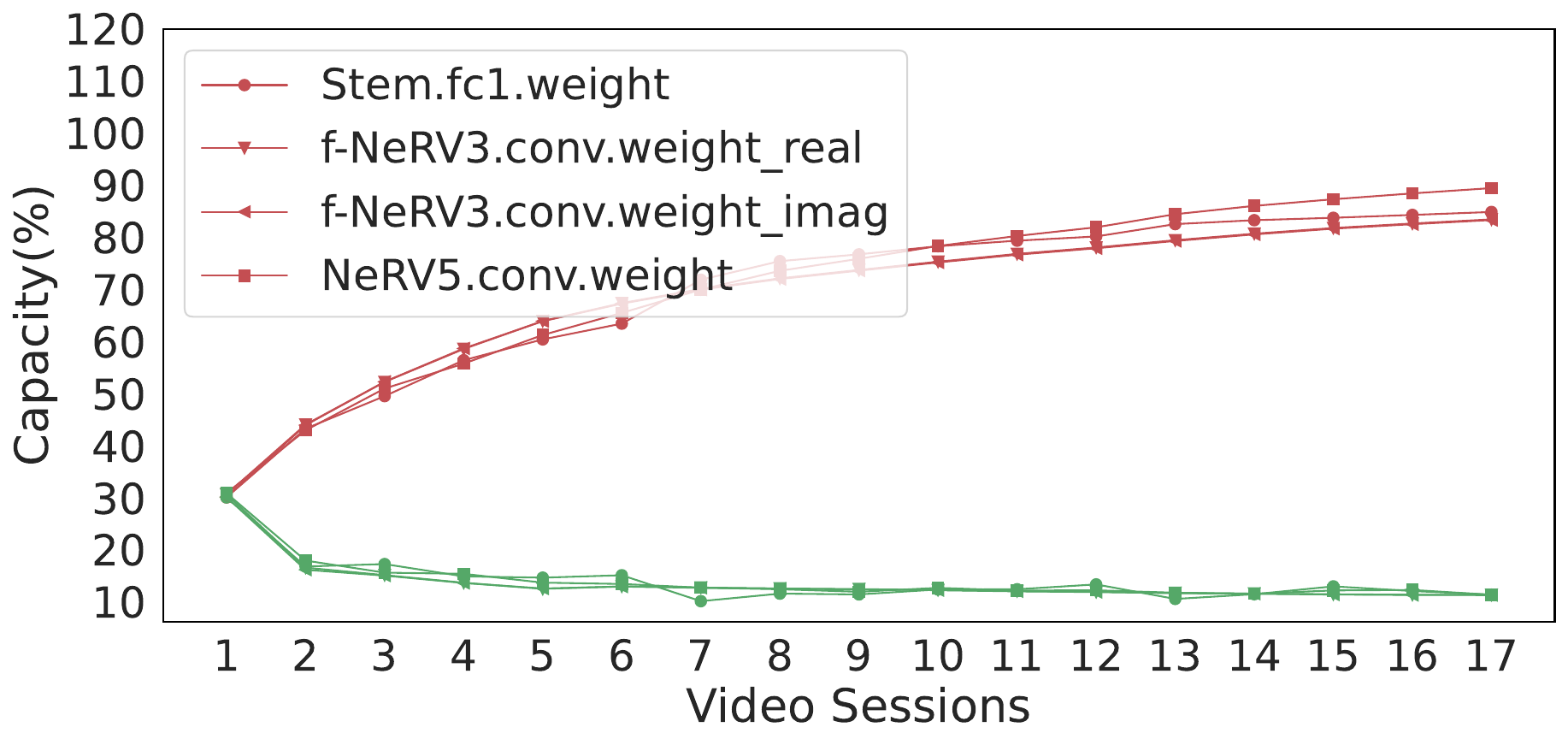} \\
    \vspace{-0.05in}
    \small (a) PSNR, $c = 30.0\%$ & \small (b) Capacity, $c = 30.0\%$ \\
    \end{tabular}
    }
    \vspace{-0.05in}
    \caption{\small \textbf{(VIL), WSN+FSO's Comparison of PSNR with others and layer-wise accumulated capacities on the UVG17 dataset.} Note that, in (b), \textbf{green} represents the percentage of reused subnetwork's parameters of Stem, $f$-NeRV3, and NeRV5 at the current session (s) obtained at the past (s-1) video sessions}
    \label{fig:psnr_cap}
    \vspace{-0.12in}
\end{figure}

%% file: FSO-materials/main_table_uvg8_psnr_fso.tex
\begin{table}[!ht]
\small
\centering
\vspace{-0.05in}
\caption{\small \textbf{(VIL)}, WSN+Fourier Subnueral Operator (FSO) layer (\textbf{$f$-NeRV$\ast$}, c=50.0 \%) on UVG8 Video Sessions with average PSNR and Backward Transfer (BWT). Note that \textit{w/o imag.} ignores the imaginary part in $f$-NeRV$\ast$.}
\resizebox{0.5\textwidth}{!}{
\begin{tabular}{lcccccccccc}
\toprule 

\multicolumn{1}{c}{\multirow{2}{*}{\textbf{Method}}} & \multicolumn{8}{c}{\textbf{Video Sessions}} & \multirow{2}{*}{\thead{\textbf{Avg. PSNR / } \\ \textbf{BWT}}} \\ 

\cline{2-9}
& \textbf{1} & \textbf{2} & \textbf{3} & \textbf{4} & \textbf{5} & \textbf{6} & \textbf{7} & \textbf{8} &  \\ \midrule

{$f$-NeRV2} & \textbf{34.46} & \textbf{33.91} & \textbf{32.17} & \textbf{36.43} & \textbf{25.26} & \textbf{20.74} & \textbf{30.18} & \textbf{25.45} & \textbf{29.82} / \textbf{0.0}  \\ 

{$f$-NeRV2} \textbf{w/o imag.} & 34.34 & 33.79 & 32.04 & 36.40 & 25.11 & 20.59 & 30.17 & 25.27 & 29.71 / 0.0  \\ 
\midrule

{$f$-NeRV3} & \textbf{36.45} & \textbf{35.15} & \textbf{35.10} & \textbf{38.57} & \textbf{28.07} & \textbf{23.06} & \textbf{32.83} & \textbf{27.70} & \textbf{32.12} / \textbf{0.0} \\ 

{$f$-NeRV3} \textbf{w/o imag.} & 35.66 & 34.65 &	34.09 &	37.95 &	25.80 &	21.94 &	32.17 &	26.91 &	31.15 / 0.0 \\ 

\bottomrule
\end{tabular}
}
\label{table:uvg8_fso_real}
\vspace{-0.1in}
\end{table}

%% file: FSO-materials/main_table_uvg8_psnr_conv.tex
\begin{table}[!ht]
\small
\centering
\vspace{-0.05in}
\caption{\small \textbf{(VIL)}, WSN+Fourier Subnueral Operator (FSO) layer (\textbf{$f$-NeRV$\ast$}, c=50.0\%) on UVG8 Video Sessions with average PSNR and Backward Transfer (BWT). Note that \textit{w/o conv.} ignores the conv. layer in $f$-NeRV$\ast$.}
\resizebox{0.5\textwidth}{!}{
\begin{tabular}{lcccccccccc}
\toprule 

\multicolumn{1}{c}{\multirow{2}{*}{\textbf{Method}}} & \multicolumn{8}{c}{\textbf{Video Sessions}} & \multirow{2}{*}{\thead{\textbf{Avg. PSNR / } \\ \textbf{BWT}}} \\ 

\cline{2-9}
& \textbf{1} & \textbf{2} & \textbf{3} & \textbf{4} & \textbf{5} & \textbf{6} & \textbf{7} & \textbf{8} &  \\ \midrule

\textcolor{black}{$f$-NeRV2} & \textbf{34.46} & \textbf{33.91} & \textbf{32.17} & \textbf{36.43} & \textbf{25.26} & \textbf{20.74} & \textbf{30.18} & \textbf{25.45} & \textbf{29.82} / \textbf{0.0}  \\ 

\textcolor{black}{$f$-NeRV2} \textbf{w/o conv.} & 30.05 &  32.10 & 30.12 &	31.82 &	24.00 &	19.60 &	28.21 & 	24.47 &	27.54 / 0.0  \\ 
\midrule

\textcolor{black}{$f$-NeRV3} & \textbf{36.45} & \textbf{35.15} & \textbf{35.10} & \textbf{38.57} & \textbf{28.07} & \textbf{23.06} & \textbf{32.83} & \textbf{27.70} & \textbf{32.12} / \textbf{0.0} \\ 

\textcolor{black}{$f$-NeRV3} \textbf{w/o conv.} & 35.46 & 35.06 & 34.98 & 38.23 & 28.00 & 22.98 & 32.57 & 27.45 & 31.84 / 0.0 \\ 

\bottomrule
\end{tabular}
}
\label{table:uvg8_fso_conv}
\vspace{-0.05in}
\end{table}

%% file: FSO-materials/main_table_uvg8_psnr_stl.tex
\begin{table}[!ht]
\small
\centering
\vspace{-0.05in}
\caption{\small \textbf{(VIL)}, WSN+Fourier Subnueral Operator (FSO) layer (\textbf{$f$-NeRV$\ast$}) on UVG8 Video Sessions with average PSNR and Backward Transfer (BWT).}
\resizebox{0.5\textwidth}{!}{
\begin{tabular}{lcccccccccc}
\toprule 

\multicolumn{1}{c}{\multirow{2}{*}{\textbf{Method}}} & \multicolumn{8}{c}{\textbf{Video Sessions}} & \multirow{2}{*}{\thead{\textbf{Avg. PSNR / } \\ \textbf{BWT}}} \\ 

\cline{2-9}
& \textbf{1} & \textbf{2} & \textbf{3} & \textbf{4} & \textbf{5} & \textbf{6} & \textbf{7} & \textbf{8} &  \\ \midrule 

STL, NeRV~\cite{chen2021nerv}$^{\ast}$  & 39.66 & 36.28 & 38.14 & 42.03 & 36.58 & 29.22 & 37.27 & 31.45 & 36.33 / ~~-~~ \\ 
\midrule 

STL, NeRV , {$f$-NeRV2}            & {39.73} & {36.30} & {38.29} & {42.03} & {36.64} & {29.25} & {37.35} & {31.65} & {36.40} / ~~-~~ \\ 


STL, NeRV , \textbf{$f$-NeRV3}  & \textbf{42.75}  & \textbf{37.65}  & \textbf{42.05} & \textbf{42.36} & \textbf{40.01} & \textbf{34.21} & \textbf{40.15}  & \textbf{36.15} & \textbf{39.41}  / ~~-~~ \\ 

\bottomrule
\end{tabular}
}
\label{table:uvg8_stl}
\vspace{-0.12in}
\end{table}

%% file: FSO-materials/plot_transfer_matrix.tex
\begin{figure}[!h]
    \centering
    \small
    \setlength{\tabcolsep}{0pt}{%
    \begin{tabular}{cc}
    \includegraphics[width=0.5\columnwidth]{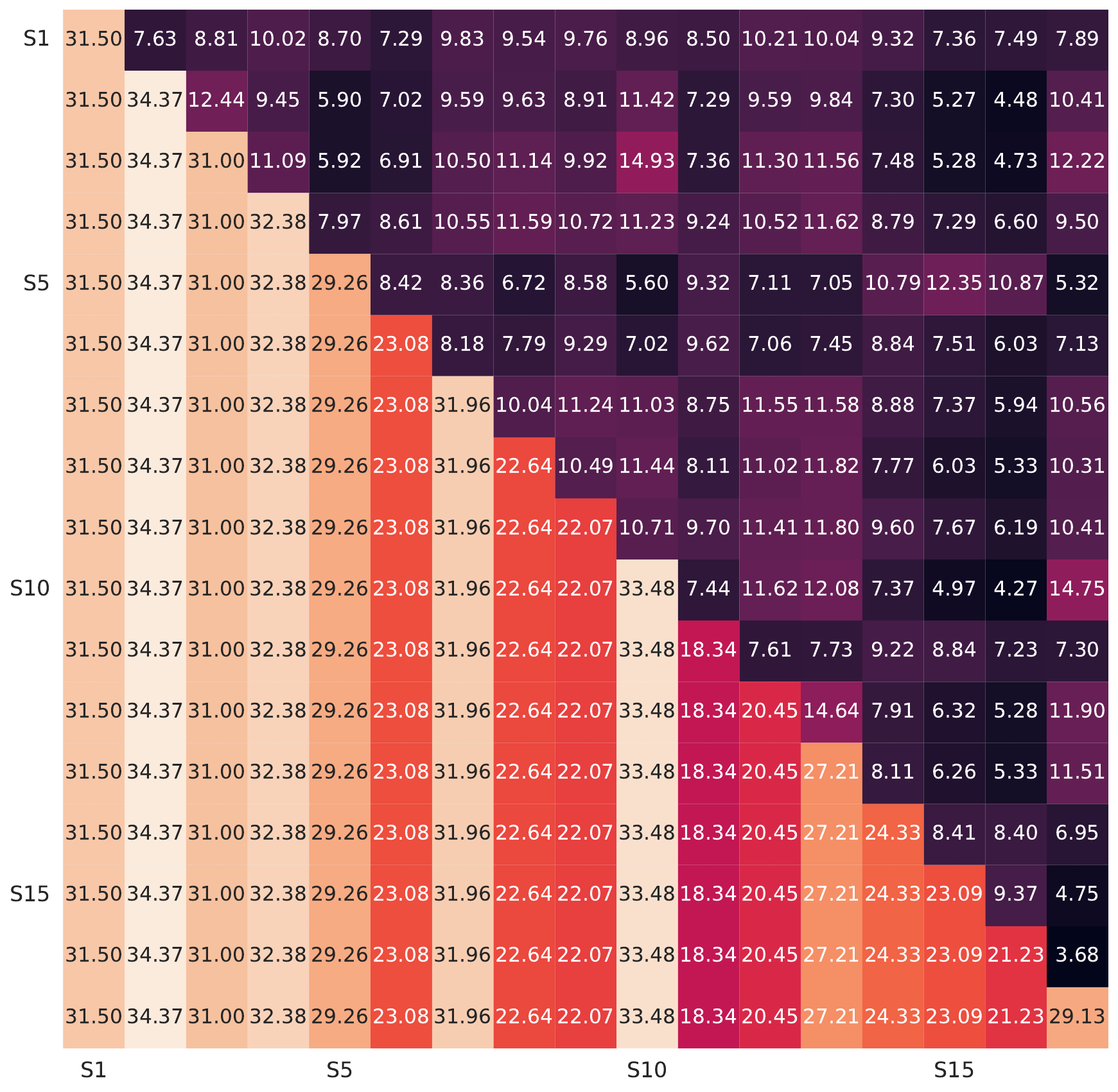} & 
    \includegraphics[width=0.5\columnwidth]{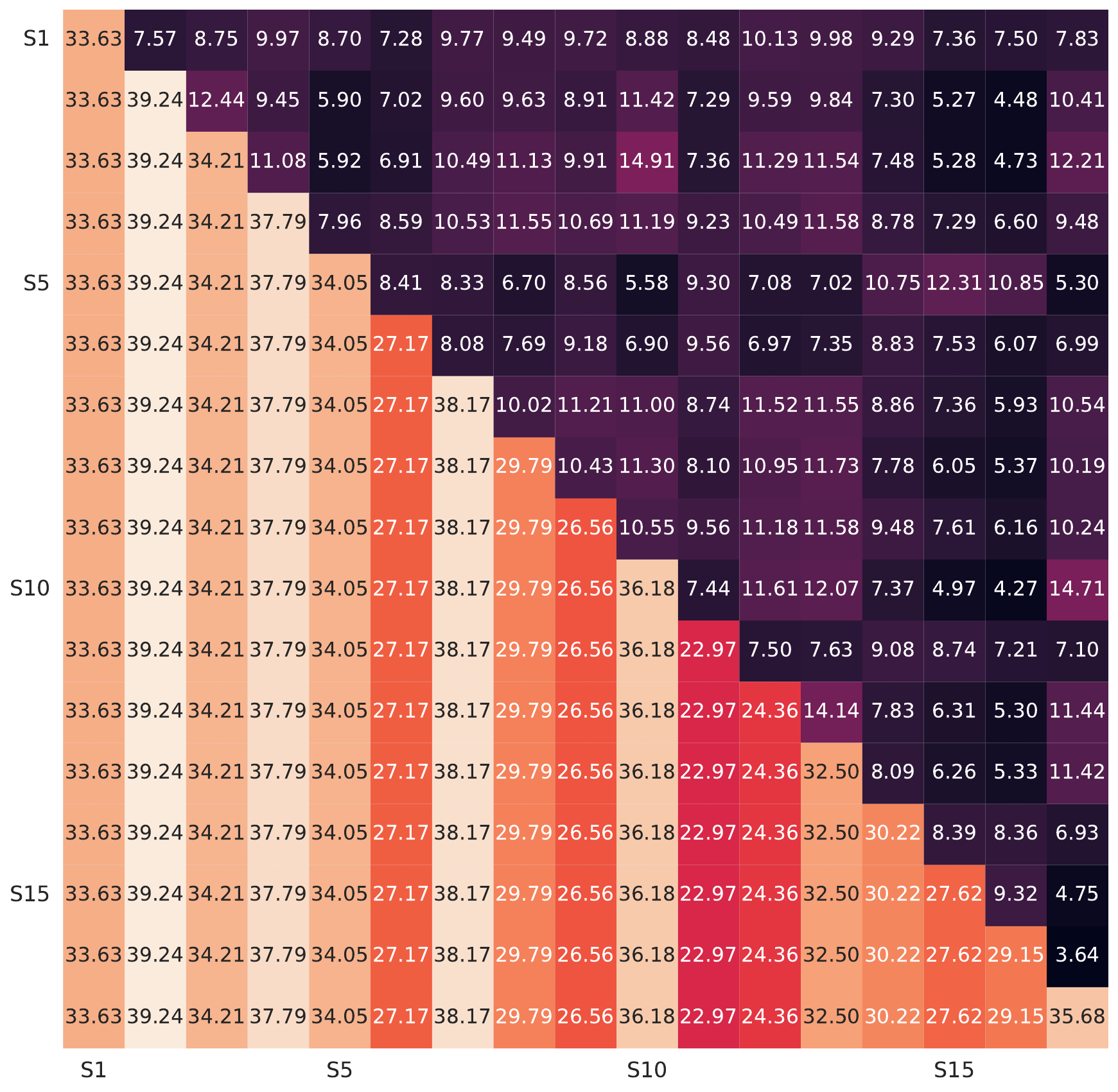} \\
    \vspace{-0.05in}
    \small (a) WSN, $c = 30.0\%$ & \small (b) WSN, $c = 30.0\%$, \textbf{$f$-NeRV3} \\
    \end{tabular}
    }
    \vspace{-0.05in}
    \caption{\small \textbf{(VIL), Transfer Matrixes on the UVG17 dataset measured by PSNR of source and target videos.}}
    \label{fig:transf_matrix}
    \vspace{-0.18in}
\end{figure}

%% file: 6_conclusion.tex
\section{Conclusion}
Inspired by \emph{Regularized Lottery Ticket Hypothesis (RLTH)}, which states that competitive subnetworks exist within a dense network in continual learning tasks, we introduce an interpretable continual learning approach referred to as \emph{Winning Subnetworks}, WSN, which leverages re-used weights within dense networks in Task Incremental Learning (TIL) and Task-agnostic Incremental Learning (TaIL) scenarios. We have also introduced variants of WSN, such as Soft-subnetwork (SoftNet), to address the overfitting in few-shot class incremental learning (FSCIL). Additionally, To overcome the limitation of WSN (sparse re-used weights) in video incremental learning (VIL), we have a newly proposed module as another variant of WSN that aims to find an adaptive and compact sub-module referred to as \emph{Fourier Subneural Operator (FSO)} in Fourier space to encode videos in each video training session. The FSO finds reusable winning subnetworks in Fourier space, providing various bandwidths. We extend Fourier representations to various continual learning scenarios such as VIL, VaIL, TIL, and FSCIL. Extensive experiments demonstrate the effectiveness of FSO in three continual learning scenarios. Overall, FSO's representation markedly enhanced task performance at different convolutional representational levels, the higher layers for TIL and FSCIL and the lower layers for VIL.

%% file: 7_appendix.tex
To help readers better understand our contributions, we have prepared additional explanations as follows:

\begin{itemize}[leftmargin=*]
    \item \textbf{.1 Continual Learning Scenarios}: The FSO has combined various continual learning scenarios through challenging points and insightful novelty.
    \item \textbf{.2 Variations of WSN through FSO}: We have investigated all variations of WSN using FSO in classification and video representations.
    \item \textbf{.3 WSN and SoftNets through FSO}: We have re-summarized the relationship between WSN and SoftNet through convergence theory and additional experiments.
    \item \textbf{.4 Advantages of WSN-based CL}: To strengthen our advantages (WSN+FSO), we have prepared the comparisons of Winning Ticket-based Continual Learning (WSN) with Prompt-Tunning-based Continual Learning and shown WSN's parameter-efficiency.

    \item \textbf{.5 Fourior Subneural Operator (FSO)}: We have shown the mechanism of FSO to represent and transfer core signals in CL properly.
\end{itemize}

\subsection{Continual Learning Scenarios}\label{app:cl_scenarios}
The Catastrophic Forgetting (CF) is inevitable in all Continual Learning (CL) scenarios. Various approaches have been taken to alleviate the CF problem. Moreover, leveraged by abstracting representation, the CL model, i.e., ResNets, which performs sequential classifications such as TIL, TaIL, CIL, and FSCIL, has another challenging point: it loses the global structure of image instances at a high layer, leading to deteriorated classification performances. In this work, we refer to the issue as an \underline{(1) abstract representation}. 

In contrast, we observed a lack of representation power in VIL for sequential implicit neural representation learning, led by the conventional convolutional layer (NeRV). We refer to this issue as a \underline{(2) lack of representation power}.

For the reader to understand the two challenges and the new architecture-based points (\underline{FSO}), we briefly summarize the specific continual learning scenarios (TIL, TaIL, CIL, FSCIL, and VIL) as follows: \\

\noindent
\textbf{Task-Incremental Learning (TIL).} Each session is a separate classification problem. The session ID of instances is provided during training and testing, and the continual learner uses ConvNet or ResNet. TIL's model trains a sequence of sessions $s_1, s_2, \cdots, s_{|\mathcal{S}|}$, incrementally. Each session $s \in \mathcal{S}$ has a training dataset $\mathcal{D}_s = \{(\bm{x}_s^i, y_s^i)\}_{i=1}^{n_s}$ where $(\bm{x}_s^i, y_s^i) \in \mathcal{X} \times \mathcal{Y}_s$ and $\mathcal{Y}_s$ are disjoint and $\cup_{s=1}^{\mathcal{|S|}}\mathcal{Y}_s = \mathcal{Y}$. The objective is to learn $f_{\bm{\theta} \odot \bm{m}_s}: \mathcal{X} \times \mathcal{S} \rightarrow \mathcal{Y}$.  

\begin{itemize}[leftmargin=*]
\item (challenging points): CF, model size, \underline{(1) abstract representation}. 
\item (novelties): \underline{FSO}, WSN (previous work's novelty).
\end{itemize}

\noindent
\textbf{Task-agnositic Incremental Learning (TaIL).} The TaIL follows the same structure as TIL. However, the session ID of instances is \textbf{NOT} provided during training and testing, and the continual learner (ConvNet or ResNet) must infer the session ID in the testing, as stated in \Cref{sub_sec:opt_til}. The TaIL is the most challenging CL scenario.

\begin{itemize}[leftmargin=*]

\item (challenging points): CF, model size, session ID inference, \underline{(1) abstract representation}.
\item (novelties): \underline{FSO}, WSN (previous work's novelty).

\end{itemize}

\noindent
\textbf{Class-Incremental Learning (CIL).} The CIL process builds a single classifier for all sessions/classes learned so far. In training, the session ID of the instances is given to a continual learner (ConvNet or ResNet). However, in testing, an instance from any class may be presented for the model to classify. There is no prior session information on the test instance. CIL's model trains a sequence of sessions $s_1, s_2, \cdots, s_{|\mathcal{S}|}$, incrementally. Each session $s \in \mathcal{S}$ has a training dataset $\mathcal{D}_s = \{(\bm{x}_s^i, y_s^i)\}_{i=1}^{n_s}$ where $(\bm{x}_s^i, y_s^i) \in \mathcal{X}_s \times \mathcal{Y}_s$ and $\mathcal{Y}_s$ are disjoint and $\cup_{s=1}^{\mathcal{|S|}}\mathcal{Y}_s = \mathcal{Y}$. The objective is to learn $f_{\bm{\theta} \odot \bm{m}_s}: \mathcal{X} \times \mathcal{S} \rightarrow \mathcal{Y}$, where session ID $s$ is to be inferred in the testing time. 

\begin{itemize}[leftmargin=*]

\item (challenging points): CF, model size, session ID inference, \underline{(1) abstract representation}.
\item (novelties): \underline{FSO}, WSN (previous work's novelty).

\end{itemize}

\noindent
\textbf{Few-Shot Class-Incremental Learning (FSCIL).}
FSCIL follows the same structure as CIL. However, in new sessions ($s\geq 1$), only a \textbf{FEW} instances are given to the model, while abundant instances are in the base session ($s=1$). This makes it harder to train the CL model to avoid overfitting and alleviating CF simultaneously under the session ID that is not given in the test times.

\begin{itemize}[leftmargin=*]

\item (challenging points): CF, model size, session ID inference, \underline{(1) abstract representation}, \underline{overfitting}.
\item (novelties): \underline{FSO}, SoftNet (previous work's novelty to avoid overfitting).

\end{itemize}

\noindent
\textbf{Video-Incremental Learning (VIL).}
Each session is a separate video representation problem. The session ID of instances is provided during training and testing, and the continual learner uses NeRVs. VIL's model trains a sequence of sessions $s_1, s_2, \cdots, s_{|\mathcal{S}|}$, incrementally. Each session $s \in \mathcal{S}$ has a training dataset $\mathcal{D}_s = \{(\bm{x}_s^i, y_s^i)\}_{i=1}^{n_s}$ where $(\bm{x}_s^i, y_s^i) \in \mathcal{X} \times \mathcal{Y}_s$ and $\mathcal{Y}_s$ are disjoint and $\cup_{s=1}^{\mathcal{|S|}}\mathcal{Y}_s = \mathcal{Y}$. The objective is to learn $f_{\bm{\theta} \odot \bm{m}_s}: \mathcal{X} \times \mathcal{S} \rightarrow \mathcal{Y}$.

\begin{itemize}[leftmargin=*]

\item (challenging points): \underline{(2) lack of representation power}
\item (novelties): \underline{FSO}, WSN (previous work's novelty).

\end{itemize}

%

\subsection{Variations of WSN through FSO}\label{app:variations_wsn}
Regarding the additional structures of WSN, SoftNet, and FSO, we could take various fusion methods such as element-wise summation and concatenations, however, these fusion methods are minor points in this work. We have investigated all variations of WSN using FSO in classification and video representations in this work. Through the inspections, we achieved the current state-of-the-art performances under continual learning scenarios. We summarize the two variations in terms of two types of architectures (ResNet, NeRVs). 

\begin{itemize}[leftmargin=*]

\item \underline{(1) abstract representation}: Through FSO, representations from the lower layer are fed to inputs at the higher layer (ResNets) to maintain global image representation, leading to the best performances in TIL, TaIL, CIL, and FSCIL.
\item \underline{(2) lack of representation power}: FSO captures global contextual representations to generate high-quality video images. We found the most parameter-efficient layer (f-NeRV3 with 8-bit qualitzation) of FSO since FSO's parameters increase at a higher layer (NeRVs).

\end{itemize}

\subsection{WSN and SoftNets through FSO}

We have unified various kinds of continual learning scenario (TIL, FSCIL, CIL, TaIL, and VIL) as well as Single Task Learning (STL, ImageNet-1K as shown in \Cref{table:imagnet1k} and \Cref{table:uvg8_stl}) through FSO, concretely. The remaining point is the relationship between WSN and SoftNet. The SoftNet, which incorporates minor winning tickets ($m < 1$), is a variant of WSN ($m \in \{0,1\}$) to alleviate the overfitting issues caused by few samples given in new session in the FSCIL scenario. Since the overfitting is not main issues in TIL, TaIL, CIL, and VIL, the effectiveness of SoftNet + FSO could be minor in those CL scenarios.

However, we observed SoftNet's powerful forward transfer ability in TIL, where SoftNet was obtained by inducing small perturbations ($U(0, 1e-3)$) into the zeros mask values of trained WSN. As shown in \Cref{tab:main_soft_til}, SoftNet obtained competitive performance with WSN and showed powerful forward transfer (FWT) ability over WSN (see \Cref{fig:conf_soft_wsn_cifar100}). We could explain this result from binary map correlations (see \Cref{fig:corr_soft_wsn_cifar100}) and the following convergence of WSN and SoftNet, stated in previous work~\cite{kang2022soft}: WSN and SoftNet have flatter minima than Dense Network, and small perturbed WSN (SoftNet) could predict unseen session instances in the flat minima. 

\noindent
\textbf{Convergences of WSN / SoftNet.}
We hypothesize that fine-tuning SoftNet on flatter minima mitigates overfitting in FSCIL more effectively than WSN or dense networks on sharper minima since SoftNet, operating in flatter minima, is less prone to converging to local minima. To analyze SoftNet's convergence behavior, we adopt the framework of Lipschitz-continuous objective gradients~\cite{boyd2004convex, bottou2018optimization}:
\begin{lemma} 
\textbf{(Lipschitz Continuity of Dense Networks)}: Let $R:\mathbb{R}^d \rightarrow \mathbb{R}$ be the objective function of a dense network, which is continuously differentiable. Its gradient function  $\nabla R: \mathbb{R}^d \rightarrow \mathbb{R}^d$, Lipschitz continuous with constant $L > 0$, satisfying the inequality:
\begin{equation} 
\begin{split}
    ||\nabla R(\bm{\theta}) - \nabla R(\bm{\theta}')||_2 & \leq  L||\bm{\theta} - \bm{\theta}'||_2, \;\; \\ 
    & \text{ for } \forall \bm{\theta}, \bm{\theta}' \in \mathbb{R}^d.
\end{split} 
\label{eq:dense_lip}
\end{equation}
\label{lm:dense_lip}
\end{lemma}

\begin{lemma}
\textbf{(Lipschitz Continuity of Subnetworks)}: For a subnetwork defined by a binary mask $\bm{m}$, the objective function $R(\bm{\theta} \odot \bm{m})$ satisfies a sparsity-aware Lipschitz continuity condition:

\begin{equation} 
\begin{split}
    || \nabla R(\bm{\theta} \odot \bm{m}) - \nabla R(\bm{\theta}' \odot \bm{m}) ||_2 &\leq  L || (\bm{\theta} - \bm{\theta}') \odot \bm{m} ||_2, \\ 
    & \text{ for } \forall \bm{\theta}, \bm{\theta}' \in \mathbb{R}^d.
\end{split}
\label{eq:subnet_lip}
\end{equation}

Additionally, using the effective mask size $||\bm{m}||_{eff} = ||\bm{m}||_1^2 / ||\bm{m}||_2^2$, the above Lipschitz continuity condition is refined as:

\begin{equation} 
\begin{split}
    || \nabla R(\bm{\theta} & \odot \bm{m}) - \nabla R(\bm{\theta}' \odot \bm{m}) ||_2   \\
    & < \frac{L}{||\bm{m}||_{eff}}||(\bm{\theta}-\bm{\theta}')\odot\bm{m}||_2
\end{split}
\label{eq:subnet_lip_tight}
\end{equation}

\label{lm:subnet_lip}
\end{lemma}

\begin{theorem}
\textbf{(Convergence Rate of Subnetworks vs. Dense Networks)}:
Subnetworks achieve a faster convergence rate compared to dense networks due to their simpler loss landscapes. For dense networks: $R(\bm{\theta}) = \mathcal{O}(d)$ where $d=||\bm{\theta}||_2$ while subnetwork: $R(\bm{\theta} \odot \bm{m})=\mathcal{O}(1/||\bm{m}||_2)$~\cite{ye2020good}. Given $||\bm{m}||_2 \ll d$, we observe the following comparison:
$$
\frac{R(\bm{\theta}\odot \bm{m})}{R(\bm{\theta})} = \frac{\mathcal{O}(1/||\bm{m}||_2)}{\mathcal{O}(d)}
$$
This suggests that subnetworks can achieve a faster convergence rate, i.e., $R(\bm{\theta} \odot \bm{m}) \ll R(\bm{\theta})$ because their effective loss landscape is less complex, leading to more efficient gradient descent steps. Using \textbf{\textit{Lemmas A.1}} and \textbf{\textit{A.2}}, we further deduce:
\begin{equation}
\begin{split}
    \frac{|| \nabla R(\bm{\theta} \odot \bm{m}) - \nabla R(\bm{\theta}' \odot \bm{m}) ||_2}{|| (\bm{\theta} - \bm{\theta}')\odot\bm{m}||_2}  & \\
    \leq \frac{|| \nabla R(\bm{\theta}) - \nabla R(\bm{\theta}') ||_2}{||\bm{\theta}- \bm{\theta}'||_2} & \leq L.
\end{split}
\end{equation}
This implies that subnetworks, due to their sparsity, exhibit a flatter loss landscape, which promotes more stable and generalizable solutions. The smaller the value is, the flatter the solution (loss landscape) is. The equation is established from the relationship $R(\bm{\theta} \odot \bm{m}) \ll R^\ast(\bm{\theta})$, where $R^\ast(\bm{\theta})$ denotes the best possible loss achievable by convex combinations of all dense network's parameters despite $|| (\bm{\theta} - \bm{\theta}') \odot \bm{m} ||_2 < ||\bm{\theta} - \bm{\theta}'||_2$. 
\end{theorem}

\begin{theorem}
\textbf{(Comparison of WSN and SoftNet)}:
Consider two subnetworks, characterized by masks $\bm{m}_{WSN}$ and $\bm{m}_{SoftNet}$, where $||\bm{m}_{WSN}||_2 < ||\bm{m}_{SoftNet}||_2$. if the corresponding losses are approximately equal, i.e., $|| R(\bm{\theta} \odot \bm{m}_{WSN})-R(\bm{\theta} \odot \bm{m}_{SoftNet})||_2 \simeq 0$, then the following inequality holds:
\begin{equation} 
\begin{split}
    & \frac{|| \nabla R(\bm{\theta} \odot \bm{m}_{WSN}) - \nabla R(\bm{\theta}' \odot \bm{m}_{WSN}) ||_2}{|| (\bm{\theta} - \bm{\theta}') \odot \bm{m}_{WSN} ||_2}  \\ 
    &~~ \geq \frac{|| \nabla R(\bm{\theta} \odot \bm{m}_{SoftNet}) - \nabla R(\bm{\theta}' \odot \bm{m}_{SoftNet}) ||_2}{|| (\bm{\theta} - \bm{\theta}') \odot \bm{m}_{SoftNet}||_2},
\end{split}
\end{equation}
Equality is achieved \textit{iff} $||\bm{m}_{WSN}||_2 = ||\bm{m}_{SoftNet}||_2$. Moreover, the effective mask size using \textbf{\textit{Lemma A.2}} refines the following bound:
$$
\frac{L}{||\bm{m}_{WSN}||_{eff}} > \frac{L}{||\bm{m}_{SoftNet}||_{eff}}.
$$
This suggests that smaller effective masks in subnetworks lead to sharper loss landscapes. These conclusions also holds for $WSN + FSO$ and $SoftNet + FSO$.
\end{theorem}
\Cref{fig:main_plot_loss_lens} visualizes the loss landscapes of dense networks, WSN, and SoftNet. The results illustrate that subnetworks (WSN and SoftNet) achieve flatter global minima than dense networks, with SoftNet exhibiting the flattest loss landscape. These findings support the theoretical claims regarding improved stability and generalization through sparse networks, such as WSN and SoftNet.


\input{supples/materials/plot_hard_soft}

\input{supples/materials/table_til_wsn_softnet_v1}
\input{supples/materials/table_conf_wsn_soft}

\input{supples/materials/table_corr_wsn_soft}

\subsection{Advantages of Winning Ticket-based CL}

First, to strengthen our advantages (WSN+FSO), we prepared the comparisons of Winning Ticket-based Continual Learning (WSN) with Prompt-Tunning-based Continual Learning, as shown in \Cref{table:wsn_prompt}. ViTs require more computational resources and longer training times due to their higher number of floating point operations (FLOPs). Higher FLOPs and retrieval test sample-specific prompts in ViTs can result in slower inference times, a bottleneck for real-time applications (on devices). Deploying ViTs effectively often requires powerful GPUs or TPUs with ample memory and computational capacity. In contrast, ResNets with task-specific binary masks can be more efficiently deployed on less powerful hardware: without dropping performances, the WSN's inference speed is even faster when applying quantization to WSN, as proven in VIL.

\input{supples/materials/wsn_prompt}

Second, when deploying CL models on edge devices, it is crucial to balance the trade-offs between accuracy, model size, computational complexity, and energy consumption. Overall, the continual learning task's accuracy of ViTs is better than WSN. However, in 5-dataset continual learning setting (Class Incremental Learning, CIL), WSN (c=50.0 \%) outperformed prompt-tuning-based CL in accuracy, model efficiency (number of parameters and FLOPs), and backward transfer (BWT), as shown in \Cref{table:5_data_acc}. Considering the critical point that Prompt-Tuning works on sample-specific prompt selections in inference, WSN's computational gain is much higher than that of Prompt-Tuning methods. 

Lastly, we have conducted additional experiments on Task-agnostic Incremental Learning (TaIL) to strengthen our core contributions toward generalized continual learning. Please see our final script. Note that the task-id is inferred by SupSup~\cite{wortsman2020supermasks}, as stated in \Cref{{sub_sec:opt_til}}.

\input{supples/materials/table_acc_wsn_prompt}

\input{supples/materials/plot_imag_part}

\subsection{Fourier Subneural Operator (FSO)}

\noindent 
\textbf{FSO of Real and Imaginary Tickets}:
To properly represent and transfer core signals in CL, we need to find relevant components in Fourier space and inverse transform them. However, we cannot manually design appropriate filters (or directly learn a selector without considering Fourier space) to represent complex real-world signals, such as more extensive scaled-video representations. To explain the concept of FSO, which adaptively selects video-relevant bandwidths in Fourier space, we assume that a complex signal is given, as shown in \Cref{fig:imag_part}(b) and (d). The object of FSO is to find critical periodic coefficients (Real and Imaginary Parts, i.e., 30, 50, and 120Hz in DFT) from the (b) complex signal or (d) DFT of a complex signal to represent (a) the original signal in high quality. We can adequately represent an origin signal if we select the core bandwidths (30, 50, and 120Hz) from (c) and (d) of real and imaginary parts. In contrast, if we select all bandwidths from random perturbed (d), we again represent a complex signal (b). As stated before, we demonstrate this concept in VIL through the two inspections of the importance of Real and Imaginary Parts (see \Cref{table:uvg8_fso_real} \& \Cref{fig:fmap_uvg17}) and diverse sparsity of FSO modules (see \Cref{fig:fmap_sparsity_uvg17}). Without selecting Imaginary Parts properly, FSO could not represent video representations, as shown in \Cref{table:uvg8_fso_real}. FSO in the NeRV3 blocks tends to select high-frequency components, leading to the best video representations, as shown in \Cref{fig:fmap_uvg17}. Moreover, the chosen adequate bandwidths in FSO are crucial for better performances, as shown \Cref{fig:fmap_sparsity_uvg17}.  In addition, our FSO adaptively finds periodic coefficients (Real and Imaginary parts) for image or video representations of one session in Fourier space and transfers them to those of others in Continual Learning Scenarios. These behaviors of FSO make WSN train faster and obtain high-quality video representations.




%% file: supples/materials/plot_hard_soft.tex
\begin{figure}[ht]
    \centering
    \setlength{\tabcolsep}{-6pt}{%
    \begin{tabular}{cc}
    \includegraphics[width=0.25\textwidth]{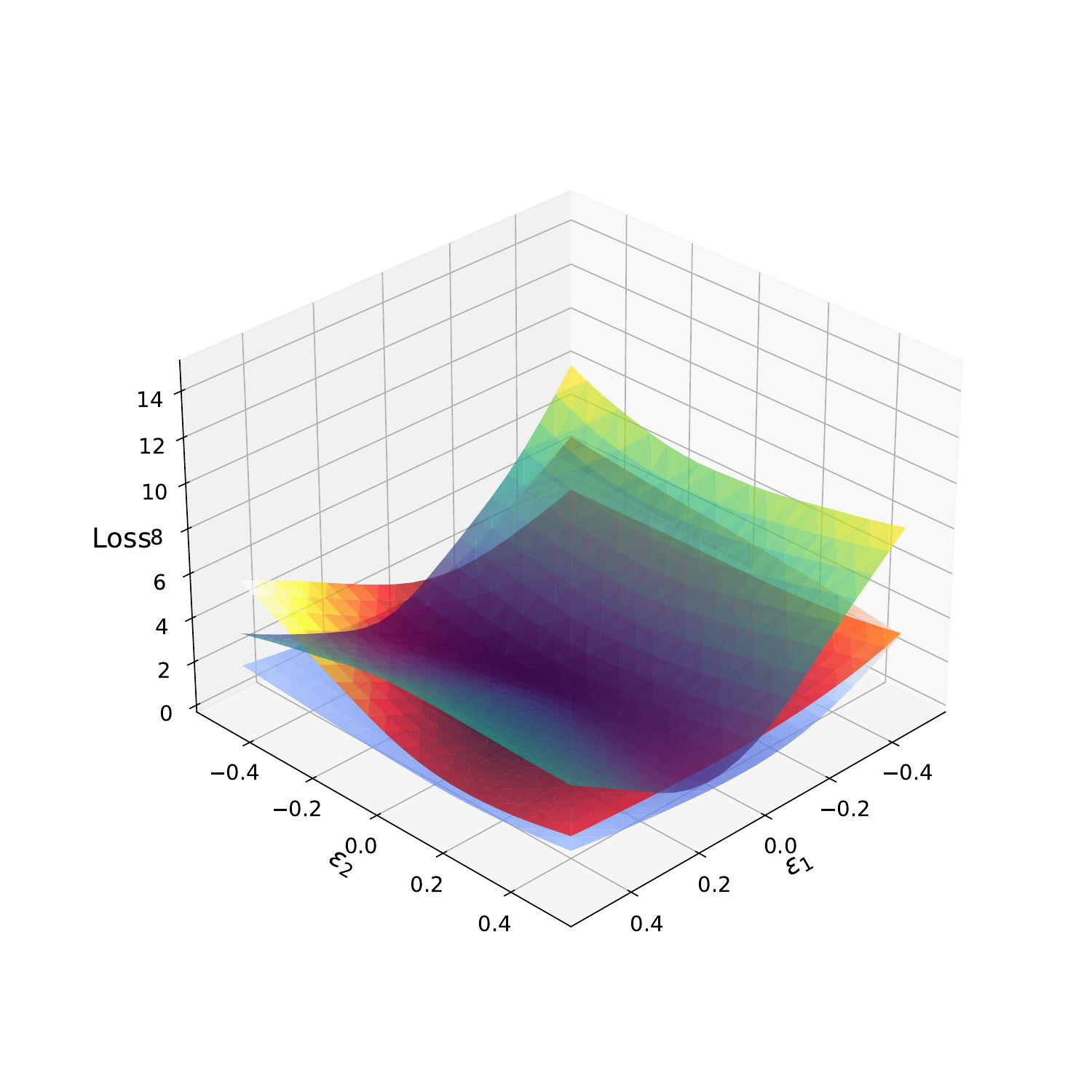} &
    \includegraphics[width=0.25\textwidth]{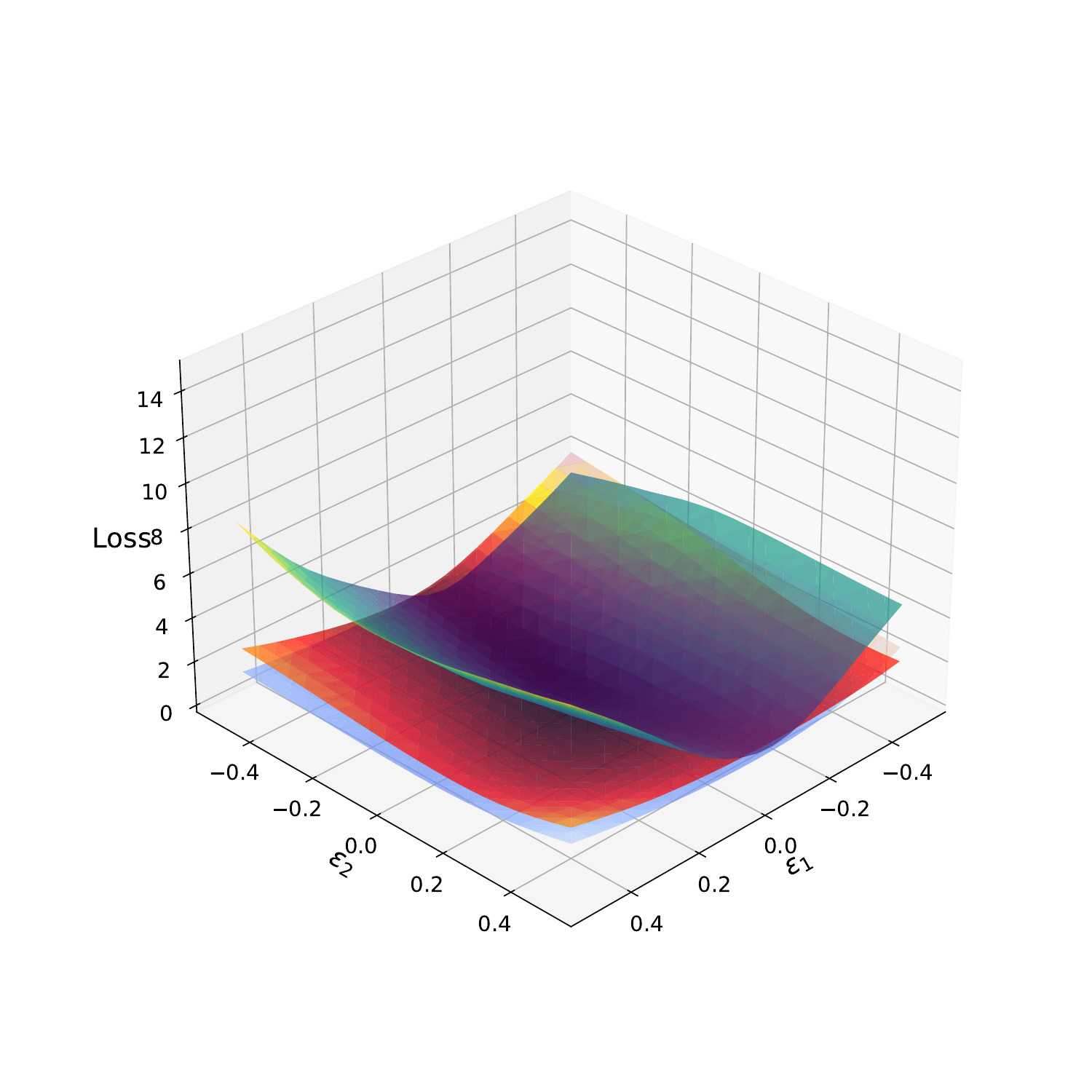} \\ 
    
    \small (a) epoch=70, $c=10.0\%$ & \small (b) epoch=100, $c=10.0\%$  \\
    \end{tabular}
    }
    \caption{\small Loss landscapes of \textcolor{teal}{Dense Network}, \textcolor{red}{WSN}, and \textcolor{cyan}{SoftNet}: Subnetworks provide a more flat global minimum than dense neural networks. To demonstrate the loss landscapes, we trained a simple three-layered, fully connected model (fc-4-25-30-3) on the Iris Flower dataset (which is three classification problem) for 100 epochs \cite{hart1968condensed,dasarathy1980nosing}.}
    \label{fig:main_plot_loss_lens}
\end{figure}

%% file: supples/materials/table_til_wsn_softnet_v1.tex
\begin{table}[ht]
\begin{center}
\caption{\small \textbf{(TIL)}, {Performance comparisons of the proposed method and other baselines} - PackNet~\cite{mallya2018packnet} and SupSup~\cite{wortsman2020supermasks} - on various benchmark datasets. We report the mean and standard deviation of the average accuracy (ACC) and average forward/backward transfer (FWT/BWT) across $5$ independent runs with five seeds under the same experimental setup \cite{deng2021flattening}. $\dagger$ $~$ denotes results reported from \cite{deng2021flattening}.}
\vspace{-0.1in}
\resizebox{0.49\textwidth}{!}{
\begin{tabular}{lcccc}
\toprule
\multicolumn{1}{c}{{\textbf{Method}}}&
\multicolumn{2}{c}{\textbf{CIFAR-100}}&
\multicolumn{2}{c}{\textbf{TinyImageNet}} \\

\midrule
 ACC (\%) & FWT / BWT (\%) & ACC (\%) & FWT / BWT (\%) \\
\midrule

La-MaML \cite{gupta2020maml} & 71.37~\scriptsize($\pm$ 0.7)$^\dagger$  & - /  -6.65~\scriptsize($\pm$ 0.9)$^\dagger$  & 66.90~\scriptsize($\pm$ 1.7)$^\dagger$  & - /  -9.13~\scriptsize($\pm$ 0.9)$^\dagger$  \\

GPM \cite{Saha2021} & 73.18~\scriptsize($\pm$ 0.5)$^\dagger$  & - / -0.37~\scriptsize($\pm$ 0.1)$^\dagger$ & 67.39~\scriptsize($\pm$ 0.5)$^\dagger$  & - / ~\textbf{1.45}~\scriptsize($\pm$ \textbf{0.2})$^\dagger$  \\

FS-DGPM \cite{deng2021flattening} & 74.33~\scriptsize($\pm$ 0.3)$^\dagger$  & - / -2.97~\scriptsize($\pm$ 0.4)$^\dagger$ & 70.41~\scriptsize($\pm$ 1.3)$^\dagger$  & - / -2.11~\scriptsize($\pm$ 0.9)$^\dagger$ \\



 
\midrule

PackNet~\cite{mallya2018packnet} & 72.39~\scriptsize($\pm$ 0.3)  & ~0.56 ($\pm 0.8$) / \textbf{0.0} & 55.46~\scriptsize($\pm$ 1.2) & -0.44 ($\pm 0.5$) / \textbf{0.0} \\

SupSup~\cite{wortsman2020supermasks} & 75.47~\scriptsize($\pm$ 0.3)  &  -0.50 ($\pm 0.6$) / \textbf{0.0} & 59.60~\scriptsize($\pm$ 1.1)  & -0.82 ($\pm 0.6$) / \textbf{0.0} \\
\midrule



WSN$^\ast$, $c=50\%$ & 
{77.46}~\scriptsize($\pm$ {0.4}) & -0.26 ($\pm 0.7$) / \textbf{0.0} & 
69.88~\scriptsize($\pm$ 1.7) & -0.33 ($\pm 0.1$) / \textbf{0.0}  \\

WSN$^\ast$, $c=50\%$ + FSO & 
{79.00}~\scriptsize($\pm$ {0.3}) & -0.25 ($\pm 0.6$) / \textbf{0.0} & 
72.04~\scriptsize($\pm$ 0.7) & -0.34 ($\pm 0.2$) / \textbf{0.0}  \\

SoftNet$^\ast$, $c=50\%$ & 
{77.46}~\scriptsize($\pm$ {0.4}) & {30.40} ($\pm 0.7$) / {0.0} & 
{69.88}~\scriptsize($\pm$ 1.7) & {47.80} ($\pm 1.1$) / {0.0}  \\

SoftNet$^\ast$, $c=50\%$ + FSO & 
\textbf{79.00}~\scriptsize($\pm$ {0.3}) & \textbf{30.42} ($\pm 0.6$) / \textbf{0.0} & 
\textbf{72.04}~\scriptsize($\pm$ 0.8) & \textbf{47.81} ($\pm 1.1$) / \textbf{0.0}  \\



\midrule

MTL (Upper-bound) & ~~61.00~\scriptsize($\pm$ 0.2)$^\dagger$ & - / - & ~~77.10~\scriptsize($\pm$ 1.1)$^\dagger$ & - / - \\

\bottomrule
\end{tabular}}
\label{tab:main_soft_til}
\end{center}
\vspace{-0.15in}
\end{table}

%% file: supples/materials/table_conf_wsn_soft.tex
\begin{figure}[ht]
    \centering
    \vspace{-0.1in}
    \setlength{\tabcolsep}{0pt}{%
    \begin{tabular}{cc}
    \includegraphics[width=0.5\columnwidth]{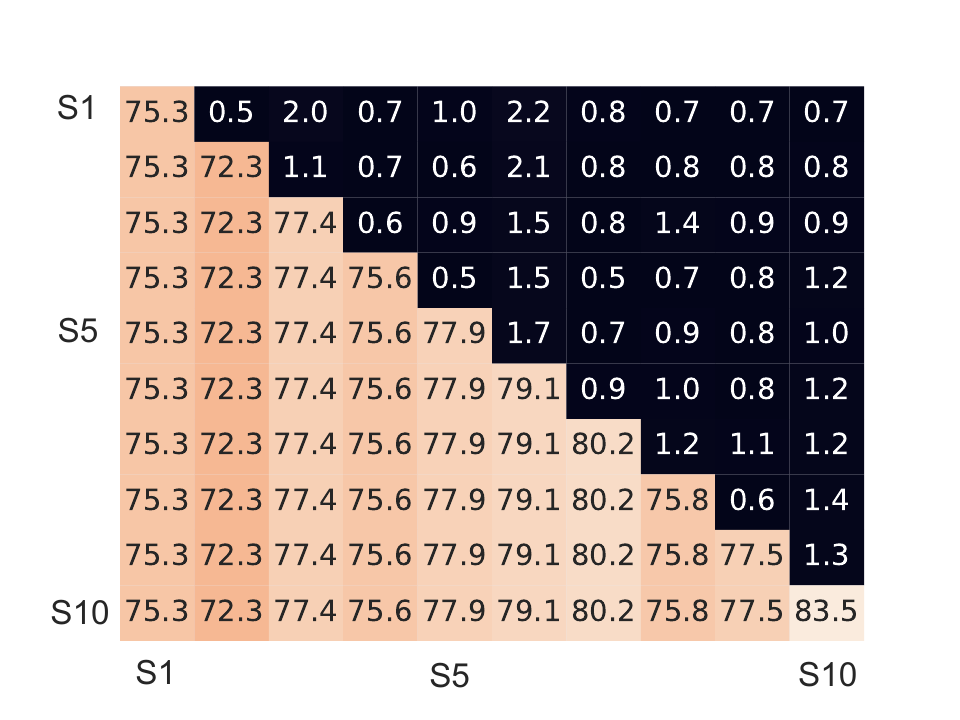} & 
    \includegraphics[width=0.5\columnwidth]{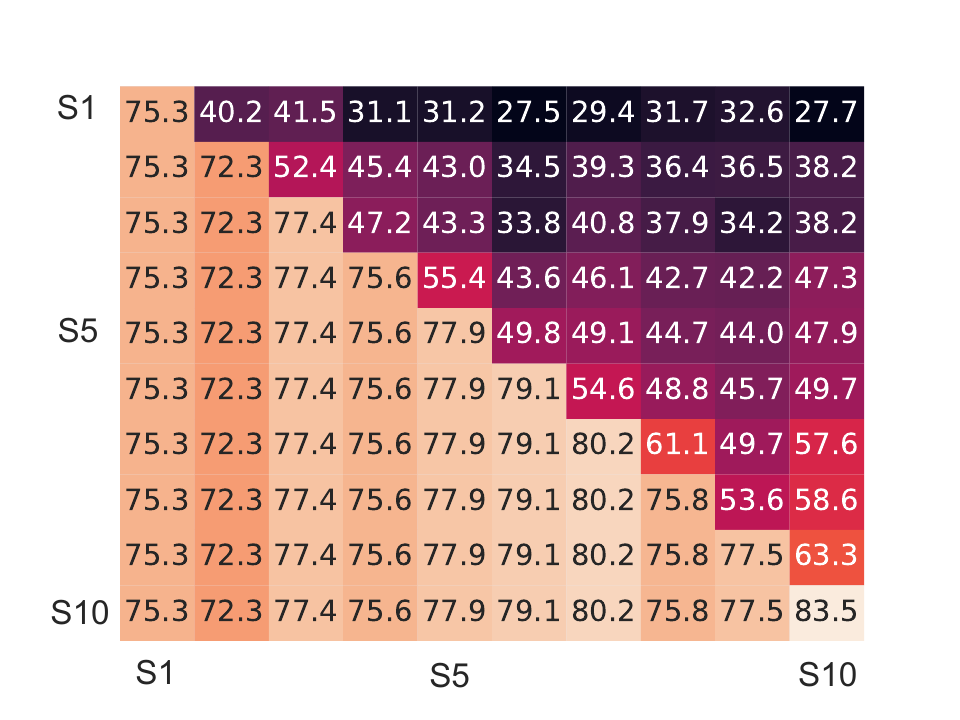} \\
    \small (a) \textcolor{red}{WSN}, $c = 50\%$  & \small (b) \textcolor{cyan}{SoftNet}, $c = 50\%$
    \end{tabular}
    }
    \caption{Forward Transfer (FWT) Matrix on CIFAR-100 Split. (a) WSN v.s. (b){SoftNet}.}
    \label{fig:conf_soft_wsn_cifar100}
    \vspace{-0.12in}
\end{figure}

%% file: supples/materials/table_corr_wsn_soft.tex
\begin{figure}[ht]
    \centering
    \vspace{-0.1in}
    \setlength{\tabcolsep}{0pt}{%
    \begin{tabular}{cc}
    \includegraphics[width=0.5\columnwidth]{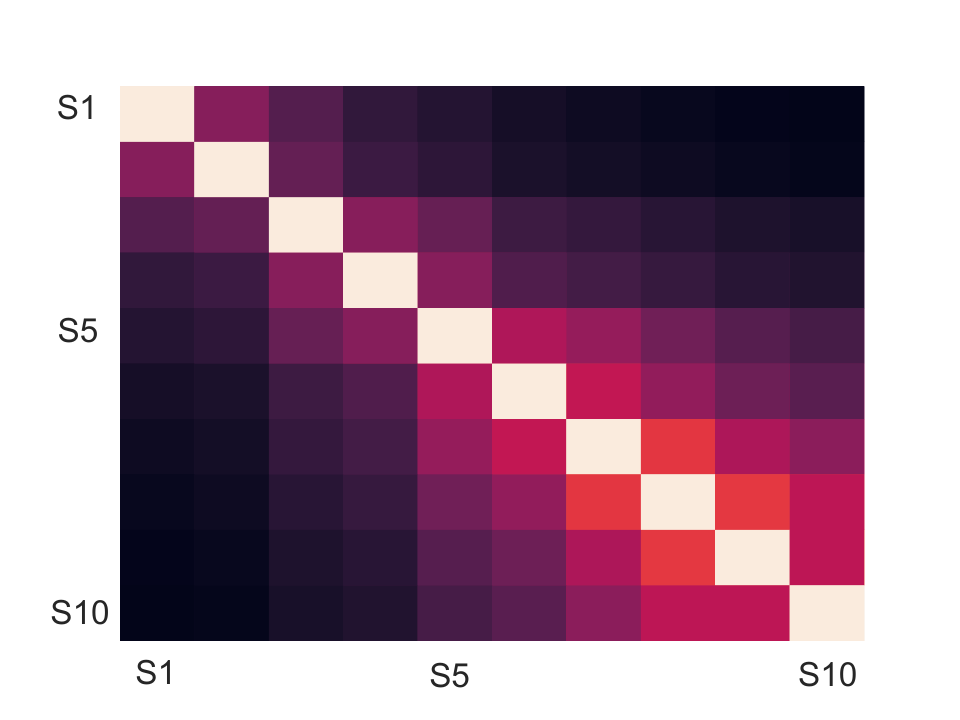} & 
    \includegraphics[width=0.5\columnwidth]{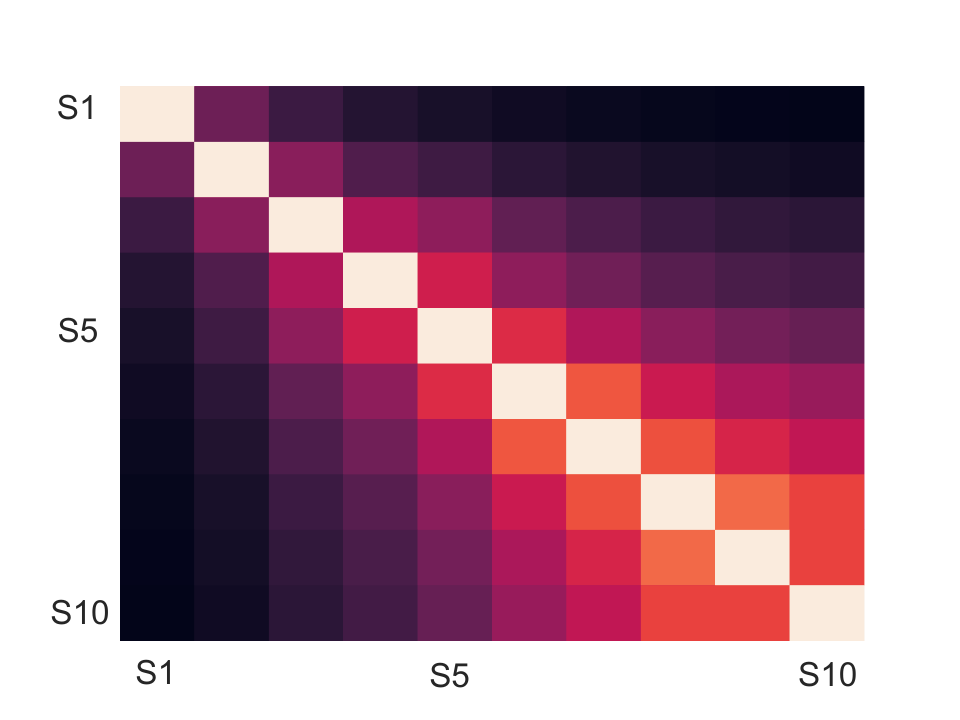} \\
    \small (a) \textcolor{red}{WSN}, $c = 50\%$  & \small (b) \textcolor{cyan}{SoftNet}, $c = 50\%$
    \end{tabular}
    }
    \caption{Average Binary Map (Subnetwork) Correlations on CIFAR-100 Split. (a) WSN v.s. (b) {SoftNet}. The higher the correlation of subnetworks, the better FWT performances.}
    \label{fig:corr_soft_wsn_cifar100}
    \vspace{-0.12in}
\end{figure}

%% file: supples/materials/wsn_prompt.tex
\begin{table}[!ht]
\small
\centering

\caption{\small \textbf{Winning Ticket (ResNets) vs Prompt-Tunning (ViTs)}, Note M (Million), B(Billion).}
\vspace{-0.05in}
\resizebox{0.5\textwidth}{!}{
\begin{tabular}{lll}
\toprule 


& \textbf{Winning Ticket}: & ~~~~~~~~ \textbf{Prompt-Tunning}: \\ 
& ~ResNets: WSN or FSO & ~~~~~~~~~~~~~~~~~ViTs \\ 
\midrule 

Usage of           &                 &  \\ 
Pre-trained on     & ~~~~~ \textbf{No}     & ~~~~~~~~~~~~~~~~~~~~ \textbf{Yes} \\ 
Large-scaled data  &                 & \\ \midrule
Saved Buffer    & \textbf{Task-specific masks} & Prompts (Task-specific prompts) \\ \midrule 
                   & ResNet18: ~~$\sim$\textbf{11M} & ViT-Base~~(12layers, 12heads): ~$\sim$86M \\
Model Capacity     & ResNet50: ~~$\sim$\textbf{25M} & ViT-Large(24layers, 16heads): $\sim$307M \\                        & ResNet101: $\sim$\textbf{44M} & ViT-Huge(32layers, 16heads): $\sim$632M \\ \midrule
                   & ResNet18: ~~$\sim$\textbf{1.8B} & ViT-Base~~(12layers, 12heads): $\sim$~17.6B \\
FLOPs              & ResNet50: ~~$\sim$\textbf{3.6B} & ViT-Large(24layers, 16heads): $\sim$~60.3B \\                       & ResNet101: $\sim$\textbf{4.1B} & ViT-Huge(32layers, 16heads): $\sim$180.8B \\

\bottomrule
\end{tabular}
}
\label{table:wsn_prompt}
\vspace{-0.1in}
\end{table}

%% file: supples/materials/table_acc_wsn_prompt.tex
\begin{table}[!ht]
\small
\centering
\caption{\small \textbf{(Class Incremental Learning, CIL)} the 5-datasets.} 
\vspace{-0.05in}
\resizebox{0.5\textwidth}{!}{
\begin{tabular}{lcccc}
\toprule 

\textbf{Method} & \textbf{Buffer size} & Model(\#params / FLOPs) & ~~~~~~~~ \textbf{5-datasets} & \\
                &     &   &  \small{ACC (\%)} & \small{BWT (\%)} \\
\midrule 
ER~\cite{chaudhry2019tiny}  &  500 & ResNet18($\sim$11M/$\sim$1.8B) & 84.26$\pm0.84$ & 15.69$\pm0.62$\\
BiC~\cite{wu2019large} &  500 & ResNet18($\sim$11M/$\sim$1.8B) & 85.53$\pm2.06$ & 10.27$\pm1.32$\\ 
DER++~\cite{buzzega2020dark} &  500 & ResNet18($\sim$11M/$\sim$1.8B) & 84.88$\pm0.57$ & 10.46$\pm1.02$\\ 
Co$^2$L~\cite{cha2021co2l} &  500 & ResNet18($\sim$11M/$\sim$1.8B) & 86.05$\pm1.03$ & 12.28$\pm1.44$\\ 
\midrule 
ER~\cite{chaudhry2019tiny}  &  250 & ResNet18($\sim$11M/$\sim$1.8B) & 80.32$\pm0.55$ & 15.69$\pm0.89$\\
BiC~\cite{wu2019large} &  250 & ResNet18($\sim$11M/$\sim$1.8B) & 78.74$\pm1.41$ & 21.15$\pm1.00$\\ 
DER++~\cite{buzzega2020dark} &  250 & ResNet18($\sim$11M/$\sim$1.8B) & 80.81$\pm0.07$ & 14.38$\pm0.35$\\ 
Co$^2$L~\cite{cha2021co2l} &  250 & ResNet18($\sim$11M/$\sim$1.8B) & 82.25$\pm1.17$ & 17.52$\pm1.35$\\ 
\midrule 
FT-seq          &  0  & ResNet18($\sim$11M/$\sim$1.8B) & 21.12$\pm0.42$     & 94.64$\pm0.68$\\
EWC~\cite{hu2018overcoming}             &  0  & ResNet18($\sim$11M/$\sim$1.8B) & 50.93$\pm0.09$     & 34.94$\pm0.07$\\
LwF~\cite{LiZ2016eccv}   &  0  & ResNet18($\sim$11M/$\sim$1.8B) & 47.91$\pm0.33$     & 38.01$\pm0.28$\\ \midrule 
\textbf{WSN, c=50.0}\% (ours)      &  -  & ResNet18($\sim$\textbf{5M}/$\sim$\textbf{1.9B}) & \textbf{93.41}$\pm0.13$     & \textbf{0.00}$\pm0.00$\\ \midrule 
L2P~\cite{wang2022learning}   &  -  & ViT-Base($\sim$86B/$\sim$17.6B) & 81.14$\pm0.93$     & 4.64$\pm0.52$\\
DualPrompt~\cite{wang2022dualprompt}     &  -  & ViT-Base($\sim$86B/$\sim$17.6B) & 88.08$\pm0.36$     & 2.21$\pm0.69$\\

\bottomrule
\end{tabular}
}
\label{table:5_data_acc}
\vspace{-0.1in}
\end{table}

%% file: supples/materials/plot_imag_part.tex
\begin{figure}[h]
    \centering
    \small
    \setlength{\tabcolsep}{0pt}{%
    \begin{tabular}{cc}
    \includegraphics[width=0.48\columnwidth]{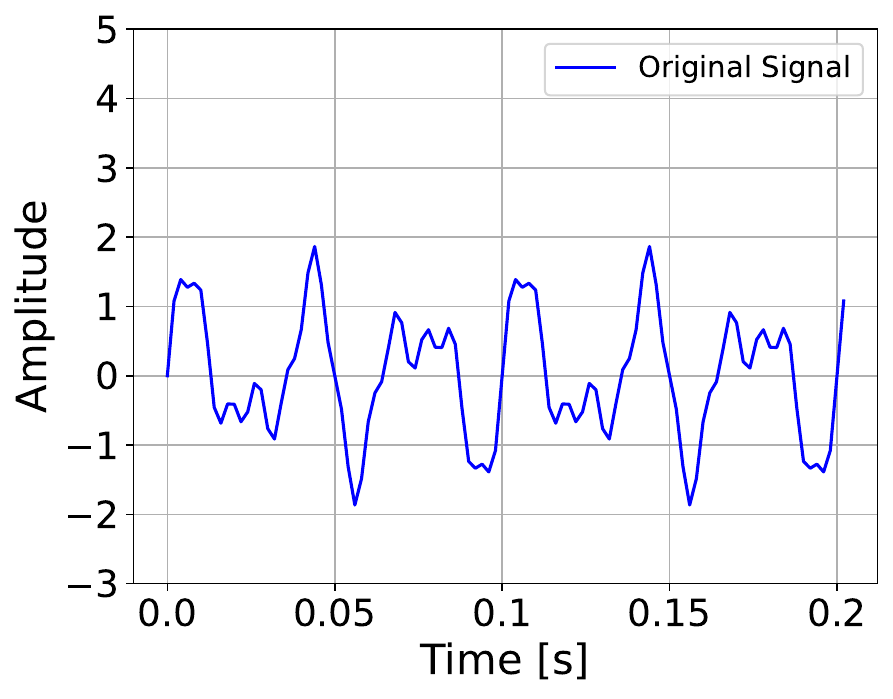} &
    \includegraphics[width=0.48\columnwidth]{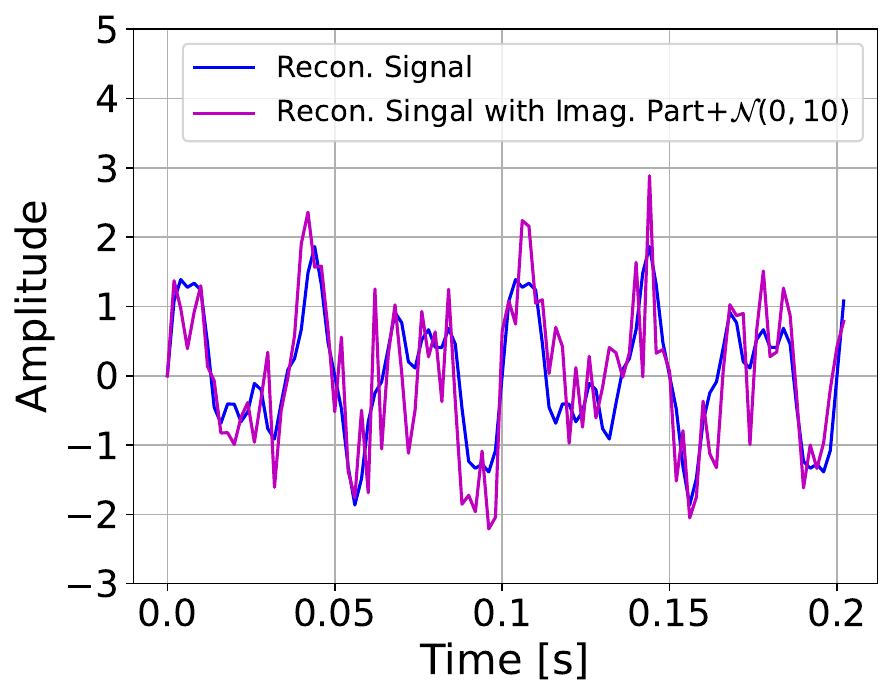} \\
    \small (a) Original Signal & \small (b) Reconstructed Signals \\
    \includegraphics[width=0.48\columnwidth]{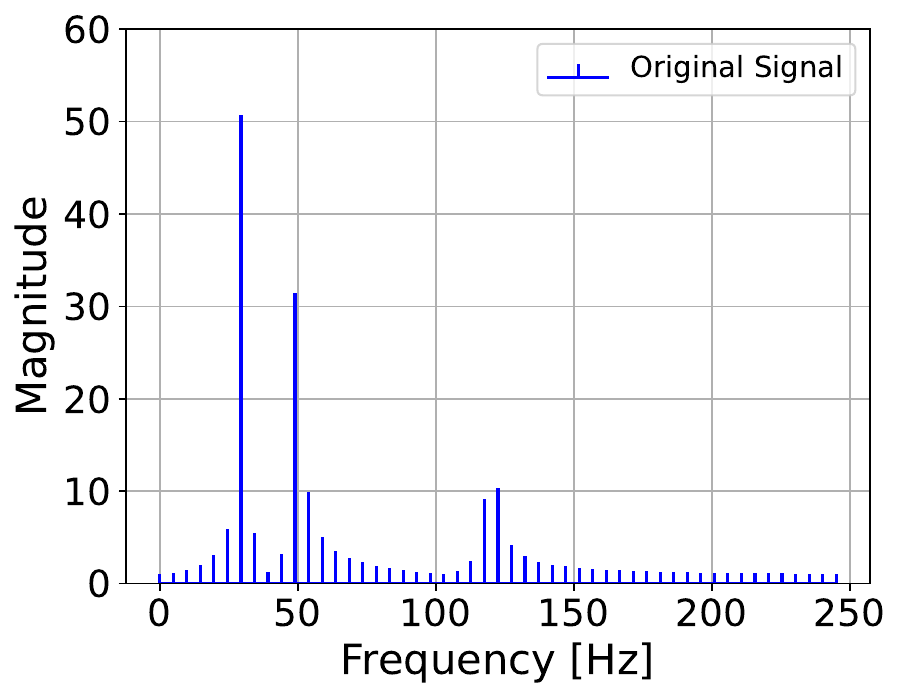} &
    \includegraphics[width=0.48\columnwidth]{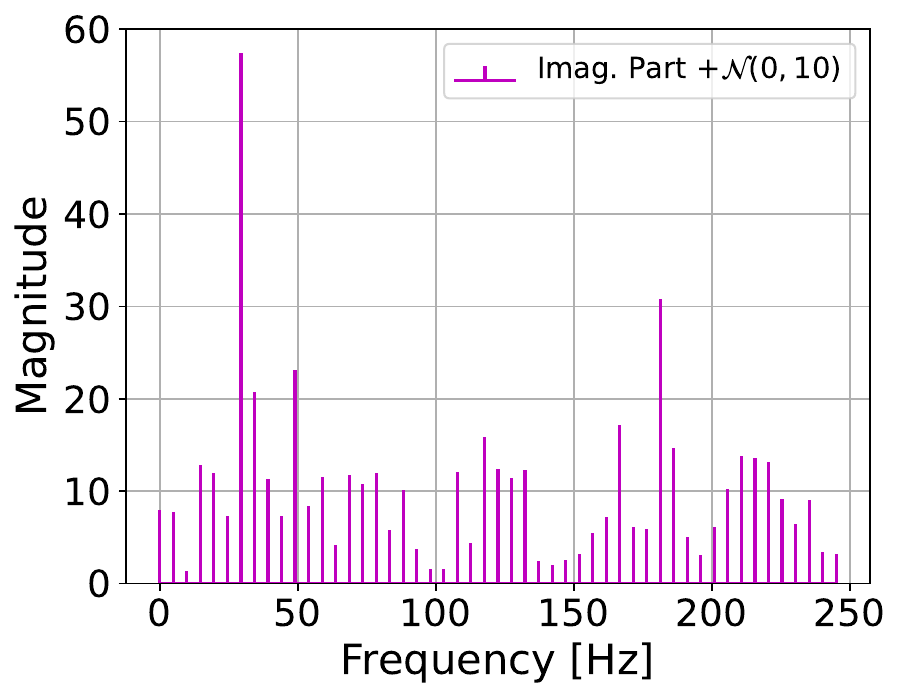} \\
    \vspace{-0.05in}
    \small (c) DFT & \small (d) DFT with Imag. Part+$\mathcal{N}(0,10)$ \\

    \end{tabular}
    }
    \vspace{-0.05in}
    \caption{\small \textbf{FSO of Imaginary Part:} (a) the original signal is a combination of three sine waves at 30, 50, and 120Hz. (b) the reconstructed signals from (c) DFT and (d) DFT with perturbed Imag. Parts. (c) The magnitude spectrum of the signal obtained from the DFT, where the three prominent peaks at 30, 50, and 120Hz. (d) The magnitude spectrum of the perturbed signal, where the random noise $\mathcal{N}(0, 10)$ is added to the Imaginary Parts.}
    \label{fig:imag_part}
    \vspace{-0.15in}
\end{figure}

%% file: main.bbl
\begin{thebibliography}{100}
\providecommand{\url}[1]{#1}
\csname url@samestyle\endcsname
\providecommand{\newblock}{\relax}
\providecommand{\bibinfo}[2]{#2}
\providecommand{\BIBentrySTDinterwordspacing}{\spaceskip=0pt\relax}
\providecommand{\BIBentryALTinterwordstretchfactor}{4}
\providecommand{\BIBentryALTinterwordspacing}{\spaceskip=\fontdimen2\font plus
\BIBentryALTinterwordstretchfactor\fontdimen3\font minus \fontdimen4\font\relax}
\providecommand{\BIBforeignlanguage}[2]{{%
\expandafter\ifx\csname l@#1\endcsname\relax
\typeout{** WARNING: IEEEtran.bst: No hyphenation pattern has been}%
\typeout{** loaded for the language `#1'. Using the pattern for}%
\typeout{** the default language instead.}%
\else
\language=\csname l@#1\endcsname
\fi
#2}}
\providecommand{\BIBdecl}{\relax}
\BIBdecl

\bibitem{ThrunS1995}
S.~Thrun, \emph{A Lifelong Learning Perspective for Mobile Robot Control}.\hskip 1em plus 0.5em minus 0.4em\relax Elsevier, 1995.

\bibitem{rusu2016progressive}
A.~A. Rusu, N.~C. Rabinowitz, G.~Desjardins, H.~Soyer, J.~Kirkpatrick, K.~Kavukcuoglu, R.~Pascanu, and R.~Hadsell, ``Progressive neural networks,'' \emph{arXiv preprint arXiv:1606.04671}, 2016.

\bibitem{zenke2017continual}
F.~Zenke, B.~Poole, and S.~Ganguli, ``Continual learning through synaptic intelligence,'' in \emph{International Conference on Machine Learning}.\hskip 1em plus 0.5em minus 0.4em\relax PMLR, 2017, pp. 3987--3995.

\bibitem{hassabis2017neuroscience}
D.~Hassabis, D.~Kumaran, C.~Summerfield, and M.~Botvinick, ``Neuroscience-inspired artificial intelligence,'' \emph{Neuron}, vol.~95, no.~2, pp. 245--258, 2017.

\bibitem{McCloskey1989}
M.~McCloskey and N.~J. Cohen, ``Catastrophic interference in connectionist networks: The sequential learning problem,'' in \emph{Psychology of learning and motivation}.\hskip 1em plus 0.5em minus 0.4em\relax Elsevier, 1989, vol.~24, pp. 109--165.

\bibitem{Kirkpatrick2017}
J.~Kirkpatrick, R.~Pascanu, N.~Rabinowitz, J.~Veness, G.~Desjardins, A.~A. Rusu, K.~Milan, J.~Quan, T.~Ramalho, A.~Grabska-Barwinska, D.~Hassabis, C.~Clopath, D.~Kumaran, and R.~Hadsell, ``Overcoming catastrophic forgetting in neural networks,'' 2017.

\bibitem{chaudhry2020continual}
A.~Chaudhry, N.~Khan, P.~K. Dokania, and P.~H. Torr, ``Continual learning in low-rank orthogonal subspaces,'' in \emph{Advances in Neural Information Processing Systems (NeurIPS)}, 2020.

\bibitem{Jung2020}
S.~Jung, H.~Ahn, S.~Cha, and T.~Moon, ``Continual learning with node-importance based adaptive group sparse regularization,'' in \emph{Advances in Neural Information Processing Systems (NeurIPS)}, 2020.

\bibitem{titsias2019functional}
M.~K. Titsias, J.~Schwarz, A.~G. d.~G. Matthews, R.~Pascanu, and Y.~W. Teh, ``Functional regularisation for continual learning with gaussian processes,'' in \emph{Proceedings of the International Conference on Learning Representations (ICLR)}, 2020.

\bibitem{mirzadeh2020linear}
S.~I. Mirzadeh, M.~Farajtabar, D.~Gorur, R.~Pascanu, and H.~Ghasemzadeh, ``Linear mode connectivity in multitask and continual learning,'' in \emph{Proceedings of the International Conference on Learning Representations (ICLR)}, 2021.

\bibitem{rebuffi2017icarl}
S.-A. Rebuffi, A.~Kolesnikov, G.~Sperl, and C.~H. Lampert, ``icarl: Incremental classifier and representation learning,'' in \emph{Proceedings of the IEEE conference on Computer Vision and Pattern Recognition}, 2017, pp. 2001--2010.

\bibitem{riemer2018learning}
M.~Riemer, I.~Cases, R.~Ajemian, M.~Liu, I.~Rish, Y.~Tu, and G.~Tesauro, ``Learning to learn without forgetting by maximizing transfer and minimizing interference,'' \emph{arXiv preprint arXiv:1810.11910}, 2018.

\bibitem{chaudhry2018efficient}
A.~Chaudhry, M.~Ranzato, M.~Rohrbach, and M.~Elhoseiny, ``Efficient lifelong learning with a-gem,'' in \emph{Proceedings of the International Conference on Learning Representations (ICLR)}, 2019.

\bibitem{chaudhry2019continual}
A.~Chaudhry, M.~Rohrbach, M.~Elhoseiny, T.~Ajanthan, P.~K. Dokania, P.~H. Torr, and M.~Ranzato, ``Continual learning with tiny episodic memories,'' \emph{arXiv preprint arXiv:1902.10486}, 2019.

\bibitem{Saha2021}
G.~Saha, I.~Garg, and K.~Roy, ``Gradient projection memory for continual learning,'' in \emph{Proceedings of the International Conference on Learning Representations (ICLR)}, 2021.

\bibitem{sarfraz2023error}
\BIBentryALTinterwordspacing
F.~Sarfraz, E.~Arani, and B.~Zonooz, ``Error sensitivity modulation based experience replay: Mitigating abrupt representation drift in continual learning,'' in \emph{The Eleventh International Conference on Learning Representations}, 2023. [Online]. Available: \url{https://openreview.net/forum?id=zlbci7019Z3}
\BIBentrySTDinterwordspacing

\bibitem{mallya2018piggyback}
A.~Mallya, D.~Davis, and S.~Lazebnik, ``Piggyback: Adapting a single network to multiple tasks by learning to mask weights,'' in \emph{Proceedings of the European Conference on Computer Vision (ECCV)}, 2018.

\bibitem{Serra2018}
J.~Serrà, D.~Suris, M.~Miron, and A.~Karatzoglou, ``Overcoming catastrophic forgetting with hard attention to the task,'' in \emph{Proceedings of the International Conference on Machine Learning (ICML)}, 2018.

\bibitem{li2019learn}
X.~Li, Y.~Zhou, T.~Wu, R.~Socher, and C.~Xiong, ``Learn to grow: A continual structure learning framework for overcoming catastrophic forgetting,'' in \emph{Proceedings of the International Conference on Machine Learning (ICML)}, 2019.

\bibitem{wortsman2020supermasks}
M.~Wortsman, V.~Ramanujan, R.~Liu, A.~Kembhavi, M.~Rastegari, J.~Yosinski, and A.~Farhadi, ``Supermasks in superposition,'' in \emph{Advances in Neural Information Processing Systems (NeurIPS)}, 2020.

\bibitem{kang2022forget}
H.~Kang, R.~J.~L. Mina, S.~R.~H. Madjid, J.~Yoon, M.~Hasegawa-Johnson, S.~J. Hwang, and C.~D. Yoo, ``Forget-free continual learning with winning subnetworks,'' in \emph{International Conference on Machine Learning}.\hskip 1em plus 0.5em minus 0.4em\relax PMLR, 2022, pp. 10\,734--10\,750.

\bibitem{kang2022soft}
H.~Kang, J.~Yoon, S.~R.~H. Madjid, S.~J. Hwang, and C.~D. Yoo, ``On the soft-subnetwork for few-shot class incremental learning,'' \emph{arXiv preprint arXiv:2209.07529}, 2022.

\bibitem{mallya2018packnet}
A.~Mallya and S.~Lazebnik, ``Packnet: Adding multiple tasks to a single network by iterative pruning,'' in \emph{Proceedings of the IEEE conference on Computer Vision and Pattern Recognition}, 2018, pp. 7765--7773.

\bibitem{golkar2019continual}
S.~Golkar, M.~Kagan, and K.~Cho, ``Continual learning via neural pruning,'' \emph{arXiv preprint arXiv:1903.04476}, 2019.

\bibitem{Yoon2020}
J.~Yoon, S.~Kim, E.~Yang, and S.~J. Hwang, ``Scalable and order-robust continual learning with additive parameter decomposition,'' in \emph{Proceedings of the International Conference on Learning Representations (ICLR)}, 2020.

\bibitem{Denil2013}
M.~Denil, B.~Shakibi, L.~Dinh, M.~A. Ranzato, and N.~de~Freitas, ``Predicting parameters in deep learning,'' in \emph{Advances in Neural Information Processing Systems (NeurIPS)}, 2013.

\bibitem{Han2016learning_both_weights_struct}
S.~Han, J.~Pool, J.~Tran, and W.~Dally, ``Learning both weights and connections for efficient neural network,'' in \emph{Proceedings of the International Conference on Learning Representations (ICLR)}, 2016.

\bibitem{Li2016pruning_convnets}
H.~Li, A.~Kadav, I.~Durdanovic, H.~Samet, and H.~P. Graf, ``Pruning filters for efficient convnets,'' \emph{arXiv preprint arXiv:1608.08710}, 2016.

\bibitem{frankle2018lottery}
J.~Frankle and M.~Carbin, ``The lottery ticket hypothesis: Finding sparse, trainable neural networks,'' in \emph{Proceedings of the International Conference on Learning Representations (ICLR)}, 2019.

\bibitem{YoonJ2018iclr}
J.~Yoon, E.~Yang, J.~Lee, and S.~J. Hwang, ``Lifelong learning with dynamically expandable networks,'' in \emph{Proceedings of the International Conference on Learning Representations (ICLR)}, 2018.

\bibitem{KumarA2012icml}
A.~Kumar and H.~Daume~III, ``Learning task grouping and overlap in multi-task learning,'' in \emph{Proceedings of the International Conference on Machine Learning (ICML)}, 2012.

\bibitem{LiZ2016eccv}
Z.~Li and D.~Hoiem, ``Learning without forgetting,'' in \emph{Proceedings of the European Conference on Computer Vision (ECCV)}, 2016.

\bibitem{deng2021flattening}
D.~Deng, G.~Chen, J.~Hao, Q.~Wang, and P.-A. Heng, ``Flattening sharpness for dynamic gradient projection memory benefits continual learning,'' in \emph{Advances in Neural Information Processing Systems (NeurIPS)}, 2021.

\bibitem{sun2023decoupling}
W.~Sun, Q.~Li, J.~Zhang, W.~Wang, and Y.-a. Geng, ``Decoupling learning and remembering: A bilevel memory framework with knowledge projection for task-incremental learning,'' in \emph{Proceedings of the IEEE/CVF Conference on Computer Vision and Pattern Recognition}, 2023, pp. 20\,186--20\,195.

\bibitem{mai2021supervised}
Z.~Mai, R.~Li, H.~Kim, and S.~Sanner, ``Supervised contrastive replay: Revisiting the nearest class mean classifier in online class-incremental continual learning,'' in \emph{Proceedings of the IEEE/CVF Conference on Computer Vision and Pattern Recognition}, 2021, pp. 3589--3599.

\bibitem{lin2023pcr}
H.~Lin, B.~Zhang, S.~Feng, X.~Li, and Y.~Ye, ``Pcr: Proxy-based contrastive replay for online class-incremental continual learning,'' in \emph{Proceedings of the IEEE/CVF Conference on Computer Vision and Pattern Recognition}, 2023, pp. 24\,246--24\,255.

\bibitem{aljundi2019online}
R.~Aljundi, E.~Belilovsky, T.~Tuytelaars, L.~Charlin, M.~Caccia, M.~Lin, and L.~Page-Caccia, ``Online continual learning with maximal interfered retrieval,'' in \emph{Advances in Neural Information Processing Systems (NeurIPS)}, 2019.

\bibitem{caccia2021new}
L.~Caccia, R.~Aljundi, N.~Asadi, T.~Tuytelaars, J.~Pineau, and E.~Belilovsky, ``New insights on reducing abrupt representation change in online continual learning,'' \emph{arXiv preprint arXiv:2104.05025}, 2021.

\bibitem{chaudhry2019tiny}
A.~Chaudhry, M.~Rohrbach, M.~Elhoseiny, T.~Ajanthan, P.~K. Dokania, P.~H. Torr, and M.~Ranzato, ``On tiny episodic memories in continual learning,'' \emph{arXiv preprint arXiv:1902.10486}, 2019.

\bibitem{liang2024loss}
Y.-S. Liang and W.-J. Li, ``Loss decoupling for task-agnostic continual learning,'' \emph{Advances in Neural Information Processing Systems}, vol.~36, 2024.

\bibitem{buzzega2020dark}
P.~Buzzega, M.~Boschini, A.~Porrello, D.~Abati, and S.~Calderara, ``Dark experience for general continual learning: a strong, simple baseline,'' \emph{Advances in neural information processing systems}, vol.~33, pp. 15\,920--15\,930, 2020.

\bibitem{ShinH2017nips}
H.~Shin, J.~K. Lee, J.~Kim, and J.~Kim, ``Continual learning with deep generative replay,'' in \emph{Advances in Neural Information Processing Systems (NeurIPS)}, 2017.

\bibitem{xu2018reinforced}
J.~Xu and Z.~Zhu, ``Reinforced continual learning,'' in \emph{Advances in Neural Information Processing Systems (NeurIPS)}, 2018.

\bibitem{yan2021dynamically}
S.~Yan, J.~Xie, and X.~He, ``Der: Dynamically expandable representation for class incremental learning,'' in \emph{Proceedings of the IEEE/CVF Conference on Computer Vision and Pattern Recognition}, 2021, pp. 3014--3023.

\bibitem{singh2020calibrating}
P.~Singh, V.~K. Verma, P.~Mazumder, L.~Carin, and P.~Rai, ``Calibrating cnns for lifelong learning,'' \emph{Advances in Neural Information Processing Systems}, vol.~33, pp. 15\,579--15\,590, 2020.

\bibitem{wang2022learning}
Z.~Wang, Z.~Zhang, C.-Y. Lee, H.~Zhang, R.~Sun, X.~Ren, G.~Su, V.~Perot, J.~Dy, and T.~Pfister, ``Learning to prompt for continual learning,'' in \emph{Proceedings of the IEEE/CVF Conference on Computer Vision and Pattern Recognition}, 2022, pp. 139--149.

\bibitem{wang2022dualprompt}
Z.~Wang, Z.~Zhang, S.~Ebrahimi, R.~Sun, H.~Zhang, C.-Y. Lee, X.~Ren, G.~Su, V.~Perot, J.~Dy \emph{et~al.}, ``Dualprompt: Complementary prompting for rehearsal-free continual learning,'' in \emph{European Conference on Computer Vision}.\hskip 1em plus 0.5em minus 0.4em\relax Springer, 2022, pp. 631--648.

\bibitem{douillard2022dytox}
A.~Douillard, A.~Ram{\'e}, G.~Couairon, and M.~Cord, ``Dytox: Transformers for continual learning with dynamic token expansion,'' in \emph{Proceedings of the IEEE/CVF Conference on Computer Vision and Pattern Recognition}, 2022, pp. 9285--9295.

\bibitem{wang2022s}
Y.~Wang, Z.~Huang, and X.~Hong, ``S-prompts learning with pre-trained transformers: An occam’s razor for domain incremental learning,'' \emph{Advances in Neural Information Processing Systems}, vol.~35, pp. 5682--5695, 2022.

\bibitem{Smith_2023_CVPR}
J.~S. Smith, L.~Karlinsky, V.~Gutta, P.~Cascante-Bonilla, D.~Kim, A.~Arbelle, R.~Panda, R.~Feris, and Z.~Kira, ``Coda-prompt: Continual decomposed attention-based prompting for rehearsal-free continual learning,'' in \emph{Proceedings of the IEEE/CVF Conference on Computer Vision and Pattern Recognition (CVPR)}, June 2023, pp. 11\,909--11\,919.

\bibitem{Smith_2023_CVPR_b}
J.~S. Smith, P.~Cascante-Bonilla, A.~Arbelle, D.~Kim, R.~Panda, D.~Cox, D.~Yang, Z.~Kira, R.~Feris, and L.~Karlinsky, ``Construct-vl: Data-free continual structured vl concepts learning,'' in \emph{Proceedings of the IEEE/CVF Conference on Computer Vision and Pattern Recognition (CVPR)}, June 2023, pp. 14\,994--15\,004.

\bibitem{Pei_2023_ICCV}
Y.~Pei, Z.~Qing, S.~Zhang, X.~Wang, Y.~Zhang, D.~Zhao, and X.~Qian, ``Space-time prompting for video class-incremental learning,'' in \emph{Proceedings of the IEEE/CVF International Conference on Computer Vision (ICCV)}, October 2023, pp. 11\,932--11\,942.

\bibitem{khan2023introducing}
M.~G. Z.~A. Khan, M.~F. Naeem, L.~Van~Gool, D.~Stricker, F.~Tombari, and M.~Z. Afzal, ``Introducing language guidance in prompt-based continual learning,'' in \emph{Proceedings of the IEEE/CVF International Conference on Computer Vision}, 2023, pp. 11\,463--11\,473.

\bibitem{qiao2024prompt}
\BIBentryALTinterwordspacing
J.~Qiao, zhizhong zhang, X.~Tan, C.~Chen, Y.~Qu, Y.~Peng, and Y.~Xie, ``Prompt gradient projection for continual learning,'' in \emph{The Twelfth International Conference on Learning Representations}, 2024. [Online]. Available: \url{https://openreview.net/forum?id=EH2O3h7sBI}
\BIBentrySTDinterwordspacing

\bibitem{he2016deep}
K.~He, X.~Zhang, S.~Ren, and J.~Sun, ``Deep residual learning for image recognition,'' in \emph{Proceedings of the IEEE conference on computer vision and pattern recognition}, 2016, pp. 770--778.

\bibitem{tan2019efficientnet}
M.~Tan and Q.~Le, ``Efficientnet: Rethinking model scaling for convolutional neural networks,'' in \emph{International conference on machine learning}.\hskip 1em plus 0.5em minus 0.4em\relax PMLR, 2019, pp. 6105--6114.

\bibitem{dosovitskiy2020image}
A.~Dosovitskiy, L.~Beyer, A.~Kolesnikov, D.~Weissenborn, X.~Zhai, T.~Unterthiner, M.~Dehghani, M.~Minderer, G.~Heigold, S.~Gelly \emph{et~al.}, ``An image is worth 16x16 words: Transformers for image recognition at scale,'' \emph{arXiv preprint arXiv:2010.11929}, 2020.

\bibitem{wang2021scaled}
C.-Y. Wang, A.~Bochkovskiy, and H.-Y.~M. Liao, ``Scaled-yolov4: Scaling cross stage partial network,'' in \emph{Proceedings of the IEEE/cvf conference on computer vision and pattern recognition}, 2021, pp. 13\,029--13\,038.

\bibitem{liu2021swin}
Z.~Liu, Y.~Lin, Y.~Cao, H.~Hu, Y.~Wei, Z.~Zhang, S.~Lin, and B.~Guo, ``Swin transformer: Hierarchical vision transformer using shifted windows,'' in \emph{Proceedings of the IEEE/CVF international conference on computer vision}, 2021, pp. 10\,012--10\,022.

\bibitem{he2017mask}
K.~He, G.~Gkioxari, P.~Doll{\'a}r, and R.~Girshick, ``Mask r-cnn,'' in \emph{Proceedings of the IEEE international conference on computer vision}, 2017, pp. 2961--2969.

\bibitem{yuan2020object}
Y.~Yuan, X.~Chen, and J.~Wang, ``Object-contextual representations for semantic segmentation,'' in \emph{Computer Vision--ECCV 2020: 16th European Conference, Glasgow, UK, August 23--28, 2020, Proceedings, Part VI 16}.\hskip 1em plus 0.5em minus 0.4em\relax Springer, 2020, pp. 173--190.

\bibitem{vinyals2015show}
O.~Vinyals, A.~Toshev, S.~Bengio, and D.~Erhan, ``Show and tell: A neural image caption generator,'' in \emph{Proceedings of the IEEE conference on computer vision and pattern recognition}, 2015, pp. 3156--3164.

\bibitem{cho2020x}
J.~Cho, J.~Lu, D.~Schwenk, H.~Hajishirzi, and A.~Kembhavi, ``X-lxmert: Paint, caption and answer questions with multi-modal transformers,'' \emph{arXiv preprint arXiv:2009.11278}, 2020.

\bibitem{ho2020denoising}
J.~Ho, A.~Jain, and P.~Abbeel, ``Denoising diffusion probabilistic models,'' \emph{Advances in neural information processing systems}, vol.~33, pp. 6840--6851, 2020.

\bibitem{blattmann2023align}
A.~Blattmann, R.~Rombach, H.~Ling, T.~Dockhorn, S.~W. Kim, S.~Fidler, and K.~Kreis, ``Align your latents: High-resolution video synthesis with latent diffusion models,'' in \emph{Proceedings of the IEEE/CVF Conference on Computer Vision and Pattern Recognition}, 2023, pp. 22\,563--22\,575.

\bibitem{Chen2021lifelonglottery}
T.~Chen, Z.~Zhang, S.~Liu, S.~Chang, and Z.~Wang, ``Long live the lottery: The existence of winning tickets in lifelong learning,'' in \emph{Proceedings of the International Conference on Learning Representations (ICLR)}, 2021.

\bibitem{mehta2021modulated}
I.~Mehta, M.~Gharbi, C.~Barnes, E.~Shechtman, R.~Ramamoorthi, and M.~Chandraker, ``Modulated periodic activations for generalizable local functional representations,'' in \emph{Proceedings of the IEEE/CVF International Conference on Computer Vision}, 2021, pp. 14\,214--14\,223.

\bibitem{sitzmann2020implicit}
V.~Sitzmann, J.~Martel, A.~Bergman, D.~Lindell, and G.~Wetzstein, ``Implicit neural representations with periodic activation functions,'' \emph{Advances in Neural Information Processing Systems}, vol.~33, pp. 7462--7473, 2020.

\bibitem{tancik2020fourier}
M.~Tancik, P.~Srinivasan, B.~Mildenhall, S.~Fridovich-Keil, N.~Raghavan, U.~Singhal, R.~Ramamoorthi, J.~Barron, and R.~Ng, ``Fourier features let networks learn high frequency functions in low dimensional domains,'' \emph{Advances in Neural Information Processing Systems}, vol.~33, pp. 7537--7547, 2020.

\bibitem{chen2019learning}
Z.~Chen and H.~Zhang, ``Learning implicit fields for generative shape modeling,'' in \emph{Proceedings of the IEEE/CVF Conference on Computer Vision and Pattern Recognition}, 2019, pp. 5939--5948.

\bibitem{park2019deepsdf}
J.~J. Park, P.~Florence, J.~Straub, R.~Newcombe, and S.~Lovegrove, ``Deepsdf: Learning continuous signed distance functions for shape representation,'' in \emph{Proceedings of the IEEE/CVF conference on computer vision and pattern recognition}, 2019, pp. 165--174.

\bibitem{mildenhall2021nerf}
B.~Mildenhall, P.~P. Srinivasan, M.~Tancik, J.~T. Barron, R.~Ramamoorthi, and R.~Ng, ``Nerf: Representing scenes as neural radiance fields for view synthesis,'' \emph{Communications of the ACM}, vol.~65, no.~1, pp. 99--106, 2021.

\bibitem{schwarz2020graf}
K.~Schwarz, Y.~Liao, M.~Niemeyer, and A.~Geiger, ``Graf: Generative radiance fields for 3d-aware image synthesis,'' \emph{Advances in Neural Information Processing Systems}, vol.~33, pp. 20\,154--20\,166, 2020.

\bibitem{chen2021nerv}
H.~Chen, B.~He, H.~Wang, Y.~Ren, S.~N. Lim, and A.~Shrivastava, ``Nerv: Neural representations for videos,'' \emph{Advances in Neural Information Processing Systems}, vol.~34, pp. 21\,557--21\,568, 2021.

\bibitem{chen2022cnerv}
H.~Chen, M.~Gwilliam, B.~He, S.-N. Lim, and A.~Shrivastava, ``Cnerv: Content-adaptive neural representation for visual data,'' \emph{arXiv preprint arXiv:2211.10421}, 2022.

\bibitem{he2023towards}
B.~He, X.~Yang, H.~Wang, Z.~Wu, H.~Chen, S.~Huang, Y.~Ren, S.-N. Lim, and A.~Shrivastava, ``Towards scalable neural representation for diverse videos,'' \emph{arXiv preprint arXiv:2303.14124}, 2023.

\bibitem{li2022nerv}
Z.~Li, M.~Wang, H.~Pi, K.~Xu, J.~Mei, and Y.~Liu, ``E-nerv: Expedite neural video representation with disentangled spatial-temporal context,'' in \emph{Computer Vision--ECCV 2022: 17th European Conference, Tel Aviv, Israel, October 23--27, 2022, Proceedings, Part XXXV}.\hskip 1em plus 0.5em minus 0.4em\relax Springer, 2022, pp. 267--284.

\bibitem{maiya2022nirvana}
S.~R. Maiya, S.~Girish, M.~Ehrlich, H.~Wang, K.~S. Lee, P.~Poirson, P.~Wu, C.~Wang, and A.~Shrivastava, ``Nirvana: Neural implicit representations of videos with adaptive networks and autoregressive patch-wise modeling,'' \emph{arXiv preprint arXiv:2212.14593}, 2022.

\bibitem{chen2023hnerv}
H.~Chen, M.~Gwilliam, S.-N. Lim, and A.~Shrivastava, ``Hnerv: A hybrid neural representation for videos,'' \emph{arXiv preprint arXiv:2304.02633}, 2023.

\bibitem{chen2022continual}
G.~Chen, W.~Zhang, H.~Lu, S.~Gao, Y.~Wang, M.~Long, and X.~Yang, ``Continual predictive learning from videos,'' in \emph{Proceedings of the IEEE/CVF Conference on Computer Vision and Pattern Recognition}, 2022, pp. 10\,728--10\,737.

\bibitem{villa2022pivot}
A.~Villa, J.~L. Alc{\'a}zar, M.~Alfarra, K.~Alhamoud, J.~Hurtado, F.~C. Heilbron, A.~Soto, and B.~Ghanem, ``Pivot: Prompting for video continual learning,'' \emph{arXiv preprint arXiv:2212.04842}, 2022.

\bibitem{cho2022streamable}
J.~Cho, S.~Nam, D.~Rho, J.~H. Ko, and E.~Park, ``Streamable neural fields,'' in \emph{European Conference on Computer Vision}.\hskip 1em plus 0.5em minus 0.4em\relax Springer, 2022, pp. 595--612.

\bibitem{kang2024progressive}
\BIBentryALTinterwordspacing
H.~Kang, J.~Yoon, D.~Kim, S.~J. Hwang, and C.~D. Yoo, ``Progressive fourier neural representation for sequential video compilation,'' in \emph{The Twelfth International Conference on Learning Representations}, 2024. [Online]. Available: \url{https://openreview.net/forum?id=rGFrRMBbOq}
\BIBentrySTDinterwordspacing

\bibitem{li2020fourier}
Z.~Li, N.~Kovachki, K.~Azizzadenesheli, B.~Liu, K.~Bhattacharya, A.~Stuart, and A.~Anandkumar, ``Fourier neural operator for parametric partial differential equations,'' \emph{arXiv preprint arXiv:2010.08895}, 2020.

\bibitem{li2020neural}
------, ``Neural operator: Graph kernel network for partial differential equations,'' \emph{arXiv preprint arXiv:2003.03485}, 2020.

\bibitem{kovachki2021neural}
N.~Kovachki, Z.~Li, B.~Liu, K.~Azizzadenesheli, K.~Bhattacharya, A.~Stuart, and A.~Anandkumar, ``Neural operator: Learning maps between function spaces,'' \emph{arXiv preprint arXiv:2108.08481}, 2021.

\bibitem{tran2021factorized}
A.~Tran, A.~Mathews, L.~Xie, and C.~S. Ong, ``Factorized fourier neural operators,'' \emph{arXiv preprint arXiv:2111.13802}, 2021.

\bibitem{hou2019learning}
S.~Hou, X.~Pan, C.~C. Loy, Z.~Wang, and D.~Lin, ``Learning a unified classifier incrementally via rebalancing,'' in \emph{Proceedings of the IEEE/CVF Conference on Computer Vision and Pattern Recognition}, 2019, pp. 831--839.

\bibitem{wu2019large}
Y.~Wu, Y.~Chen, L.~Wang, Y.~Ye, Z.~Liu, Y.~Guo, and Y.~Fu, ``Large scale incremental learning,'' in \emph{Proceedings of the IEEE/CVF Conference on Computer Vision and Pattern Recognition}, 2019, pp. 374--382.

\bibitem{yoon2022online}
\BIBentryALTinterwordspacing
J.~Yoon, D.~Madaan, E.~Yang, and S.~J. Hwang, ``Online coreset selection for rehearsal-based continual learning,'' in \emph{Proceedings of the International Conference on Learning Representations (ICLR)}, 2022. [Online]. Available: \url{https://openreview.net/forum?id=f9D-5WNG4Nv}
\BIBentrySTDinterwordspacing

\bibitem{mazumder2021few}
P.~Mazumder, P.~Singh, and P.~Rai, ``Few-shot lifelong learning,'' \emph{arXiv preprint arXiv:2103.00991}, 2021.

\bibitem{Hinton2012}
G.~Hinton, ``Neural networks for machine learning,'' 2012.

\bibitem{Bengio2013}
Y.~Bengio, N.~L{\'{e}}onard, and A.~C. Courville, ``Estimating or propagating gradients through stochastic neurons for conditional computation,'' \emph{CoRR}, 2013.

\bibitem{Ramanujan2020}
V.~Ramanujan, M.~Wortsman, A.~Kembhavi, A.~Farhadi, and M.~Rastegari, ``What's hidden in a randomly weighted neural network?'' in \emph{Proc. IEEE Conf. Comput. Vis. and Pattern Recognit.}, 2020.

\bibitem{shi2016real}
W.~Shi, J.~Caballero, F.~Husz{\'a}r, J.~Totz, A.~P. Aitken, R.~Bishop, D.~Rueckert, and Z.~Wang, ``Real-time single image and video super-resolution using an efficient sub-pixel convolutional neural network,'' in \emph{Proceedings of the IEEE conference on computer vision and pattern recognition}, 2016, pp. 1874--1883.

\bibitem{shi2021overcoming}
G.~Shi, J.~Chen, W.~Zhang, L.-M. Zhan, and X.-M. Wu, ``Overcoming catastrophic forgetting in incremental few-shot learning by finding flat minima,'' \emph{Advances in Neural Information Processing Systems}, vol.~34, 2021.

\bibitem{mensink2013distance}
T.~Mensink, J.~Verbeek, F.~Perronnin, and G.~Csurka, ``Distance-based image classification: Generalizing to new classes at near-zero cost,'' \emph{IEEE transactions on pattern analysis and machine intelligence}, vol.~35, no.~11, pp. 2624--2637, 2013.

\bibitem{Krizhevsky2009}
A.~Krizhevsky, G.~Hinton \emph{et~al.}, ``Learning multiple layers of features from tiny images,'' 2009.

\bibitem{Stanford}
Stanford, ``Available online at http://cs231n.stanford.edu/tiny-imagenet-200.zip,'' \emph{CS 231N}, 2021.

\bibitem{krizhevsky2012imagenet}
A.~Krizhevsky, I.~Sutskever, and G.~E. Hinton, ``Imagenet classification with deep convolutional neural networks,'' \emph{Advances in neural information processing systems}, vol.~25, pp. 1097--1105, 2012.

\bibitem{LeCun1998}
Y.~LeCun, ``The mnist database of handwritten digits,'' 1998.

\bibitem{gupta2020maml}
G.~Gupta, K.~Yadav, and L.~Paull, ``La-maml: Look-ahead meta learning for continual learning,'' in \emph{Advances in Neural Information Processing Systems (NeurIPS)}, 2020.

\bibitem{chrysakis2020online}
A.~Chrysakis and M.-F. Moens, ``Online continual learning from imbalanced data,'' in \emph{Proceedings of the International Conference on Machine Learning (ICML)}.\hskip 1em plus 0.5em minus 0.4em\relax PMLR, 2020, pp. 1952--1961.

\bibitem{le2015tiny}
Y.~Le and X.~Yang, ``Tiny imagenet visual recognition challenge,'' \emph{CS 231N}, vol.~7, no.~7, p.~3, 2015.

\bibitem{tao2020few}
X.~Tao, X.~Hong, X.~Chang, S.~Dong, X.~Wei, and Y.~Gong, ``Few-shot class-incremental learning,'' in \emph{Proceedings of the IEEE/CVF Conference on Computer Vision and Pattern Recognition}, 2020, pp. 12\,183--12\,192.

\bibitem{chen2020incremental}
K.~Chen and C.-G. Lee, ``Incremental few-shot learning via vector quantization in deep embedded space,'' in \emph{International Conference on Learning Representations}, 2020.

\bibitem{kang2019decoupling}
B.~Kang, S.~Xie, M.~Rohrbach, Z.~Yan, A.~Gordo, J.~Feng, and Y.~Kalantidis, ``Decoupling representation and classifier for long-tailed recognition,'' \emph{arXiv preprint arXiv:1910.09217}, 2019.

\bibitem{loshchilov2016sgdr}
I.~Loshchilov and F.~Hutter, ``Sgdr: Stochastic gradient descent with warm restarts,'' \emph{arXiv preprint arXiv:1608.03983}, 2016.

\bibitem{cheraghian2021semantic}
A.~Cheraghian, S.~Rahman, P.~Fang, S.~K. Roy, L.~Petersson, and M.~Harandi, ``Semantic-aware knowledge distillation for few-shot class-incremental learning,'' in \emph{Proceedings of the IEEE/CVF Conference on Computer Vision and Pattern Recognition}, 2021, pp. 2534--2543.

\bibitem{zhang2021few}
C.~Zhang, N.~Song, G.~Lin, Y.~Zheng, P.~Pan, and Y.~Xu, ``Few-shot incremental learning with continually evolved classifiers,'' in \emph{Proceedings of the IEEE/CVF Conference on Computer Vision and Pattern Recognition}, 2021, pp. 12\,455--12\,464.

\bibitem{zhou2022few}
D.-W. Zhou, H.-J. Ye, L.~Ma, D.~Xie, S.~Pu, and D.-C. Zhan, ``Few-shot class-incremental learning by sampling multi-phase tasks,'' \emph{IEEE Transactions on Pattern Analysis and Machine Intelligence}, 2022.

\bibitem{chi2022metafscil}
Z.~Chi, L.~Gu, H.~Liu, Y.~Wang, Y.~Yu, and J.~Tang, ``Metafscil: A meta-learning approach for few-shot class incremental learning,'' in \emph{Proceedings of the IEEE/CVF Conference on Computer Vision and Pattern Recognition}, 2022, pp. 14\,166--14\,175.

\bibitem{peng2022few}
C.~Peng, K.~Zhao, T.~Wang, M.~Li, and B.~C. Lovell, ``Few-shot class-incremental learning from an open-set perspective,'' in \emph{European Conference on Computer Vision}.\hskip 1em plus 0.5em minus 0.4em\relax Springer, 2022, pp. 382--397.

\bibitem{liu2022few}
H.~Liu, L.~Gu, Z.~Chi, Y.~Wang, Y.~Yu, J.~Chen, and J.~Tang, ``Few-shot class-incremental learning via entropy-regularized data-free replay,'' \emph{arXiv preprint arXiv:2207.11213}, 2022.

\bibitem{hersche2022constrained}
M.~Hersche, G.~Karunaratne, G.~Cherubini, L.~Benini, A.~Sebastian, and A.~Rahimi, ``Constrained few-shot class-incremental learning,'' in \emph{Proceedings of the IEEE/CVF Conference on Computer Vision and Pattern Recognition}, 2022, pp. 9057--9067.

\bibitem{zhou2022forward}
D.-W. Zhou, F.-Y. Wang, H.-J. Ye, L.~Ma, S.~Pu, and D.-C. Zhan, ``Forward compatible few-shot class-incremental learning,'' in \emph{Proceedings of the IEEE/CVF conference on computer vision and pattern recognition}, 2022, pp. 9046--9056.

\bibitem{kim2023warping}
\BIBentryALTinterwordspacing
D.-Y. Kim, D.-J. Han, J.~Seo, and J.~Moon, ``Warping the space: Weight space rotation for class-incremental few-shot learning,'' in \emph{The Eleventh International Conference on Learning Representations}, 2023. [Online]. Available: \url{https://openreview.net/forum?id=kPLzOfPfA2l}
\BIBentrySTDinterwordspacing

\bibitem{akyurek2021subspace}
A.~F. Aky{\"u}rek, E.~Aky{\"u}rek, D.~Wijaya, and J.~Andreas, ``Subspace regularizers for few-shot class incremental learning,'' \emph{arXiv preprint arXiv:2110.07059}, 2021.

\bibitem{zhu2021self}
K.~Zhu, Y.~Cao, W.~Zhai, J.~Cheng, and Z.-J. Zha, ``Self-promoted prototype refinement for few-shot class-incremental learning,'' in \emph{Proceedings of the IEEE/CVF Conference on Computer Vision and Pattern Recognition}, 2021, pp. 6801--6810.

\bibitem{yang2023neural}
Y.~Yang, H.~Yuan, X.~Li, Z.~Lin, P.~Torr, and D.~Tao, ``Neural collapse inspired feature-classifier alignment for few-shot class incremental learning,'' \emph{arXiv preprint arXiv:2302.03004}, 2023.

\bibitem{boyd2004convex}
S.~Boyd, S.~P. Boyd, and L.~Vandenberghe, \emph{Convex optimization}.\hskip 1em plus 0.5em minus 0.4em\relax Cambridge university press, 2004.

\bibitem{bottou2018optimization}
L.~Bottou, F.~E. Curtis, and J.~Nocedal, ``Optimization methods for large-scale machine learning,'' \emph{Siam Review}, vol.~60, no.~2, pp. 223--311, 2018.

\bibitem{ye2020good}
M.~Ye, C.~Gong, L.~Nie, D.~Zhou, A.~Klivans, and Q.~Liu, ``Good subnetworks provably exist: Pruning via greedy forward selection,'' in \emph{International Conference on Machine Learning}.\hskip 1em plus 0.5em minus 0.4em\relax PMLR, 2020, pp. 10\,820--10\,830.

\bibitem{hart1968condensed}
P.~Hart, ``The condensed nearest neighbor rule (corresp.),'' \emph{IEEE transactions on information theory}, vol.~14, no.~3, pp. 515--516, 1968.

\bibitem{dasarathy1980nosing}
B.~V. Dasarathy, ``Nosing around the neighborhood: A new system structure and classification rule for recognition in partially exposed environments,'' \emph{IEEE Transactions on Pattern Analysis and Machine Intelligence}, no.~1, pp. 67--71, 1980.

\bibitem{cha2021co2l}
H.~Cha, J.~Lee, and J.~Shin, ``Co2l: Contrastive continual learning,'' in \emph{Proceedings of the IEEE/CVF International conference on computer vision}, 2021, pp. 9516--9525.

\bibitem{hu2018overcoming}
W.~Hu, Z.~Lin, B.~Liu, C.~Tao, Z.~Tao, J.~Ma, D.~Zhao, and R.~Yan, ``Overcoming catastrophic forgetting for continual learning via model adaptation,'' in \emph{International conference on learning representations}, 2018.

\end{thebibliography}
